\documentclass[conference]{IEEEtran}
\IEEEoverridecommandlockouts
\usepackage{cite}
\usepackage{amsmath,amssymb,amsfonts}
\usepackage{algorithmic}
\usepackage{graphicx}
\usepackage{textcomp}
\usepackage{xcolor}
\usepackage{gensymb}
\usepackage{caption}
\usepackage{subcaption}
\usepackage{multirow,booktabs}

\usepackage{makecell}

\def\BibTeX{{\rm B\kern-.05em{\sc i\kern-.025em b}\kern-.08em
    T\kern-.1667em\lower.7ex\hbox{E}\kern-.125emX}}
\begin{document}

\title{Lifting 2d Human Pose to 3d : A Weakly Supervised Approach}
\author{\IEEEauthorblockN{Sandika Biswas,
Sanjana Sinha, Kavya Gupta and
Brojeshwar Bhowmick}
\IEEEauthorblockA{Embedded Systems and Robotics,\\
TCS Research and Innovation\\
Email: \{{biswas.sandika, sanjana.sinha, gupta.kavya, b.bhowmick\}}@tcs.com,
}
}

\maketitle

\begin{abstract}

Estimating 3d human pose from monocular images is a challenging problem due to the variety and complexity of
human poses and the inherent ambiguity in recovering depth from the single view. Recent deep learning based methods show promising results by using supervised learning on 3d pose annotated datasets. However, the lack of large-scale 3d annotated training data captured under in-the-wild settings makes the 3d pose estimation difficult for in-the-wild poses. Few approaches have utilized training images from both 3d and 2d pose datasets in a weakly-supervised manner for learning 3d poses in unconstrained settings. In this paper, we propose a method which can effectively predict 3d human pose from 2d pose using a deep neural network trained in a weakly-supervised manner on a combination of ground-truth 3d pose and ground-truth 2d pose. Our method uses re-projection error minimization as a constraint to predict the 3d locations of body joints, and this is crucial for training on data where the 3d ground-truth is not present. Since minimizing re-projection error alone may not guarantee an accurate 3d pose, we also use additional geometric constraints on skeleton pose to regularize the pose in 3d. We demonstrate the superior generalization ability of our method by cross-dataset validation on a challenging 3d benchmark dataset MPI-INF-3DHP containing in the wild 3d poses. 

\end{abstract}


\section{Introduction}

Human pose estimation from images and videos is a fundamental problem in computer vision which has a variety of applications such as virtual reality, gaming, surveillance, human-computer interaction,  health-care \cite{upperbody,SLS}, etc. Estimating the shape of the human skeleton in 3d from a single image \cite{mehta2017monocular, pavlakos2017coarse, zhou2017towards, luo2018orinet, tome2017lifting, yang20183d, lee2018propagating} or video \cite{lin2017recurrent,hossain2018exploiting,zhou2016sparseness,dabral2018learning} is a much more challenging problem than estimating the pose in 2d \cite{newell2016stacked,cao2017realtime} due to the inherent ambiguity of estimating depth from a single view. 
Due to the availability of large scale 2d pose annotated datasets \cite{andriluka14cvpr}, the state-of-the-art deep supervised learning-based methods for 2d pose estimation have successfully been able to generalize to new ``in-the-wild" images \cite{newell2016stacked}. These are the images that are not captured under any specific scene settings or pose restrictions. However, the well-known 3d pose datasets \cite{sigal2010humaneva,h36m_pami} contain 3d motion capture (MoCap) data recorded in controlled setup in indoor settings. Hence, 3d supervised learning methods  \cite{martinez2017simple,pavlakos2017coarse,lee2018propagating,tome2017lifting} do not generalize well to datasets in the wild where 3d ground-truth is not present. 

\begin{figure}[t!]
        \centering
        \begin{tabular}{cc}
            \includegraphics[keepaspectratio=true, scale = 0.23]{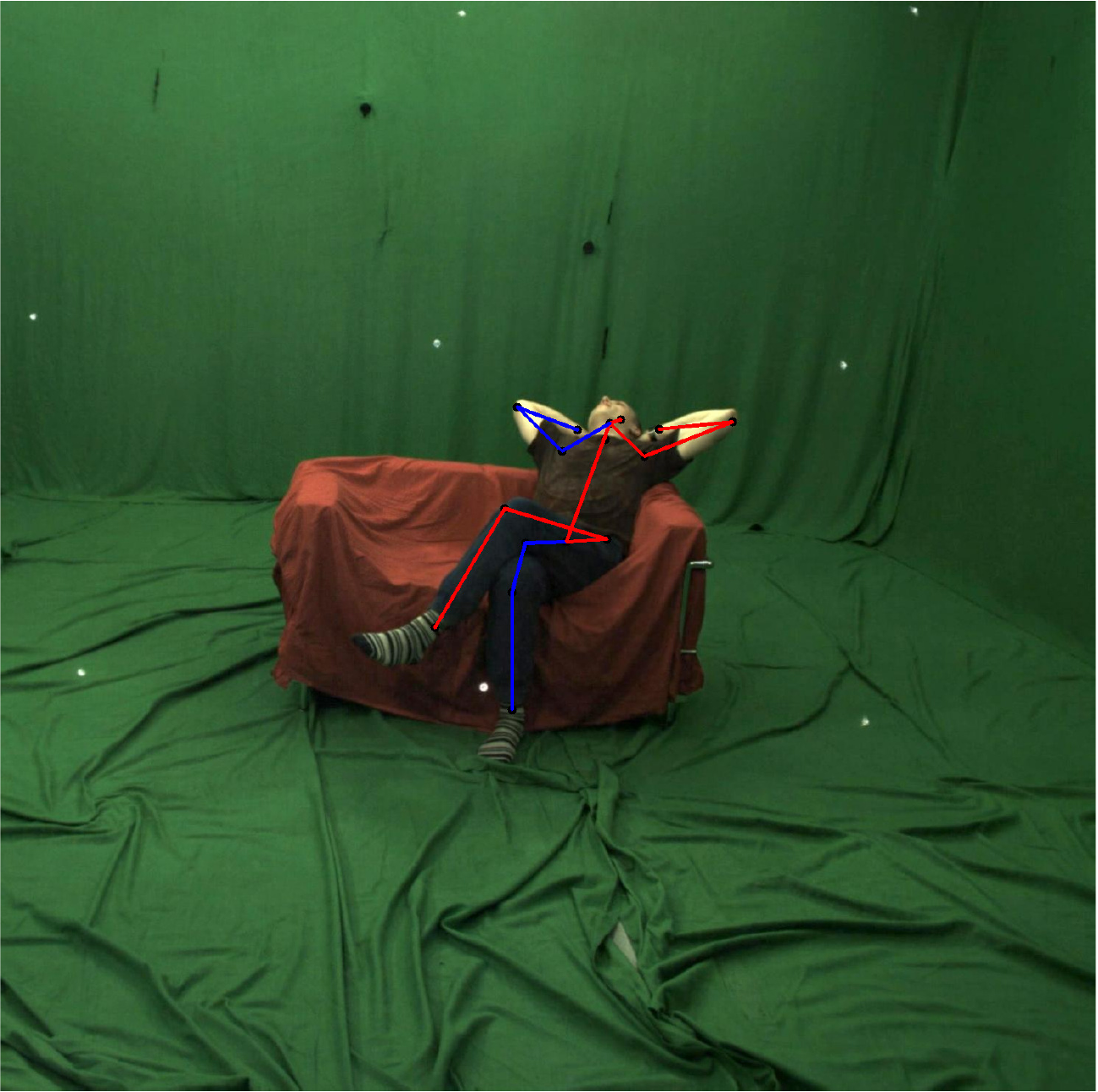} & \includegraphics[keepaspectratio=true, scale = 0.24]{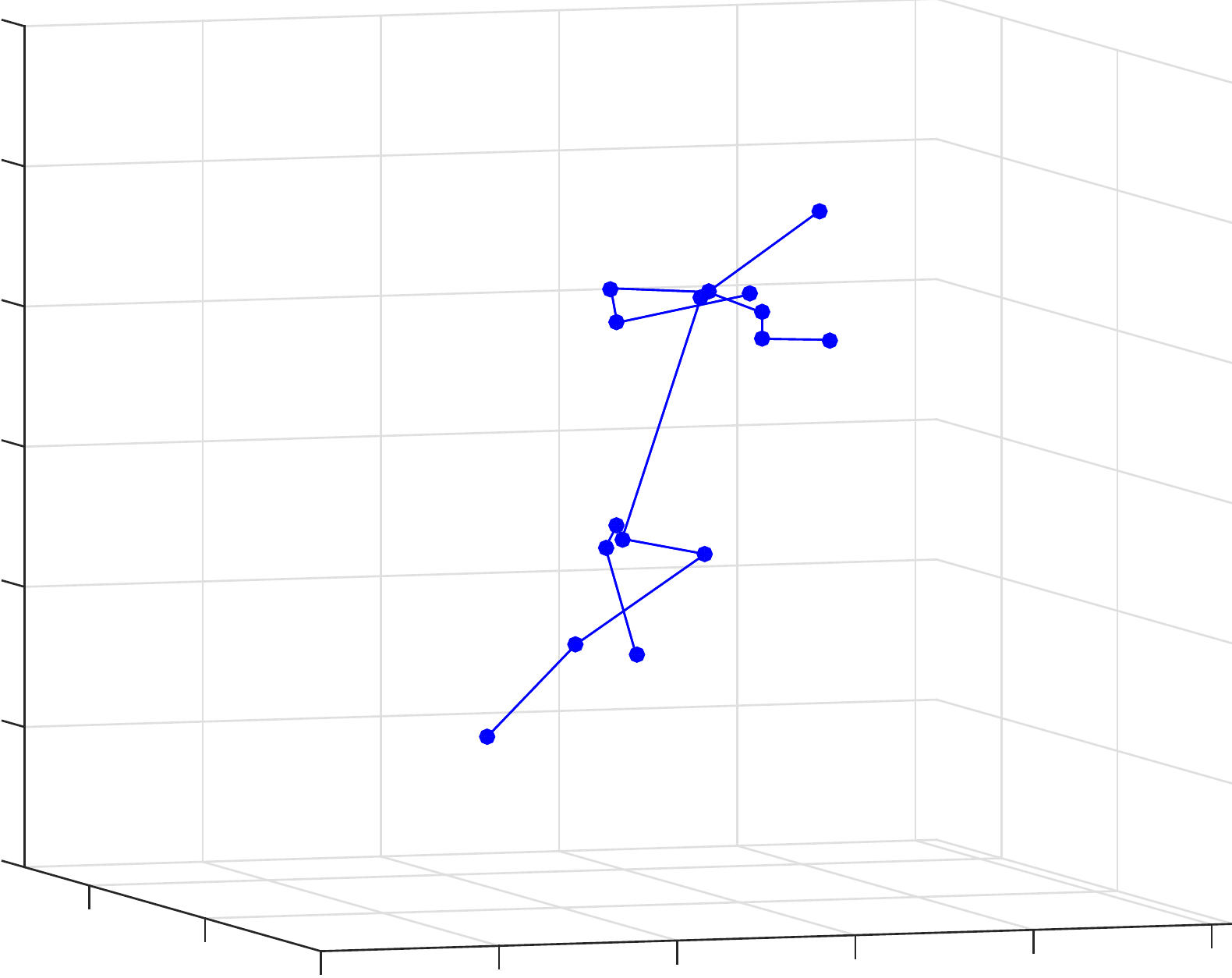} \\
                    \end{tabular}
        \caption{\textbf{Left}: Image from MPI-INF-3DHP test dataset. \textbf{Right}: 3d pose prediction from state-of-the-art weakly-supervised method proposed by Zhou et al. which does not capture the pose correctly.}
        \label{fig:errorweakly}
    \end{figure}
    
    \begin{figure}[t!]
        \centering
        \begin{tabular}{cc}
            \includegraphics[keepaspectratio=true, scale = 0.21]{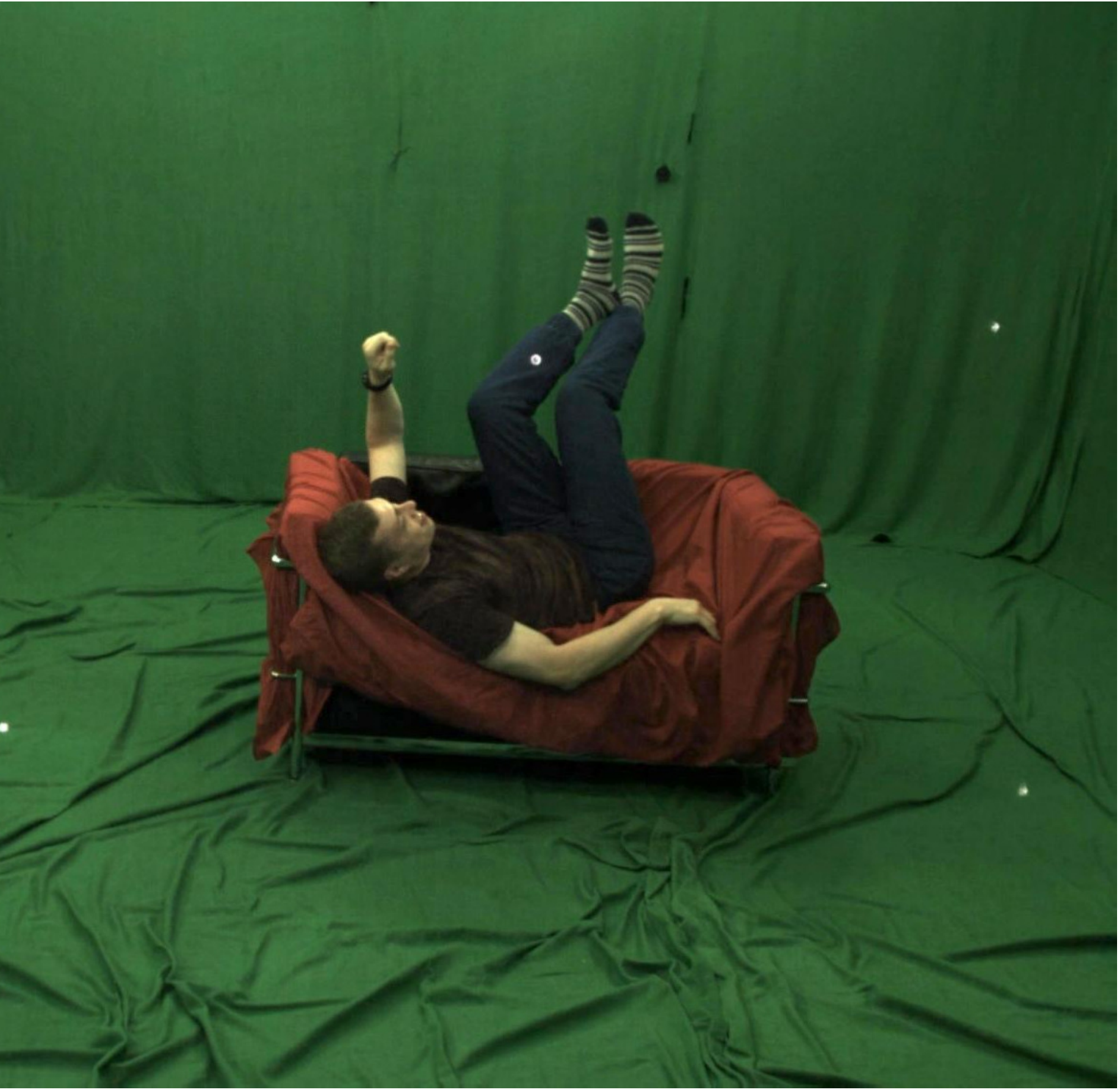} & \includegraphics[keepaspectratio=true, scale = 0.33]{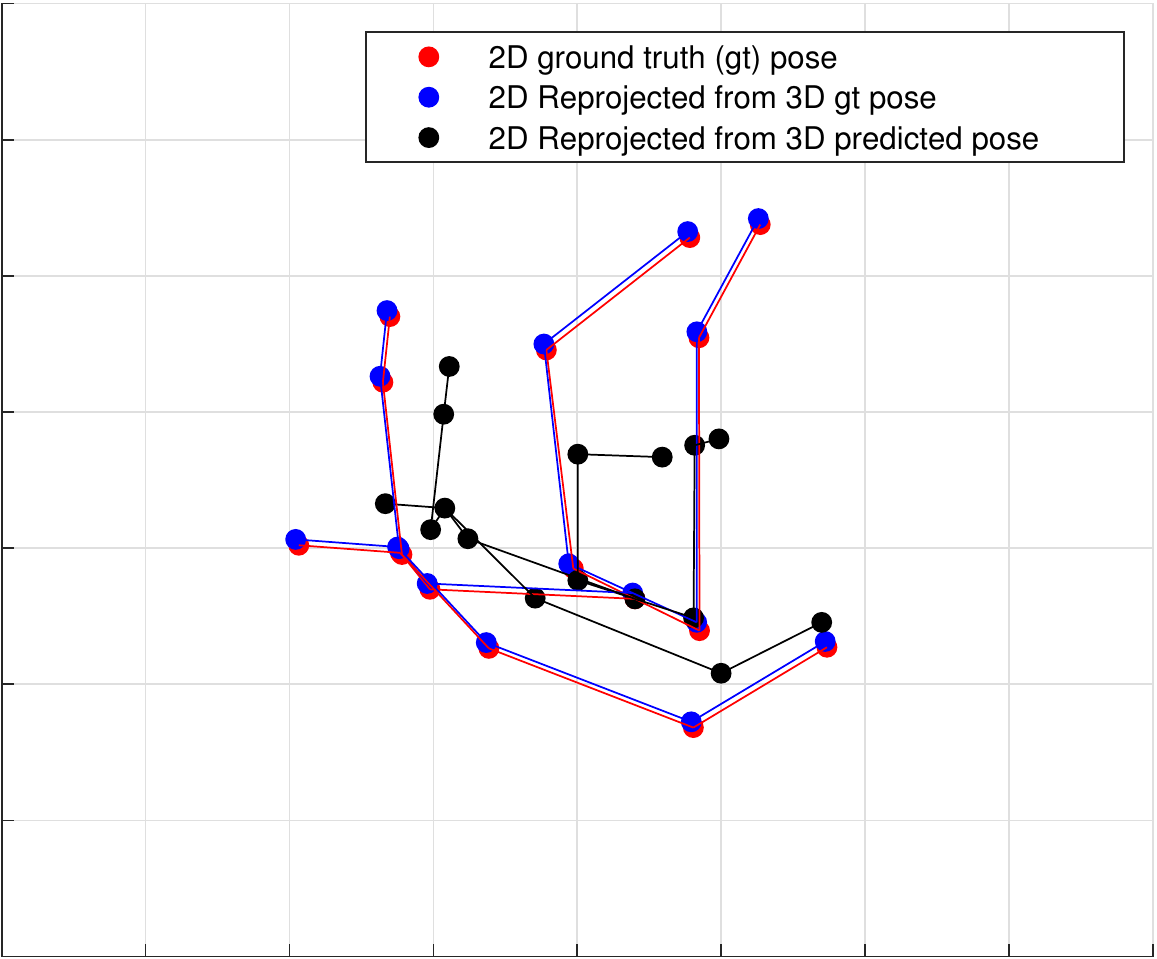} \\
                    \end{tabular}
        \caption{\textbf{Left}: Image from MPI-INF-3dHP test dataset. \textbf{Right}: Comparison of 2d ground truth pose (red), 2d re-projection of 3d ground truth pose (blue) and 2d re-projection of predicted 3d pose (using camera parameters) from state-of-the-art method (black). There is a significant error between 2d re-projection of predicted 3d pose and ground truth 2d pose which we want to minimize to improve the predicted 3d pose. }
        \label{fig: motivation}
    \end{figure}

Almost all the recent methods for monocular 3d pose estimation fall under one of these three approaches - (i) Estimating 3d pose from images directly using full 3d supervision \cite{pavlakos2017coarse,mehta2017monocular,mehta2017vnect}, (ii) Estimating 3d pose from ground-truth 2d pose using full 3d supervision \cite{martinez2017simple} (iii) Estimating 3d pose directly from images using weakly-supervised learning \cite{dabral2018learning,zhou2017towards}. Approach (ii) has been shown to be more effective than approach (i) since the 2d pose input makes the process of lifting 2d pose to 3d invariant to image-related factors such as illumination, background, occlusion, etc. which adversely affect the overall accuracy of 3d pose estimation. Both the approaches (i) and (ii) produce very high accuracy on the popular 3d benchmark datasets which are captured under controlled settings, but may fail to generalize well if the pose or scene is very different from 3d training examples \cite{martinez2017simple}. On the other hand, weakly-supervised methods use 2d pose ground-truth from 2d pose datasets as weak labels in addition to the 3d pose ground-truth from 3d pose datasets. Since 2d datasets contain poses in the wild \cite{andriluka14cvpr}, the generalization of these methods is higher than the fully-supervised methods following approaches in (i) and (ii). However, the current methods of weakly-supervised learning are carried out as a two-step approach by first predicting 2d pose from image and then regressing joint depth in a single end-to-end network \cite{zhou2017towards,dabral2018learning}. Training a network using such an approach is crucially dependent on the accuracy of the 2d pose detector.
Due to the depth regression, the accuracy of 2d pose prediction is adversely affected in an end-to-end pipeline \cite{zhou2017towards}. If the 2d pose accuracy is lowered, the 3d pose accuracy also degrades.
Figure~\ref{fig:errorweakly} shows an example where a weakly supervised approach \cite{zhou2017towards}, which compute 2d pose and 3d pose jointly, produces incorrect joint locations due to incorrect estimation of intermediate 2d pose during training. 

In this work, we address the following problem - given ground-truth 2d poses in the wild, can 3d poses be recovered with sufficient accuracy even in the absence of 3d ground truth? To the best of our knowledge, this is the first work that addresses this specific problem that combines the motivation behind both approaches of (ii) and (iii). We propose a weakly supervised approach for 3d pose estimation from given 2d pose. We use a simple deep network that consists of a 2d-to-3d pose regression module and a 3d-to-2d pose re-projection module. The advantage of using our network is that it can be simultaneously trained on data from both 3d pose datasets and 2d pose datasets (without 3d pose annotations).
Our 2d-to-3d pose regression module is similar to the state-of-the-art network \cite{martinez2017simple} with the difference that a) our learning is weakly-supervised instead of being fully supervised and b) it can be trained on any dataset containing only ground-truth 2d labels. Our 3d-to-2d pose re-projection module is designed to ensure that the predicted 3d pose re-projects correctly to input 2d pose, which is not incorporated in the existing fully supervised method of \cite{martinez2017simple}, as shown in Figure~\ref{fig: motivation}. In the absence of ground truth 3d pose, the predicted 3d pose is constrained in our method by minimization of re-projection error with respect to the 2d pose input. Our 3d-to-2d regression network does not require the knowledge of camera parameters and hence can be used on any arbitrary images without known camera parameters (e.g., images from MPII dataset \cite{andriluka14cvpr}). This simple approach of ``lifting" 2d pose to 3d and subsequently re-projecting 3d to 2d enables joint training on in-the-wild 2d pose datasets that do not contain 3d pose ground-truth. Our approach differs from other weakly-supervised methods for 3d pose estimation as we do not address the problem of 2d pose estimation and focus only on the effective learning of 3d poses from ground truth 2d poses, even when datasets do not contain 3d pose labels. \\
\vspace{-6pt}
The main contributions of our paper are outlined below:
\begin{itemize}
\item We propose a network for predicting 3d human pose from 2d pose that can be trained in a weakly-supervised manner using 3d pose annotated data as well as data with only 2d pose annotations.
\item In addition to the standard 2d-to-3d pose regression, we introduce a 3d-to-2d re-projection network that minimizes 3d pose re-projection error in order to predict accurate 3d pose in the absence of 3d ground-truth. This re-projection error is also used to refine predicted 3d pose while 3d ground truth is available. The 3d-to-2d projection network can be trained on data with unknown camera parameters. 
\item By training on a mixture of 2d and 3d pose datasets, our method outperforms state-of-the-art 3d pose estimation methods on a benchmark 3d dataset containing challenging poses in the wild.
\end{itemize}

\begin{figure*}
\centering
\includegraphics[keepaspectratio=true, scale = 0.7]{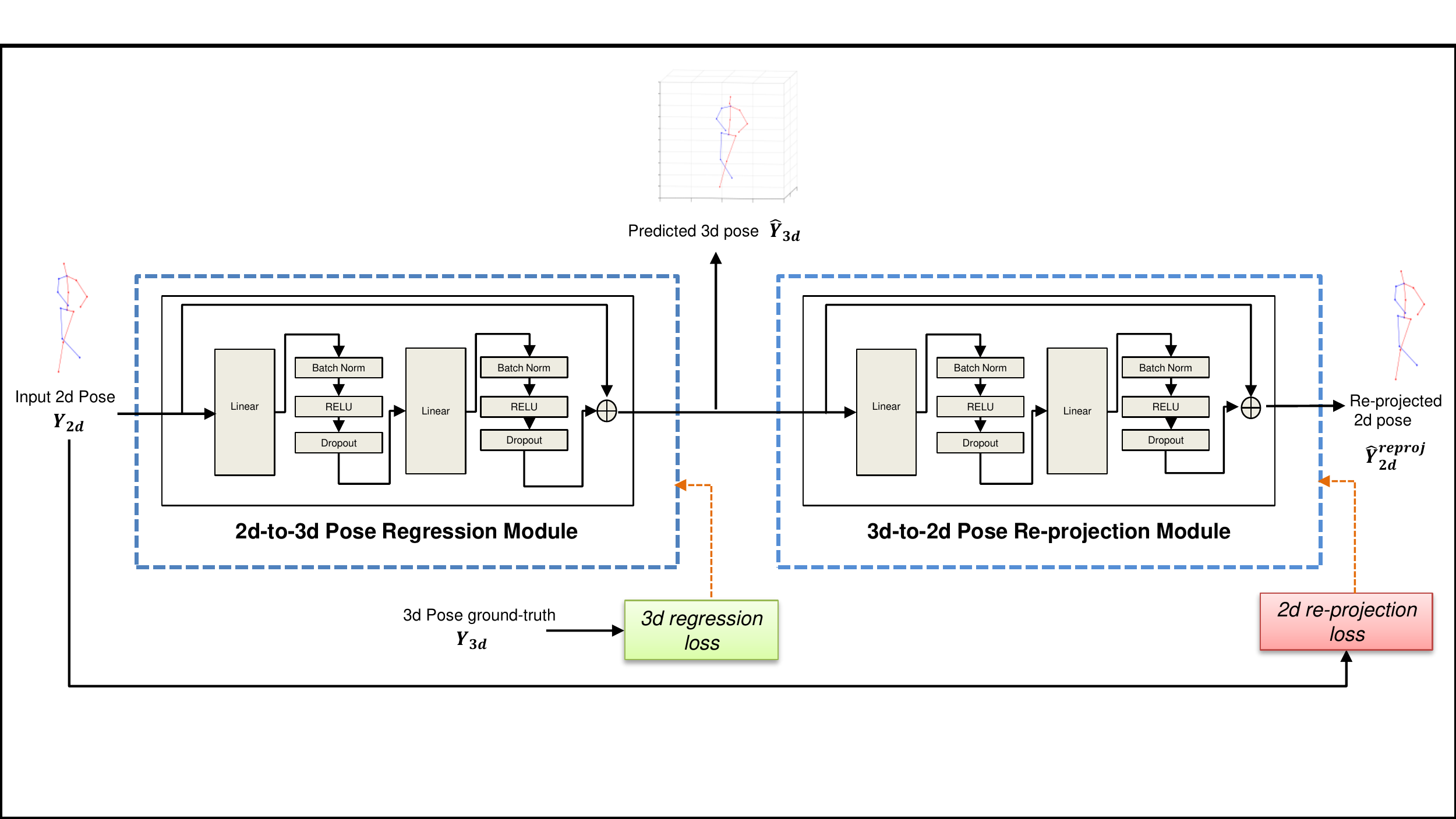}
\caption{Proposed network architecture. Each of the two network modules, i.e 2d-to-3d regression module and 3d-to-2d re-projection module use the architecture of Martinez et al. }
\label{fig:block_diagram}
\end{figure*}

\section{Related Work}

\textbf{Monocular 3d human pose estimation:} The monocular 3d human pose estimation problem is to learn the 3d structure of a human skeleton from a single image or a video, without using depth information or multiple views. It is a severely ill-posed problem which has been formulated in recent literature \cite{pavlakos2017coarse,zhou2017towards,luo2018orinet,tome2017lifting,yang20183d,lee2018propagating,lin2017recurrent,hossain2018exploiting,zhou2016sparseness,dabral2018learning,mehta2017monocular,mehta2017vnect} as a supervised learning problem given the availability of 3d human motion capture datasets \cite{h36m_pami,sigal2010humaneva}. Most of these works focus on end-to-end 3d pose estimation from single images \cite{zhou2017towards,luo2018orinet,tome2017lifting,yang20183d,lee2018propagating,zhou2016sparseness,dabral2018learning,mehta2017monocular,mehta2017vnect}, while some utilize temporal sequences for estimating 3d pose from video \cite{hossain2018exploiting,lin2017recurrent}. Our work is related to the problem of 3d pose estimation from a single image, which can be applicable for videos, but without utilizing any temporal information.

\textbf{2d pose to 3d pose:} Recent research works have approached the problem of estimating 3d poses from 2d poses, which are learned apriori from images \cite{hossain2018exploiting,martinez2017simple,fang2018learning}. These methods use 2d pose detections from state-of-the-art 2d pose detector \cite{newell2016stacked}, which provides invariance to illumination changes, background clutter, clothing variation, etc. By decoupling the two stages, it is also possible to infer the accuracy of ``lifting" ground truth 2d pose to 3d. The current state-of-the-art work by Martinez et al.\cite{martinez2017simple} uses a deep feedforward network that takes a 2d human pose as input and estimates 3d pose with very high accuracy using a simple network. Their results suggest the effectiveness of decoupling the 3d pose estimation problem into two separate problems - namely, 2d pose estimation from an image and 3d pose estimation from a 2d pose. Their 3d pose detector trained on ground-truth 3d poses achieved a remarkable improvement in accuracy (30 \%), leading to the implication that the accuracy of 2d pose estimation remains a bottleneck in end-to-end 3d pose estimation. Inspired by their work, we address the problem of learning 3d poses from known 2d poses of high accuracy. However, the fully supervised method of Martinez et al. \cite{martinez2017simple} can fail to recover 3d pose from ground truth 2d pose accurately if the 2d pose is considerably different from Human3.6m training examples or contain occluded or cropped human poses\cite{martinez2017simple}. This has led us to address the problem of effectively learning 3d poses from 2d pose data with greater pose diversity along with the existing 3d pose datasets. 

\textbf{Weakly-supervised learning of 3d pose:} In this approach, 
2d pose datasets (without 3d annotations) have been used for training 3d pose estimation network for simultaneous 2d and 3d pose prediction \cite{sun2017compositional} and weakly-supervised learning of 3d poses from images \cite{zhou2017towards,dabral2018learning}. In the absence of 3d ground truth labels, there can be a number of 3d poses, which when projected back gives the same pose in 2d; hence the 3d poses must follow geometric validity constraints, such as bone length ratio \cite{zhou2017towards}, illegal angle constraints \cite{dabral2018learning,luo2018orinet}. However, current approaches for weakly-supervised learning work directly on images. Hence the accuracy of the predicted 3d pose is affected by the accuracy of the 2d pose learned in intermediate stages of the network \cite{zhou2017towards}. Hence it is difficult to identify whether 3d pose failure on an arbitrary image is due to a noisy estimate of 2d pose or an inaccurate ``lifting" of 2d pose to 3d. In this work we carry out the weakly-supervised learning directly on 2d poses instead of images, to investigate the accuracy of learning 3d poses from ground-truth 2d pose in the absence of ground truth 3d pose.

\vspace{-3pt}
\textbf{In-the-wild 3d pose:}
The widely used 3d pose benchmark datasets Human3.6m \cite{h36m_pami} and HumanEva \cite{sigal2010humaneva} are captured using MoCap systems in controlled lab environments and do not contain sufficient pose variability and scene diversity. On the other hand, datasets such as MPII \cite{andriluka14cvpr} contain large scale in-the-wild data with ground truth annotations for 2d pose obtained from crowdsourcing. This has led to greater generalization of 2d pose estimation methods to the in-the-wild images. 3d pose estimation methods that utilize these 2d pose datasets for weak-supervision can only be assessed in a qualitative manner.  Recently a more challenging 3d pose dataset MPI-INF-3DHP \cite{mono-3dhp2017} has been introduced for more generalized 3d human pose estimation, as it contains some ``in-the-wild" 3d pose training examples. Recent works have used this dataset for supervised learning and cross-dataset evaluation  \cite{luo2018orinet,mehta2017monocular,zhou2017towards,yang20183d,mehta2017vnect,dabral2018learning}. We use this dataset to demonstrate generalization of our method.

\section{Proposed Method}


Our goal is to learn the 3d human pose \textit{$Y_{3d} \in \mathbb{R}^{3 \times J}$} (a set of J body joint locations in 3-dimensional space), given the 2d pose \textit{$Y_{2d} \in \mathbb{R}^{2 \times J}$} in 2-dimensional image coordinates. The 3d pose is learnt in a weakly-supervised manner from a dataset having samples with 2d-3d ground-truth pose pairs (\textit{$Y_{2d},Y_{3d}$}) as well as samples with only 2d pose labels \textit{$Y_{2d}$}. 
For any given training sample, 3d pose prediction \textit{$\hat{Y}_{3d}$} is learnt using supervised learning when ground-truth  \textit{$Y_{3d}$} is present and in an unsupervised manner from input \textit{$Y_{2d}$} when \textit{$Y_{3d}$} is not present. \textit{$Y_{2d}$} is also used as labels for increasing the re-projection accuracy of predicted 3d pose, while ground-truth  \textit{$Y_{3d}$} is present.
The proposed network architecture is illustrated in Figure~\ref{fig:block_diagram}. The network consists of 
(i) 2d-to-3d pose regression module (Section ~\ref{sec:2dto3d}) for predicting 3d pose \textit{$\hat{Y}_{3d}$} from given 2d pose \textit{$Y_{2d}$} and (ii) 3d-to-2d re-projection module (Section ~\ref{sec:3dto2d}) for aligning the 2d re-projection \textit{$\hat{Y}^{reproj}_{2d}$} of predicted 3d pose \textit{$\hat{Y}_{3d}$} with input 2d pose \textit{$Y_{2d}$}. For both the 2d-to-3d and 3d-to-2d network, we adopt similar network architecture as proposed by Martinez et al.\cite{martinez2017simple}.


\subsection{2d-to-3d Pose Regression}
\label{sec:2dto3d}
Our 2d-to-3d pose regression is carried out using the network proposed by Martinez et al.\cite{martinez2017simple}. This method \cite{martinez2017simple} uses a deep feedforward neural network that effectively learns to predict 3d pose from input 2d pose, using a fully supervised 3d loss, which is defined as, 

\begin{equation}
\label{eqn:supervised}
L_{3d} = \frac{1}{N}\sum_{i=1}^{N} (\hat{Y}_{3d} - Y_{3d})^2
\end{equation}

Where, $\hat{Y}_{3d}$ is the predicted 3d pose, $Y_{3d}$ is ground truth 3d pose, and $N$ is the number of training samples.
When our training sample contains 3d ground-truth pose, our network minimizes 3d supervised loss defined in Equation 1.

\subsection{3d Pose Re-projection}
\label{sec:3dto2d}

The predicted 3d pose is valid if it projects back correctly to the input 2d pose. 
The re-projection error is minimized to constrain the predicted 3d pose. This re-projection loss ($L_{2d}$) is defined as:

\begin{equation}
L_{2d} = \frac{1}{N}\sum_{i=1}^{N} (\hat{Y}^{reproj}_{2d} - Y_{2d})^2
\end{equation}

Where, $\hat{Y}^{reproj}_{2d}$ is re-projection of predicted 3d pose $\hat{Y}_{3d}$ and $Y_{2d}$ is input 2d pose.

An infinite number of 3d poses can be re-projected to a single 2d pose, but all of them may not be a physically plausible human pose. Hence, the solution space is further restricted to ensure plausibility in predicted 3d poses by introducing structural constraints on the bone lengths.

\subsection{Geometric Constraints on 3d Human Poses}
\subsubsection*{Bone length symmetry loss}


To ensure symmetry between contra-lateral segments of the human pose, bone length symmetry loss  ($L^{symm}_{3d}$) has been applied on predicted limb lengths. Bone lengths of leg, arm, between neck to shoulder and hip to pelvis, remain same for left and right segments of the body. This constraint has been enforced on predicted 3d pose, using symmetry loss $L^{symm}_{3d}$, which is defined as:

\begin{equation}
L^{symm}_{3d} = \frac{1}{N}\sum_{i=1}^{N} \Big(\frac{1}{|Rs_i|}\sum_{e \in Rs_i}(B_e^l - B_e^r)^2\Big)
\end{equation}

Where, $Rs_i$ represents a set of skeleton segments \\ $\{arm, \, leg, \, neck\_shoulder, \, hip\_pelvis\}$. $B_e^l$ and $B_e^r$ are bone lengths of left and right side of each of the segments $e$.

\vspace{1em}
\noindent
The total loss ($L$) minimized by our full network is defined as:
\begin{equation}
L = \alpha L_{3d} + \beta L_{2d} + \gamma L^{symm}_{3d}
\end{equation}

Here $\alpha$, $\beta$, $\gamma \in (0,1)$ are scalar values denoting the weightage of each loss term in total loss. In the absence of 3d pose ground truth, $\alpha$ is set to 0.

\subsection{Network architecture}
Figure~\ref{fig:block_diagram} shows the overall architecture for our proposed network. Both the 2d-to-3d and 3d-to-2d network modules use the architecture similar to Martinez et al. \cite{martinez2017simple}. Each module consists of two residual blocks, where each residual block contains two linear layers, with batch normalization, dropout and ReLu activation after each layer. A residual connection is added from the initial to the final layer of each block. Also, the input and output of each module are connected to two fully connected layers that map the input dimension to the dimension of intermediate layers and back to output dimension respectively.  
  
\section{Experimental Setup}

\begin{table*}[htbp]
	\centering
	\scriptsize
	\resizebox{\textwidth}{!}{
	\begin{tabular}{c c c c c c c c c}
		\hline
		\textbf{Method} &\textbf{Direct.}& \textbf{Discuss} & \textbf{Eating} & \textbf{Greet} & \textbf{Phone} & \textbf{Photo} & \textbf{Pose} & \textbf{Purch.}\\
		\hline
		\hline
		
		Zhou et al. \cite{zhou2017towards} &54.8 &60.7 & 58.2 &71.4 &62.0 &53.8 &55.6 &75.2\\
		Dabral et al. \cite{dabral2018learning} &44.8 &50.4 &44.7&49.0&52.9 &61.4 &43.5 &45.5 \\
		Yang et al. \cite{yang20183d}&51.5&58.9&50.4&57.0&62.1&65.4&49.8&52.7\\
		Luo et al. \cite{luo2018orinet}&49.2& 57.5& 53.9 &55.4 &62.2 &73.9 &52.1 &60.9 \\
		Sun et al. \cite{sun2017compositional}&42.1&44.3&45.0&45.4&51.5&\textbf{43.2}&\textbf{41.3} &59.3\\
		Martinez et al. (GT) \cite{martinez2017simple}&37.7&44.4& 40.3 & 42.1 & 48.2 & 54.9 & 44.4 & 42.1 \\
		\hline
		Ours (GT)  &\textbf{35.74}&\textbf{42.39}&\textbf{39.06}&\textbf{40.55}& \textbf{44.37}&52.54 &42.86 & \textbf{38.83} \\
		\hline
		\hline
		\textbf{Method} & \textbf{Sitting} & \textbf{SittingD} & \textbf{Smoke} & \textbf{Wait} & \textbf{WalkD} & \textbf{Walk} & \textbf{WalkT} & \textbf{Avg.} \\
		\hline

		Zhou et al. \cite{zhou2017towards} &111.6 &64.1& 65.5 &66.0 &51.4&63.2&55.3& 64.9\\
		Dabral et al. \cite{dabral2018learning} &63.1 &87.3 & 51.7 &48.5&52.2 &37.6 &41.9 &52.1 \\
		Yang et al. \cite{yang20183d}&69.2&85.2&57.4&58.4&43.6&60.1&47.7&58.6\\
		Luo et al. \cite{luo2018orinet}&73.8 &96.5 &60.4 &55.6 &69.5 &46.6 &52.4 &61.3 \\
		Sun et al. \cite{sun2017compositional}& 73.3 &\textbf{51.0} &53.0 &44.0 &\textbf{38.3} &48.0 &44.8 &48.3\\
		Martinez et al. (GT)\cite{martinez2017simple}&54.6 & 58.0 &45.1 &46.4 &47.6 &36.4 & 40.4& 45.5 \\
		\hline
		Ours (GT)  &\textbf{53.08} & 53.90 & \textbf{42.10}&\textbf{43.36}& 43.92&  \textbf{33.31}& \textbf{36.54} & \textbf{42.84} \\
		\hline

	\end{tabular}
	}
	\caption{MPJE (Mean Per Joint Error, mm) metric on Human3.6m dataset under defined protocol i.e. no rigid alignment of predicted 3d pose with ground truth 3d pose. GT denotes training on ground-truth 2d pose labels. Except Martinez et al. all other state-of-the-art methods use images for training instead of 2d pose labels.
Our model achieves least MPJE for majority of the actions.}
	\label{table1}
\end{table*}


\subsection{Dataset Description}

\textbf{Human3.6m} \cite{ionescu2014human3} is the largest publicly available 3d human pose benchmark dataset, with ground truth annotations captured with four RGB cameras and motion capture (MoCap) system. The dataset consists of 3.6 million images featuring 11 professional actors (only 7 used in the experimental setup) performing 15 everyday activities such as walking, eating, sitting, discussing, taking photo, etc. This dataset has both 2d and 3d joint locations along with camera parameters and body proportions for all the actors. Each pose has annotations for 32 joints; however, only the major 17 joints are used in the experimental setup of most of the state-of-the-art methods \cite{martinez2017simple,zhou2017towards}. We evaluate the performance of our proposed method using standard protocol \cite{li20143d} of evaluating Human3.6m, which uses actors 1, 5, 6, 7 and 8 for training and actors 9 and 11 for testing.  

\textbf{MPII} \cite{andriluka14cvpr} is the benchmark dataset for 2d human pose estimation. The images were collected from short YouTube videos covering daily human activities with complex poses and variant image appearances. Poses are annotated by human with sixteen 2d joints. It contains around 25k training images and 2957 validation images. Since the dataset has been collected randomly (not in controlled lab setup), it consists of a large variety of poses. Hence, 3d pose estimation methods can use this data for better generalization to in-the-wild human poses.

\textbf{MPI-INF-3DHP} \cite{mehta2017monocular} is a newly released 3d human pose dataset of 6 subjects performing 7 actions in indoor settings (background with a green screen (GS) and no green screen(NoGS)), captured with MoCap System and 14 RGB cameras and 2 subjects performing actions in outdoor in-the-wild settings. This makes it more challenging dataset than Human3.6M, which has data captured only in indoor settings. We use MPI-INF-3DHP dataset to test the generalization ability of our proposed model to in-the-wild 3d poses. The testing split consists of 2935 valid frames.

\subsection{Data Pre-processing}

While no augmentation is done for 2d poses and 3d poses of Human3.6m and MPI-INF-3DHP datasets, MPII is augmented for 35 times (rotation and scaling of 2d poses) for training, 
to make it compatible with the size of Human3.6m dataset.
Like previous work \cite{martinez2017simple}, we apply standard normalization (zero mean unit standard deviation) on 2d and 3d poses. We use root-centered 3d poses (skeleton with origin at pelvis joint) for 2d-to-3d regression like many other state-of-the-art methods. We also use root-centering on 2d pose labels for the 3d-to-2d re-projection module. 

\subsection{Training Strategy}

Our network is trained in three consecutive phases. In the first phase, only the 3d-to-2d regression module is trained with full supervision on 3d pose ground truth. In the second phase, the 2d-to-3d re-projection module is trained with the ground truth 3d pose input to predict re-projected 2d pose. In the third phase, both pre-trained 3d-to-2d and 2d-to-3d modules are fine-tuned simultaneously. During this final phase, the 2d re-projection module is fine-tuned using predicted 3d pose instead of ground truth 3d pose. To understand the generalization ability of our proposed network, three variants of models have been trained.
\begin{itemize}

\item \textit{Model I}, Trained on Human3.6m dataset.

\item \textit{Model II}, Model I fine tuned on Human3.6m and MPII dataset.

\item \textit{Model III}, Model I fine tuned on MPI-INF-3DHP dataset.

\end{itemize}

Except for Model II, all other models are trained with the 3d supervised loss, 2d re-projection loss and bone-length symmetry loss. In Model II, for MPII dataset only unsupervised losses i.e. 2d re-projection loss and bone-length symmetry loss have been used. 

\subsection{Implementation details}
\label{Implementation details}

2d-to-3d regression and 3d-to-2d re-projection modules of Model I and III (MPI-INF-3DHP fine tuning) are individually pre-trained on ground truth pose for 50 epochs. After pre-training, these modules are trained simultaneously for another 100 epochs with predicted 3d pose as input to 3d-to-2d re-projection module.  For Model II (MPII fine tuning), 
both modules are fine-tuned simultaneously for 200 epochs using training samples from both Human 3.6M and MPII (1:1 ratio in a batch). For training samples from MPII, the value of $\alpha $ is set to 0 since MPII does not contain ground-truth 3d pose for supervision. In all other cases, the value of $\alpha $, $\beta$ and $\gamma$ are empirically set to 0.5, 0.5 and 1.0 respectively during end-to-end training of full network. Learning rate is 1e-4, and batch size is 64, during the training of all models.

\begin{table*}[htbp]
	\centering
	\scriptsize
	\scalebox{0.8}{
	\resizebox{\textwidth}{!}{
	\renewcommand\arraystretch{0.75}
	\begin{tabular}{|c|c|c|c|c|c|c|}
		\hline
	\multirow{2}{*}{\textbf{Method}} &\multirow{2}{*}{\textbf{Training Data}} & \multicolumn{4}{|c|}{\textbf{PCK}}  & \textbf{AUC}\\
	\cline{3-7}
	&&GS&	NoGS&	Outdoor&	ALL&	ALL\\
	
	\hline
	Zhou et al. \cite{zhou2017towards}&	H36m&	45.6&	45.1&	14.4&	37.7&	20.9\\	
	\hline
    Martinez et al.\cite{martinez2017simple}&	H36m&62.8*&	58.5*&	62.2*&	62.2*&27.7* \\
    \hline
	Mehta et al.\cite{mehta2017monocular} &	H36m &	70.8&	62.3&	58.5&	64.7&	31.7\\
	
    \hline
	Luo et al. \cite{luo2018orinet}&	H36m&	71.3*&	59.4*&	65.7*&	65.6*&	33.2*\\
	\hline 
	Yang et al.\cite{yang20183d}& H36M+MPII &-&-&-&69.0&32.0\\
    \hline
	Zhou et al.\cite{zhou2017towards}&	H36m+MPII&	71.1&	64.7&	\textbf{72.7}&	69.2&	32.5\\
    \hline
    Ours (Model I) &H36m&	66.9*&	63.0*&	67.4*&	65.8*&	31.2*\\
    \hline
    Ours (Model II) &	H36m+MPII&	\textbf{74.2}*&	\textbf{66.9}*&	71.4*&	\textbf{70.8}*&	\textbf{34.5}*\\
    \hline

	\end{tabular}
	}
	}
	\caption{Results on MPI-INF-3DHP test-set by scene. Higher PCK(\%) and AUC indicates better performance. $-$ means values are not given in original paper. $*$ denotes re-targeting of predicted 3d pose using ground truth limb length. Our model shows best performance among state-of-the-art methods while fine tuned on MPII dataset.}
	\label{table2}
\end{table*}

\begin{table*}[htbp]
	\centering
	\scriptsize
	\resizebox{\textwidth}{!}{
	\renewcommand\arraystretch{1.5}
	\begin{tabular}{|c|c|c|c|c|c|c|c|c|c|c|c|}
		\hline
	\multirow{2}{*}{\textbf{Method}} &\multirow{2}{*}{\textbf{Training Data}} &  \textbf{Walk}	&\textbf{Exer.}	&\textbf{Sit}&	\textbf{Reach}&	\textbf{Floor} & \textbf{Sport} & \textbf{Misc.}&\multicolumn{3}{|c|}{\textbf{Total}}\\
 	\cline{3-12}
  & &PCK&	PCK&	PCK&	PCK&	PCK	&PCK&	PCK&	PCK&	AUC&	MPJE\\
	\hline
	\multirow{1}{*}{Mehta et al. \cite{mehta2017monocular}}&\multirow{1}{*}{(MPII+LSP)H3.6M+3DHP$a$}&86.6&75.3&74.8&73.7&52.2&82.1&77.5&75.7&39.3&117.6\\
	\hline
	Mehta et al.\cite{mehta2017vnect}&	(MPII+LSP)H3.6M+3DHP$a$&	87.7&77.4&74.7&72.9&51.3&83.3&80.1&76.6&40.4&124.7
\\	
\hline
Dabral et al.\cite{dabral2018learning}& H3.6M+3DHP&-&-&-&-&-&-&-&76.7&39.1&103.8
\\
	\hline
Luo et al.\cite{luo2018orinet}&	(MPII)H3.6M+3DHP&90.5*&80.9*&90.0*&85.6*&70.2*&93.0*&92.9*&83.8*&47.7*&85.0*
\\
	\hline
    Ours&	H3.6M+3DHP&	\textbf{97.3*}&\textbf{93.0*}&\textbf{92.3*}&\textbf{95.3*}&\textbf{86.4*}&\textbf{94.6*}&\textbf{94.3*}&\textbf{85.4*}&\textbf{55.8*}&\textbf{71.40*}
\\
    \hline  

	\end{tabular}
	}
	\caption{Activity-wise performance on MPI-INF-3DHP test-set using standard metrics PCK (\%), AUC and MPJE (mm). (MPII) means pretrained on MPII dataset. $a$ denotes background augmentation in training data. $-$ means values are not given in original paper. $*$ denotes the re-targeting of predicted 3d pose using ground truth limb length. Higher PCK, AUC and lower MPJE indicates better performance. We have achieved significantly better performance than the state-of-the-art methods on all the actions in terms of all the metrics.}
	\label{table3}
\end{table*}

\begin{figure*}[t!]
        \centering
        \begin{tabular}{cccccc}
            \includegraphics[keepaspectratio=true, scale = 0.18]{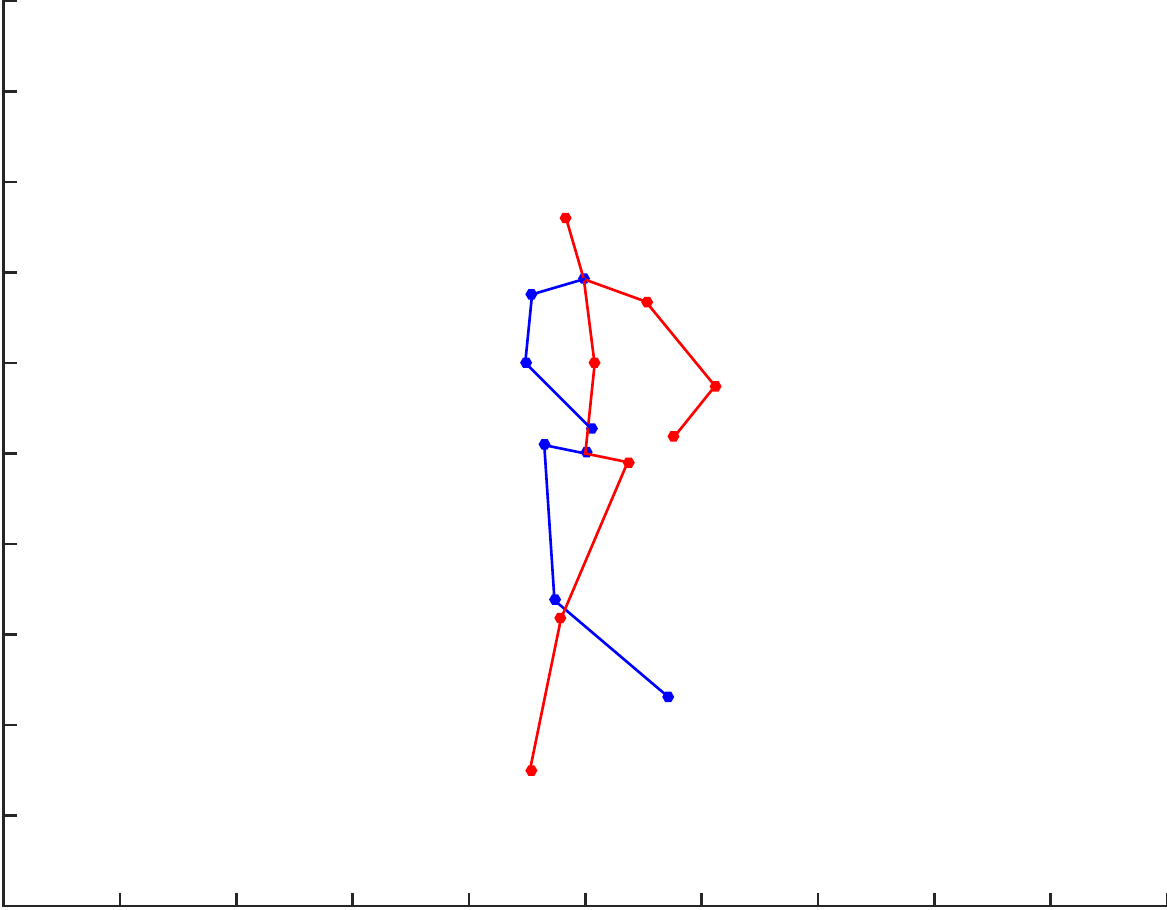} & \includegraphics[keepaspectratio=true, scale = 0.17]{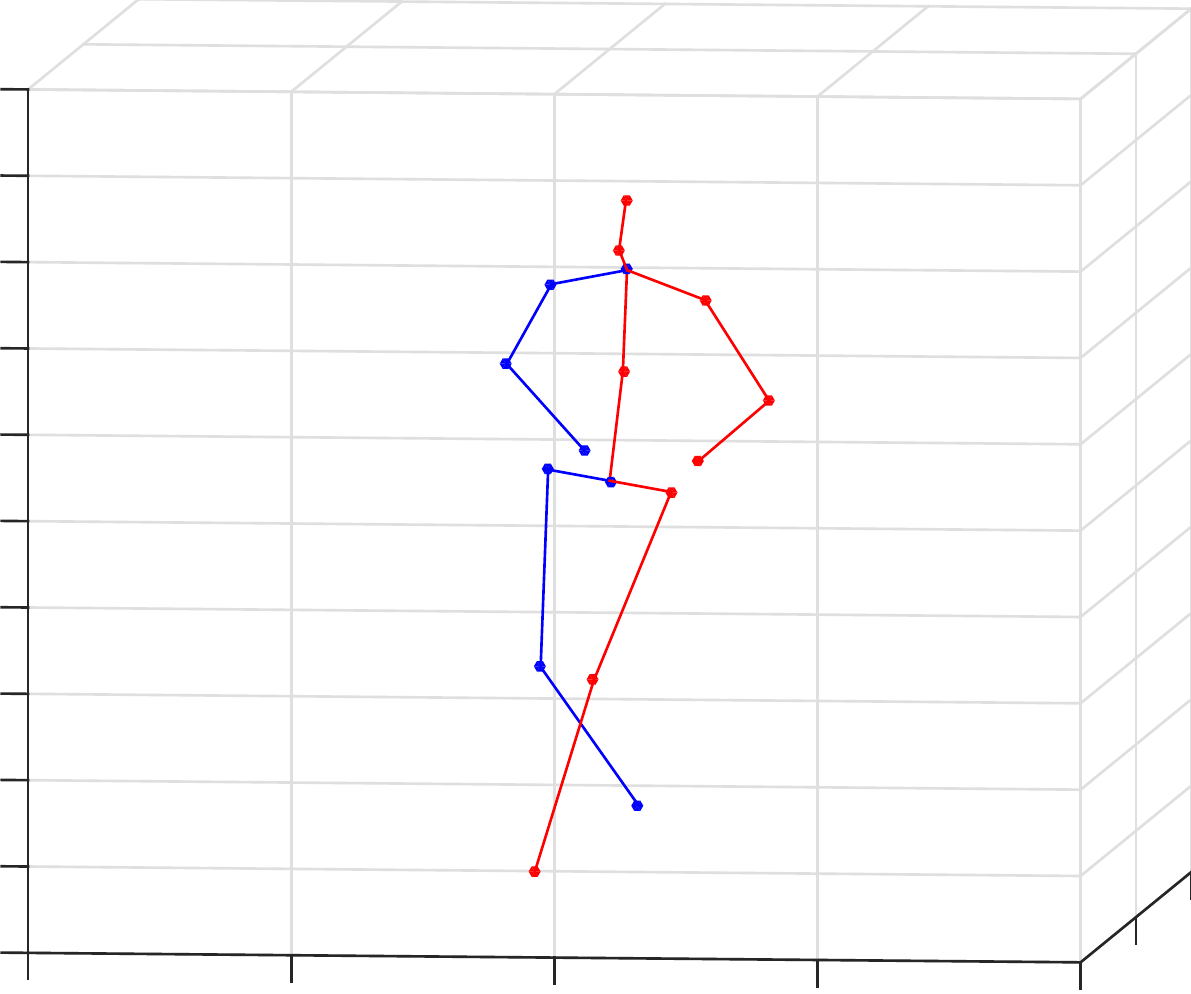} & \includegraphics[keepaspectratio=true, scale = 0.18]{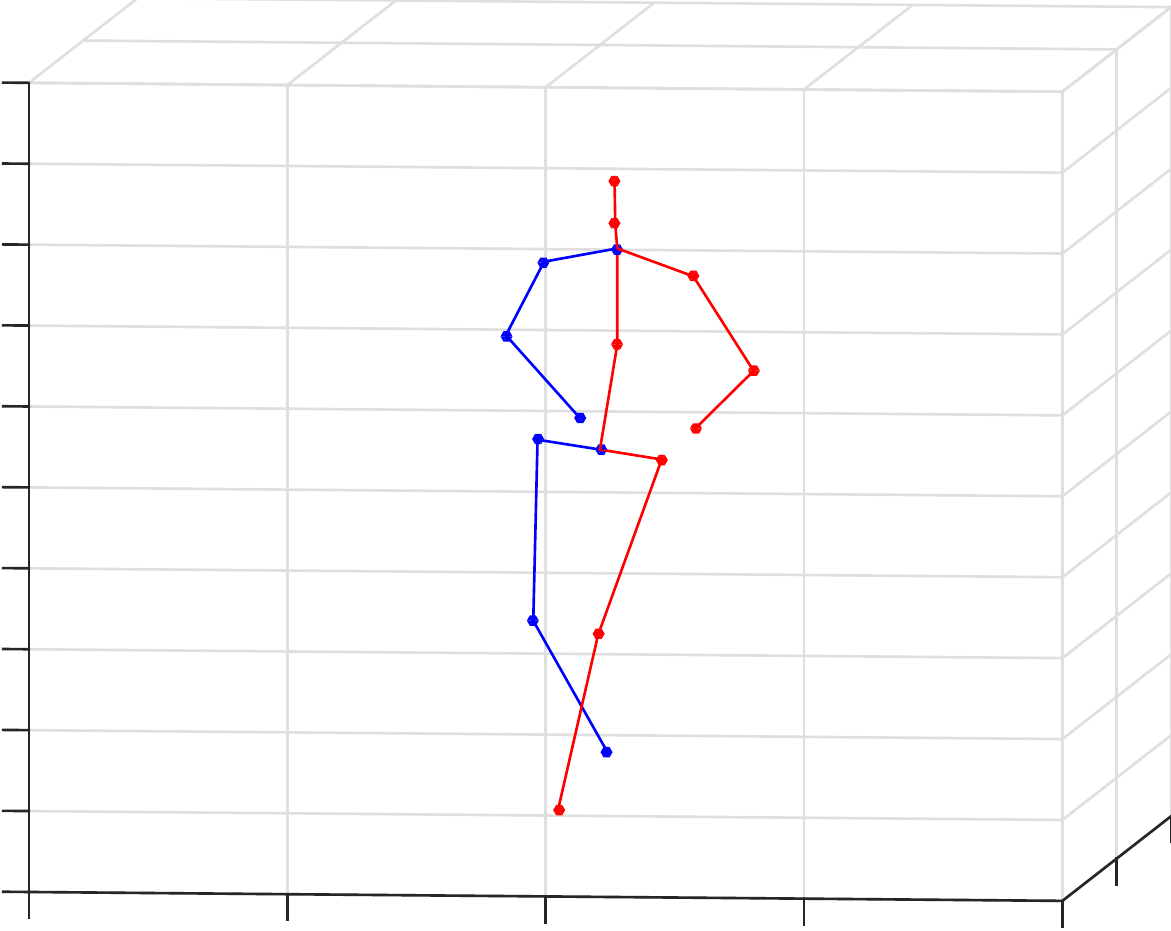} & \includegraphics[keepaspectratio=true, scale = 0.18]{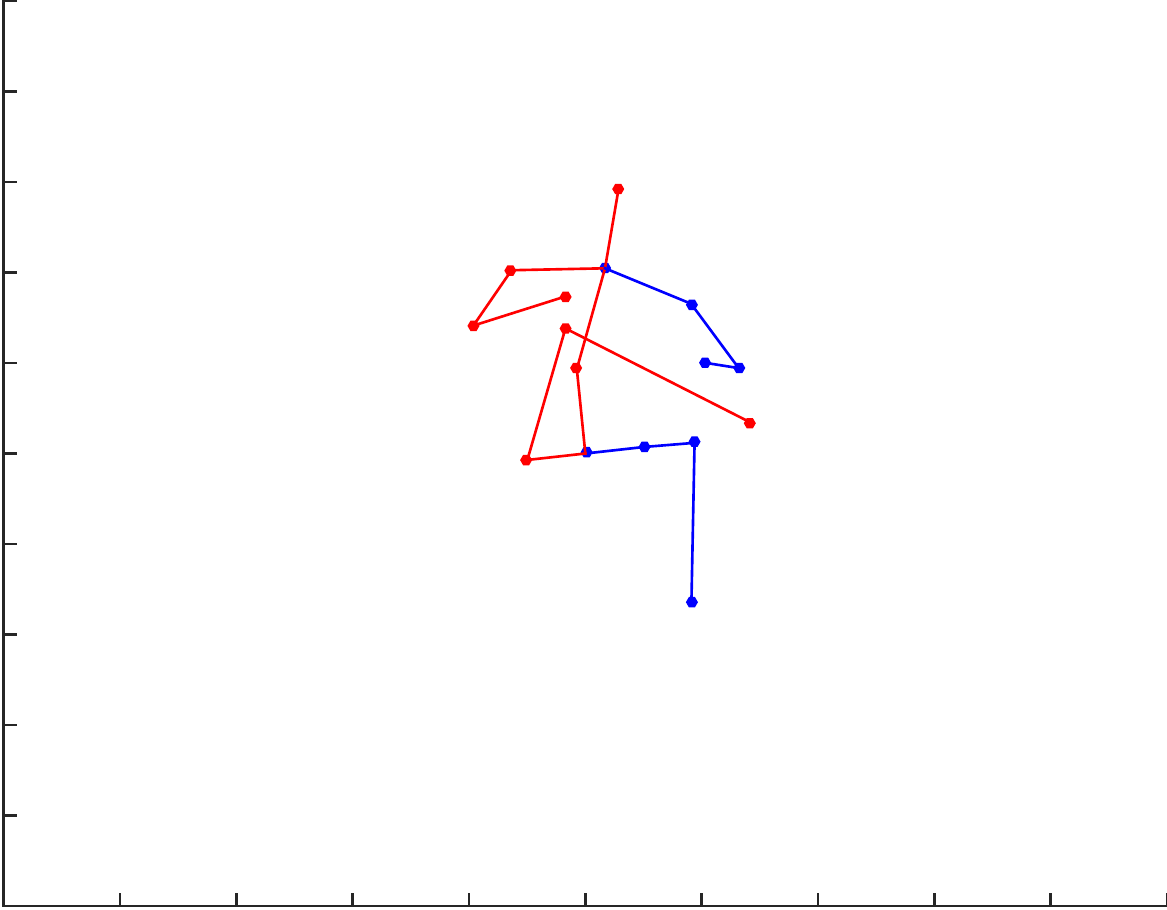} & \includegraphics[keepaspectratio=true, scale = 0.177]{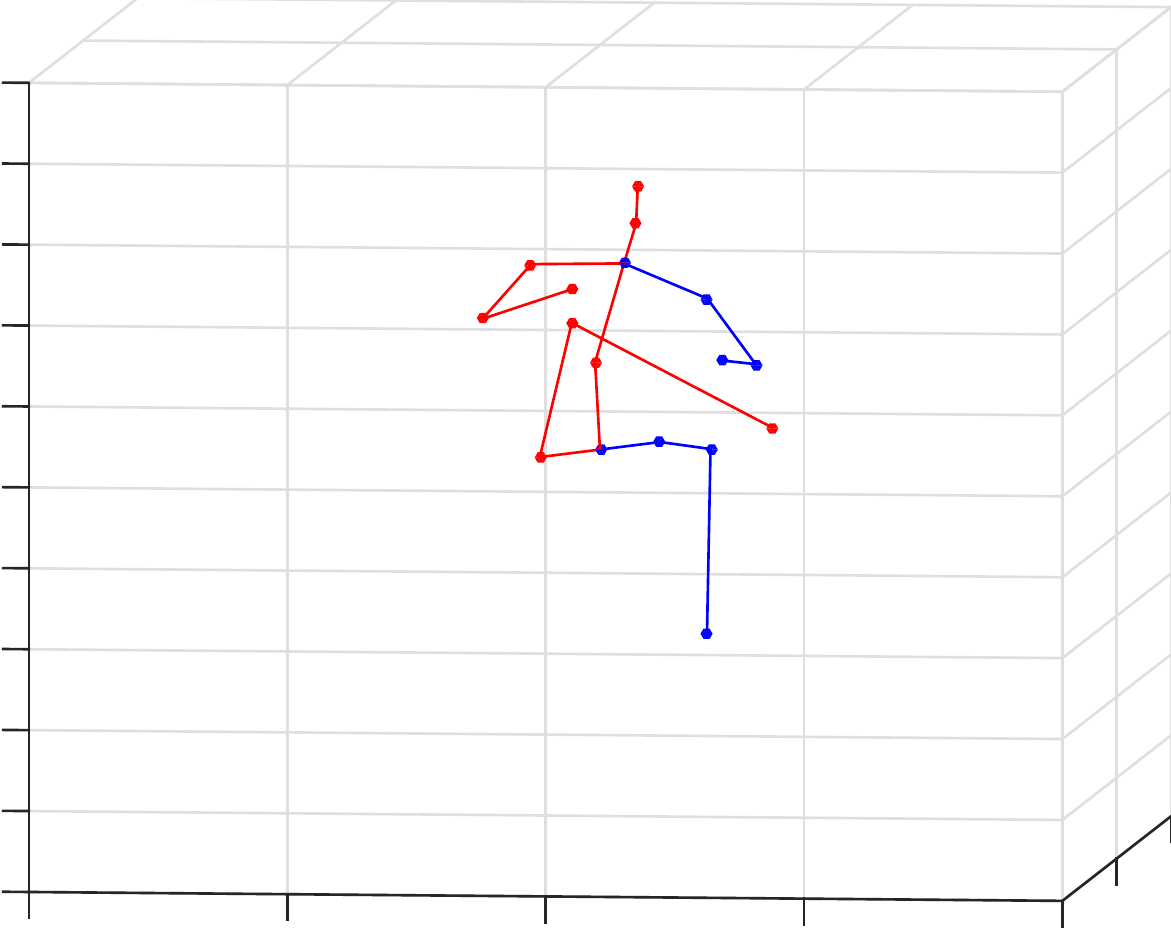} & \includegraphics[keepaspectratio=true, scale = 0.175]{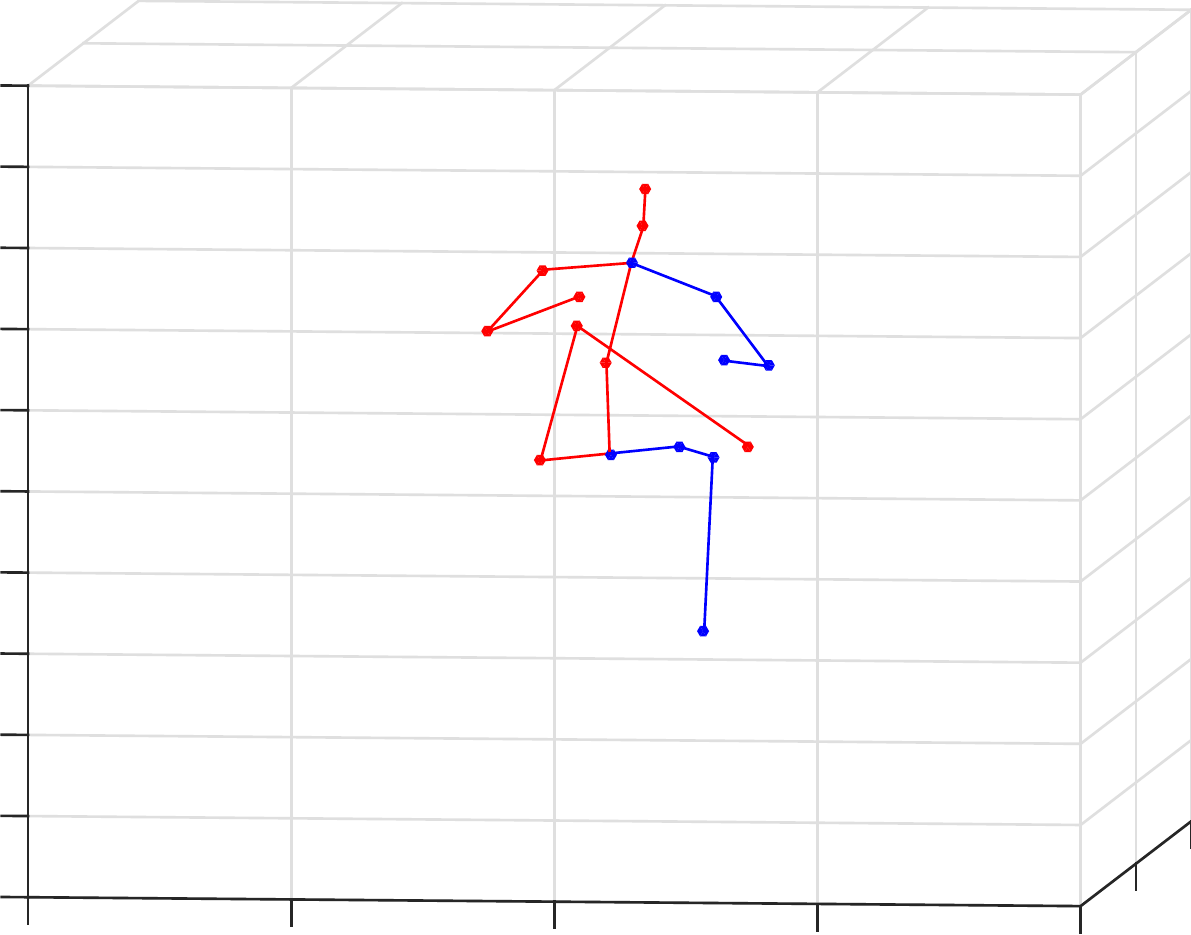}\\            
            \includegraphics[keepaspectratio=true, scale = 0.18]{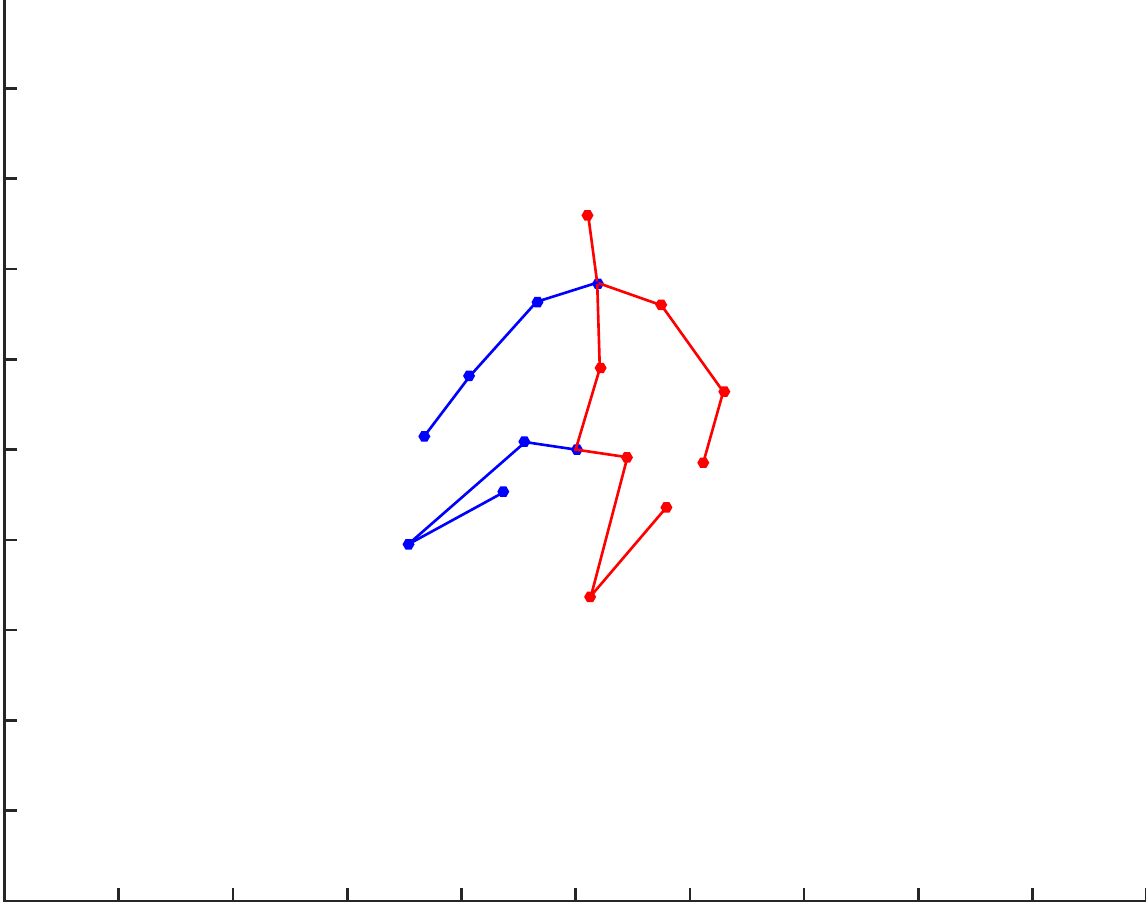} & \includegraphics[keepaspectratio=true, scale = 0.175]{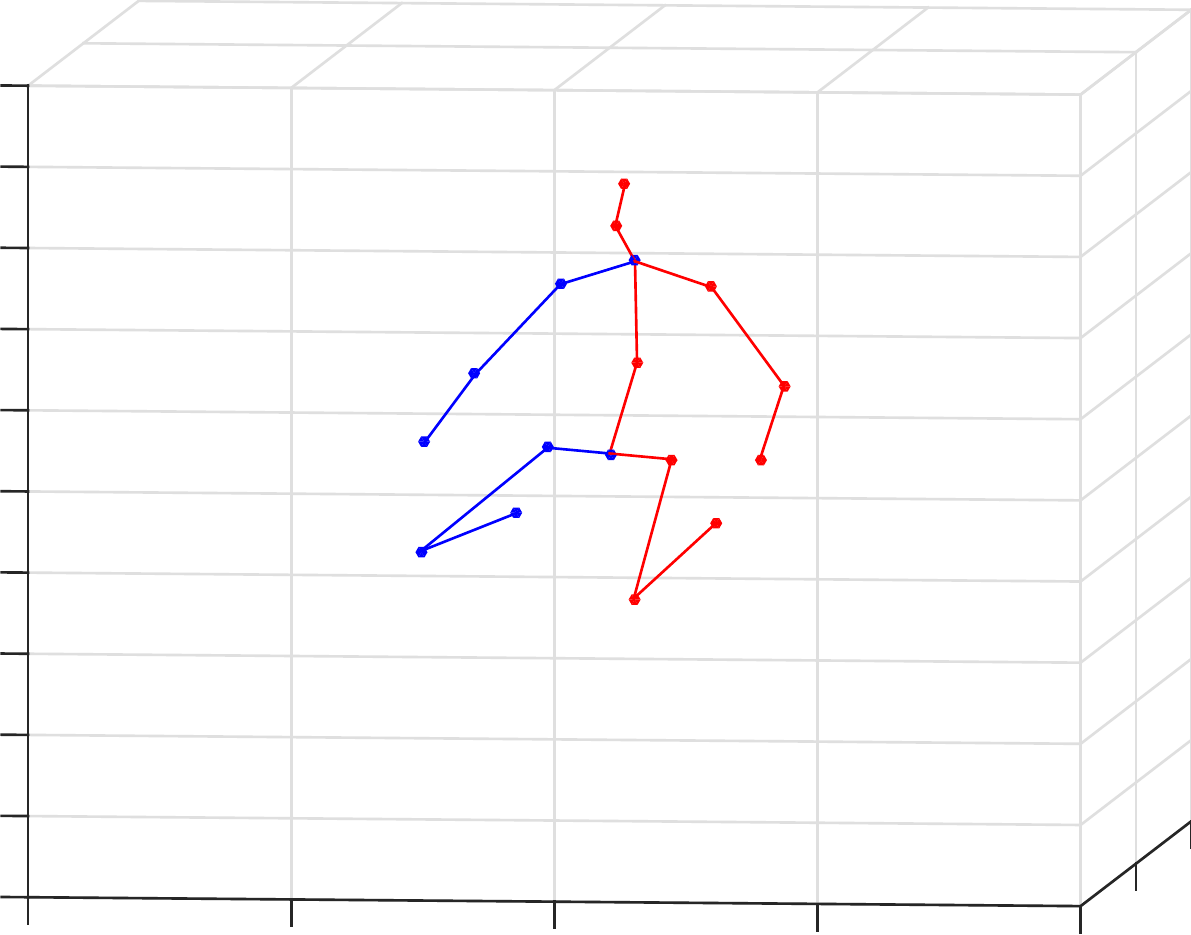} & \includegraphics[keepaspectratio=true, scale = 0.177]{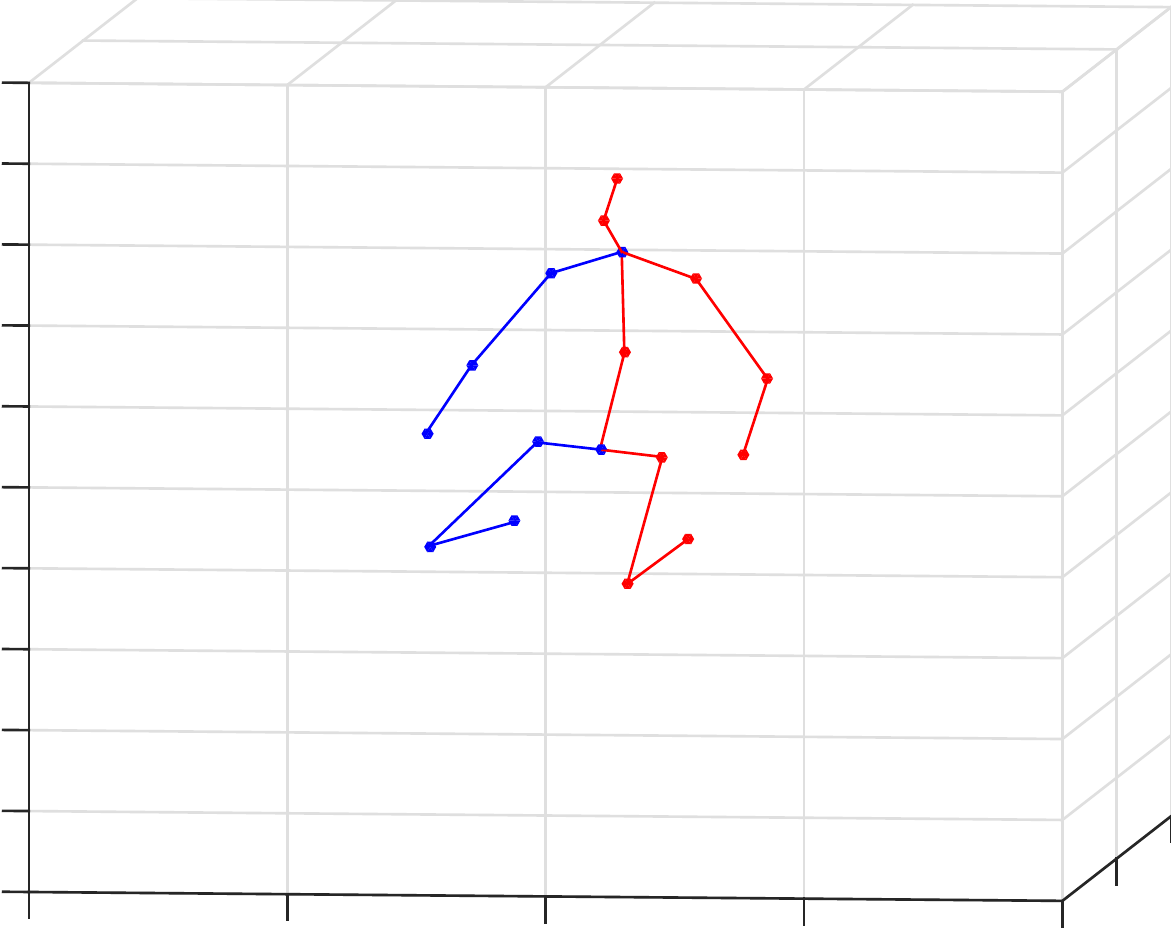} & \includegraphics[keepaspectratio=true, scale = 0.18]{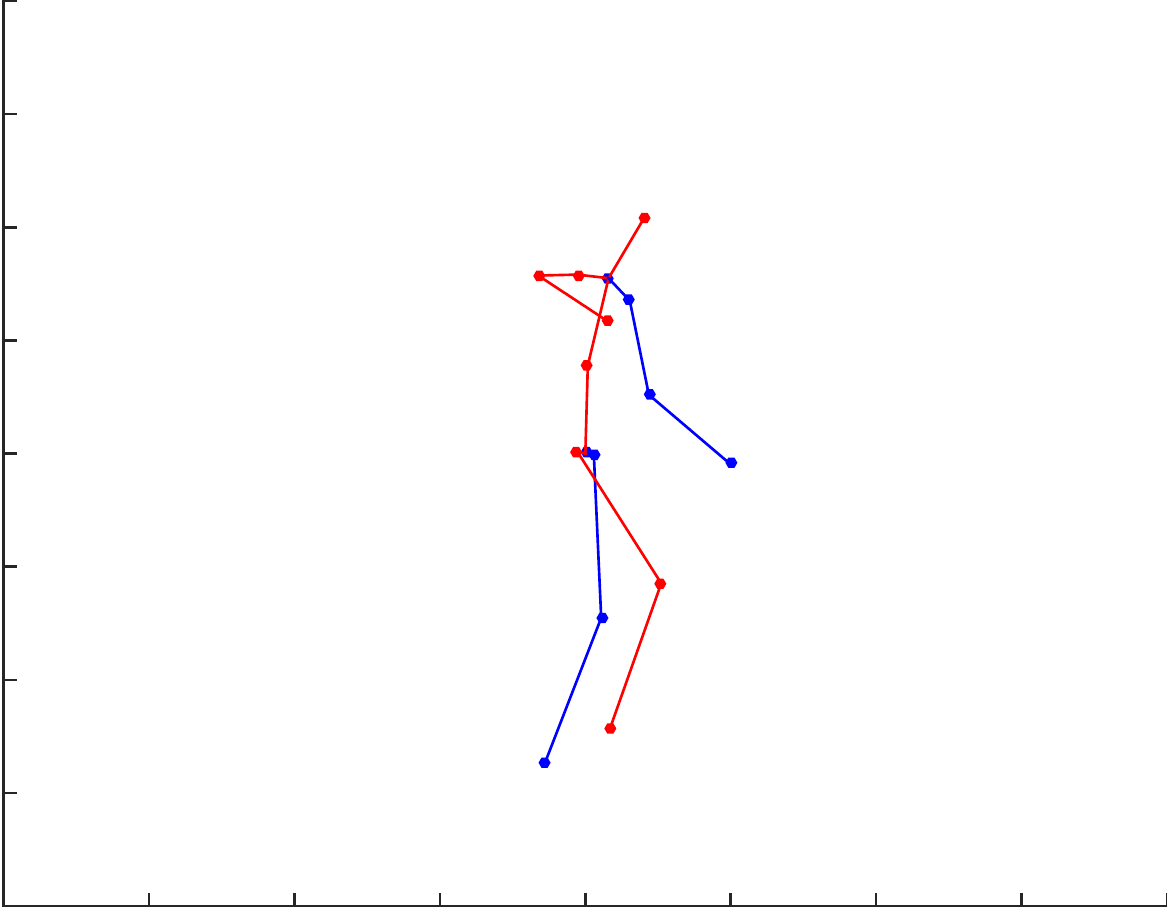} & \includegraphics[keepaspectratio=true, scale = 0.177]{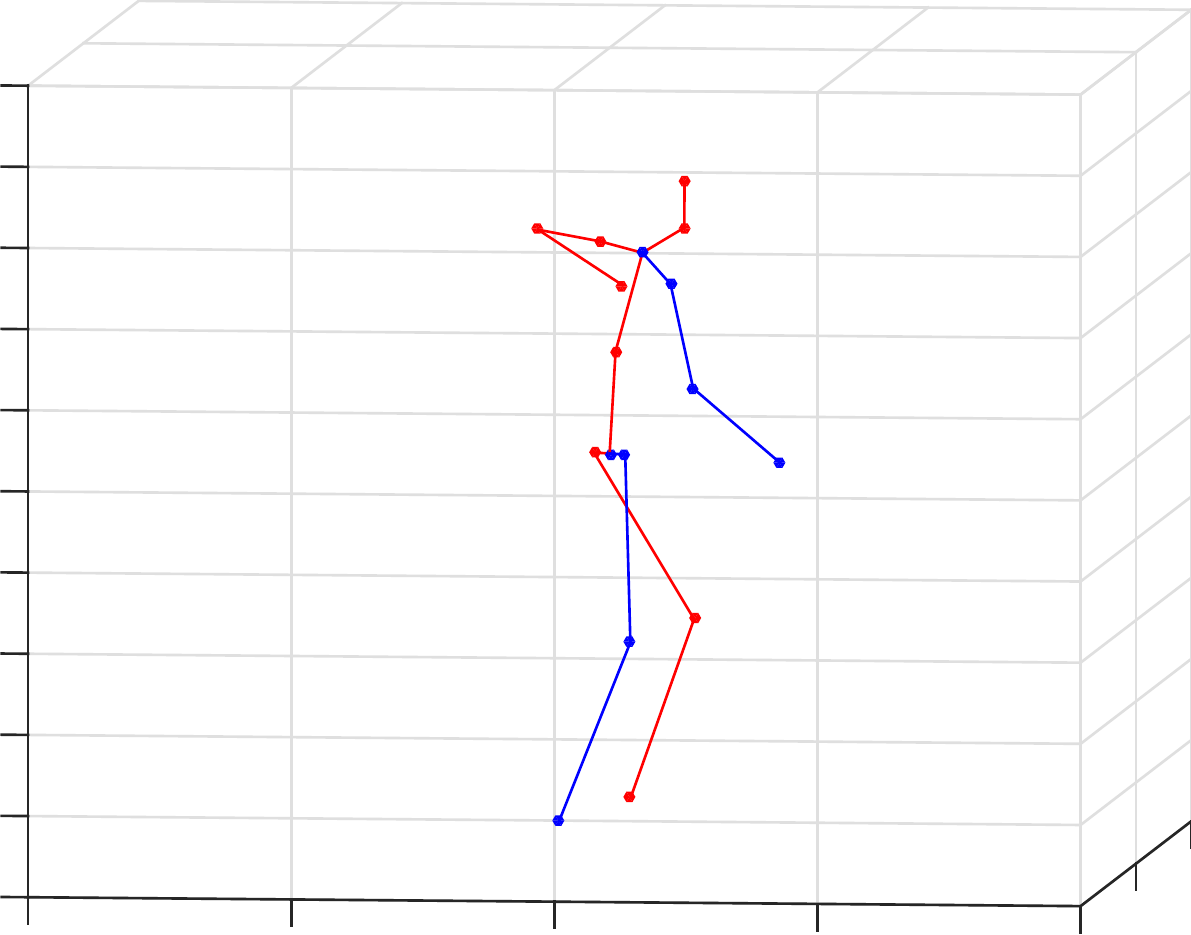} & \includegraphics[keepaspectratio=true, scale = 0.175]{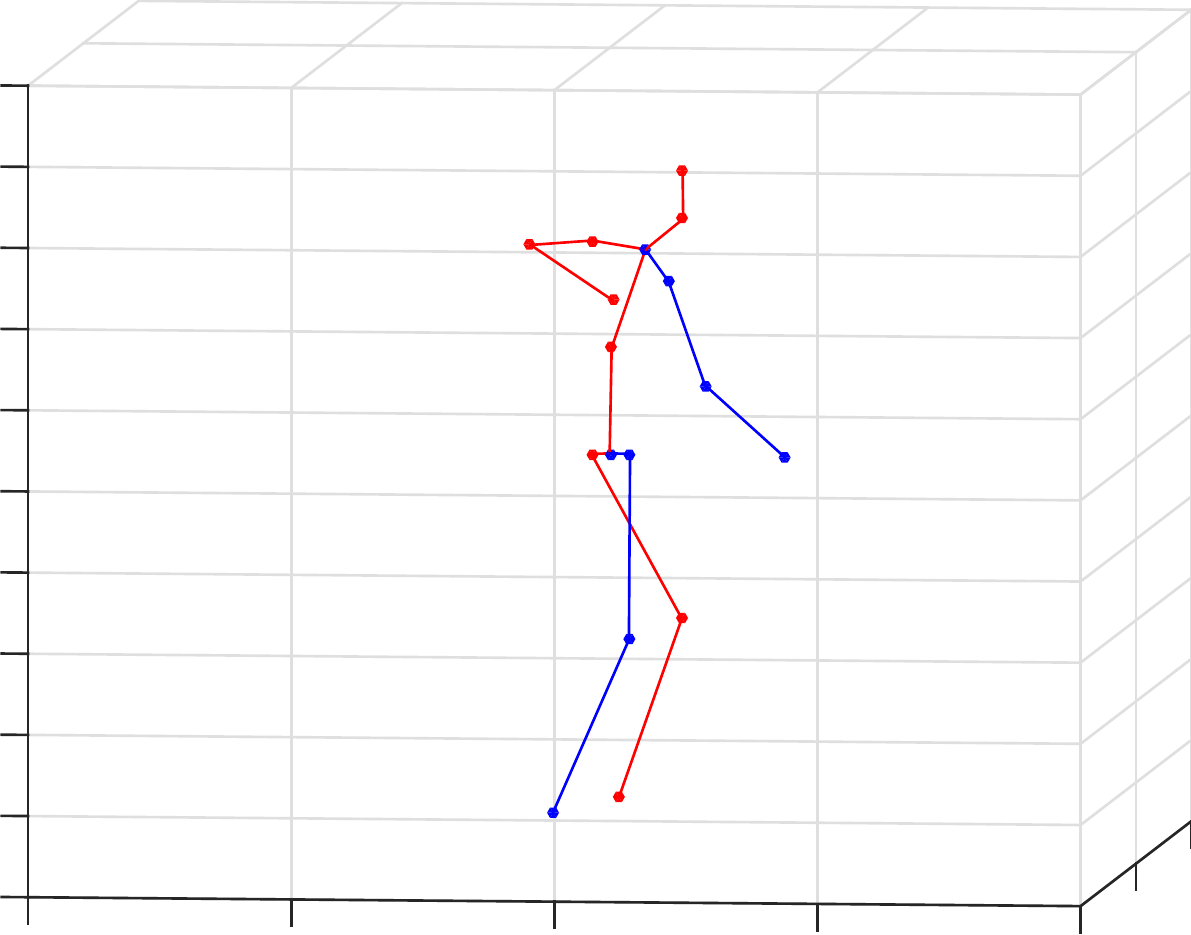}\\            
            \includegraphics[keepaspectratio=true, scale = 0.18]{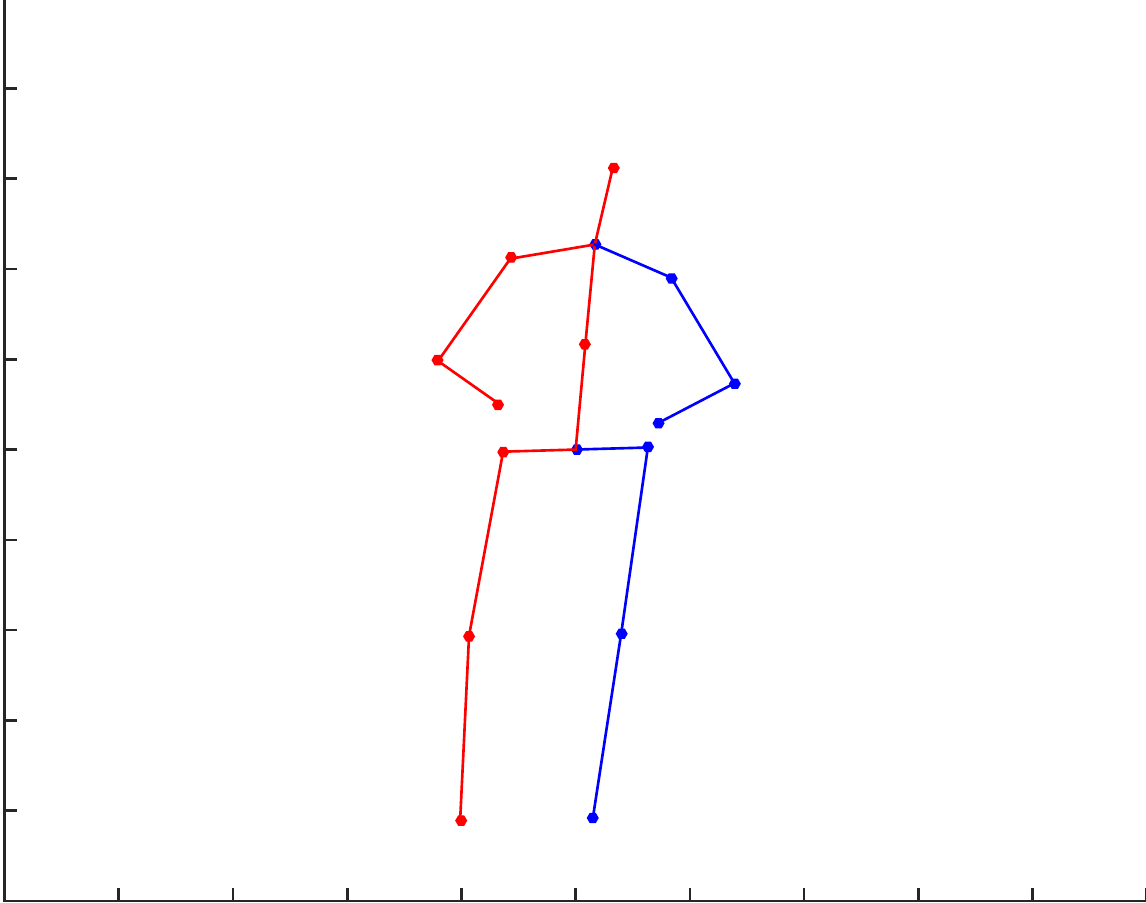} & \includegraphics[keepaspectratio=true, scale = 0.18]{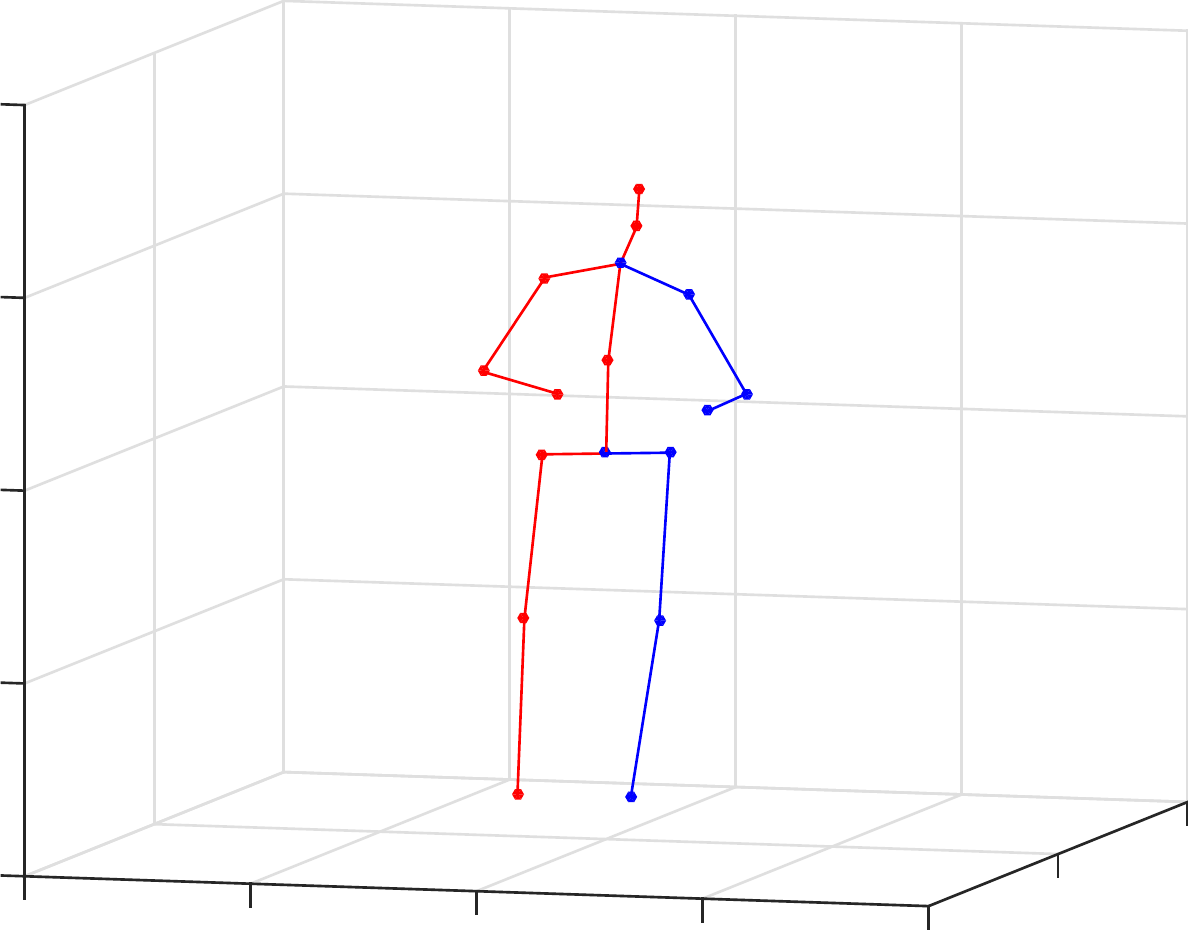} & \includegraphics[keepaspectratio=true, scale = 0.177]{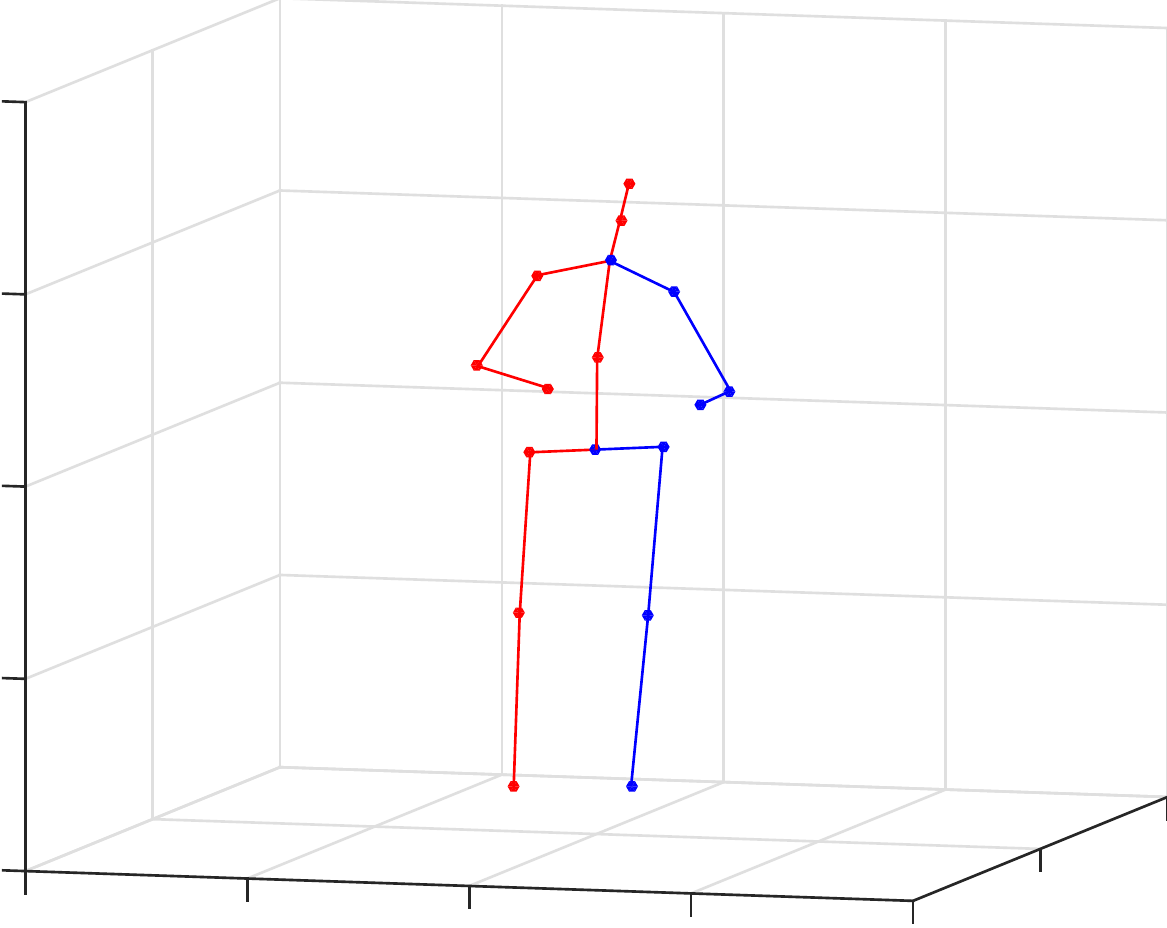} & \includegraphics[keepaspectratio=true, scale = 0.18]{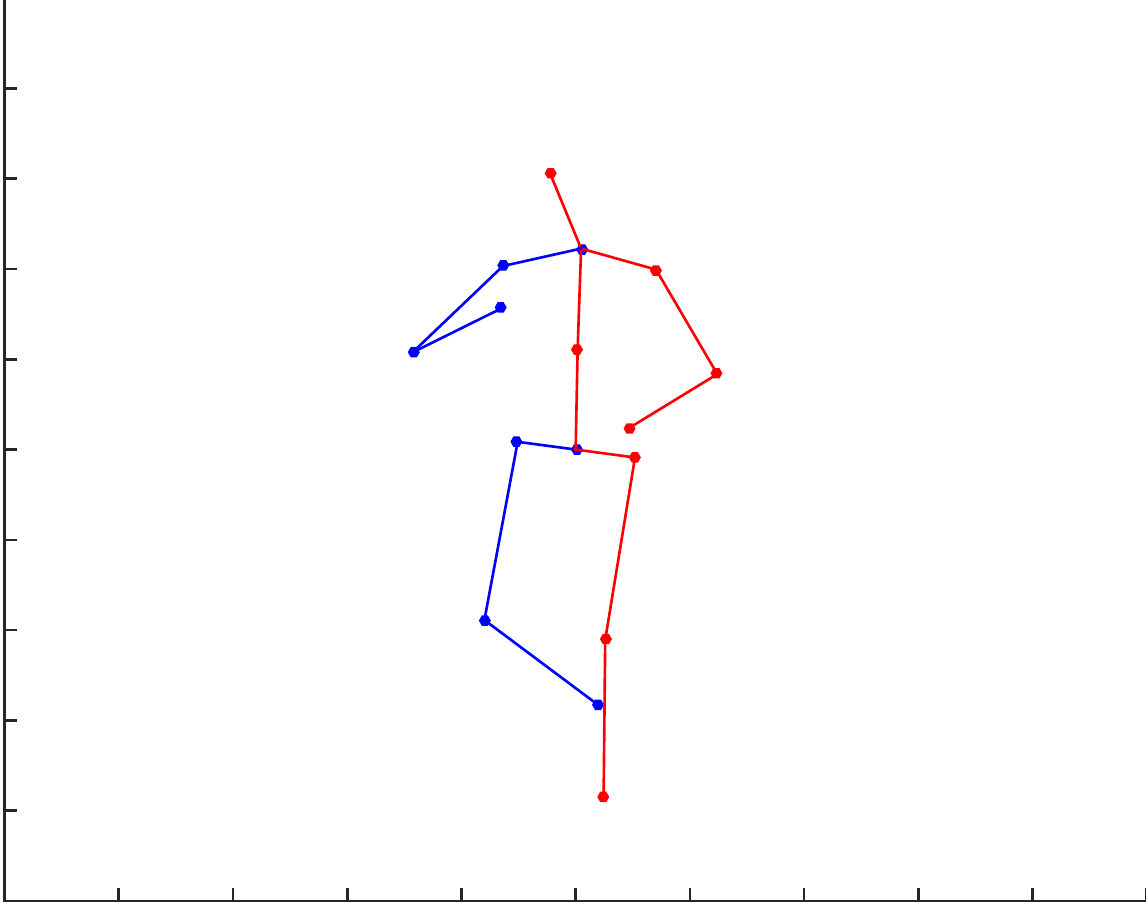} & \includegraphics[keepaspectratio=true, scale = 0.177]{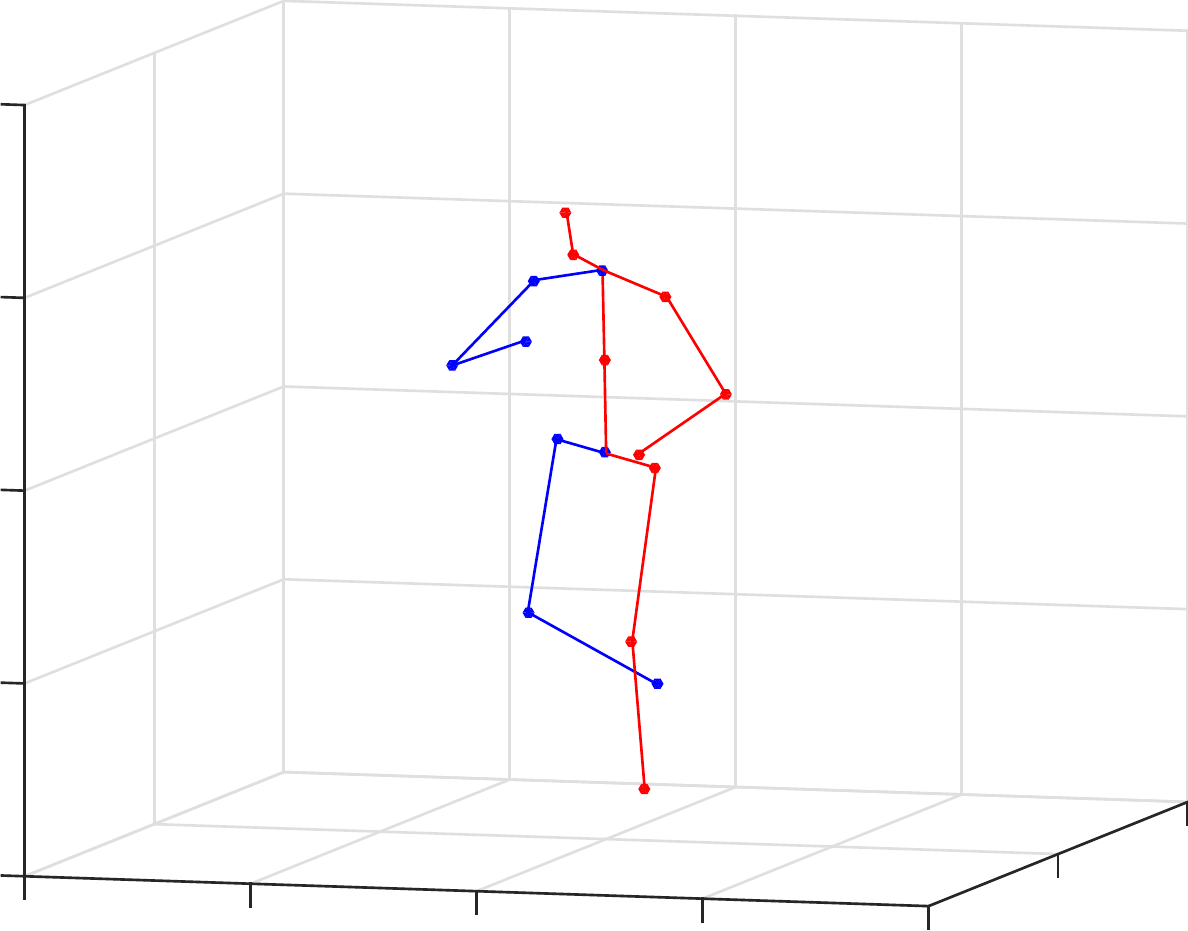} & \includegraphics[keepaspectratio=true, scale = 0.175]{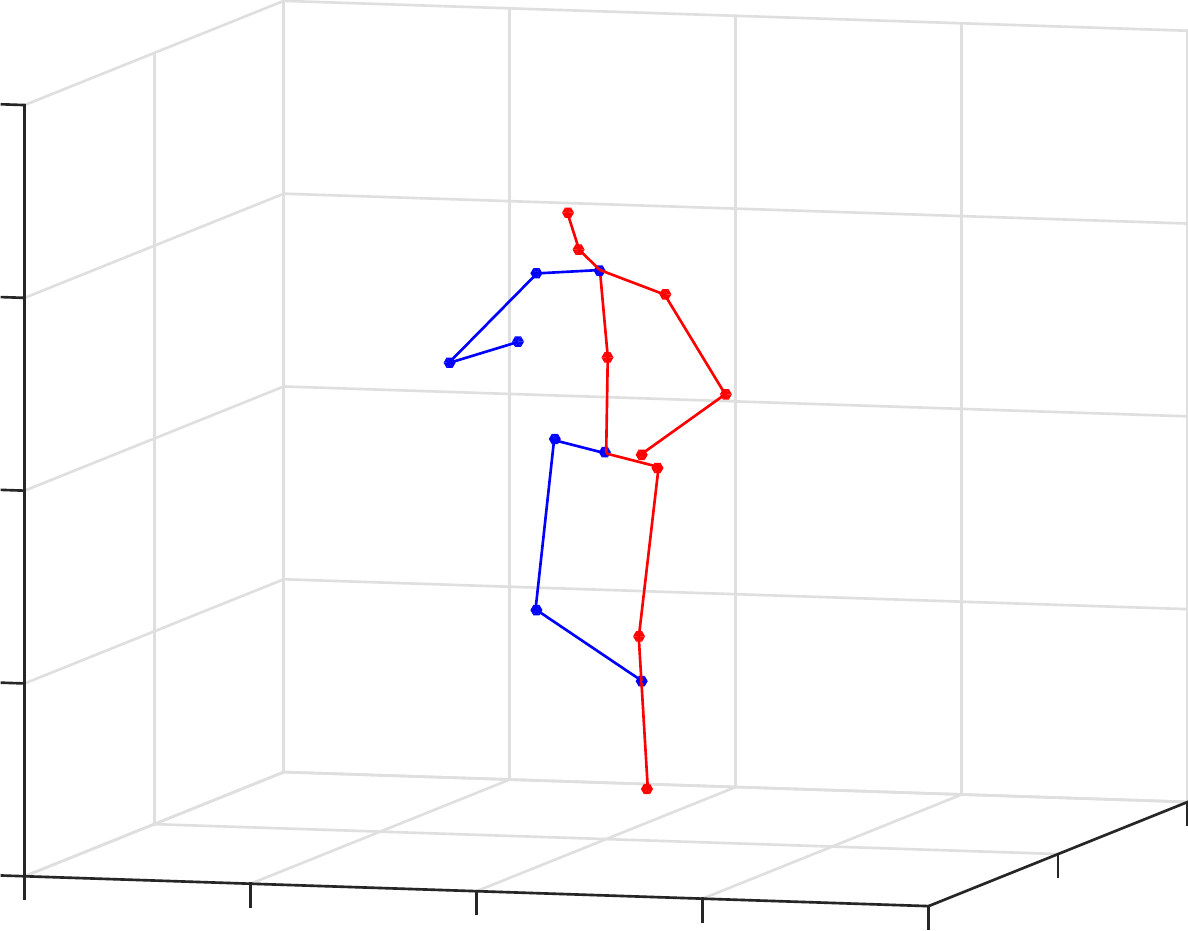}\\
        \end{tabular}
        \caption{Qualitative evaluation of Model I on Human3.6m dataset. \textbf{First} and \textbf{Fourth} column: Input 2d poses. \textbf{Second} and \textbf{Fifth} column: Ground truth 3d poses. 
        \textbf{Third} and \textbf{Sixth} column: 3d pose prediction using proposed Model I (model trained on Human3.6m with re-projection loss). Our model captures all the poses very accurately and performs better than the recent state-of-the-art methods. Quantitative results are given in Table~\ref{table1}. }
        \label{fig:Human36m images}
    \end{figure*}    
    
In the following section, we present an experimental evaluation of our proposed network on different benchmark datasets. We use Human3.6m and MPII datasets for training and show the quantitative performance of the proposed method on Human3.6m and qualitative performance on MPII to compare with state-of-the-art methods. For showing the generalization ability of our method we perform cross-dataset validation on MPI-INF-3DHP, and also present results after fine-tuning on MPI-INF-3DHP. We present a comparative analysis with different state-of-the-art methods for 3d pose estimation directly from 2d pose as well as the methods which give 3d pose prediction from images (end-to-end approaches). 

\section{Experimental Evaluation}
\subsection{Quantitative Results}
Evaluation on test datasets is done using standard 3d pose estimation metrics, MPJE (Mean Per Joint Error in mm) for Human3.6m dataset, along with PCK (Percent of Correct Keypoints) and AUC (Area Under the Curve)\cite{mono-3dhp2017,mehta2017vnect,luo2018orinet,dabral2018learning,zhou2017towards} for MPI-INF-3DHP, which are more robust and stronger metrics in identifying the incorrect joint predictions. To follow the evaluation convention in the existing works \cite{mono-3dhp2017}, we chose a threshold of 150 mm in the calculation of PCK. 
For quantitative evaluation on MPI-INF-3DHP, to account for the depth scale difference between datasets Human3.6m and MPI-INF-3DHP, the predicted 3d pose is re-targeted to the ground-truth ``universal" skeleton of MPI-INF-3DHP. This is done by scaling the predicted pose using ground-truth bone lengths while preserving directions of bones, following standard practices \cite{mehta2017vnect,luo2018orinet, zhou2017towards}. Moreover, we also account for the difference in pelvis joint definitions between Human3.6m and MPI-INF-3DHP during the evaluation of our model II on MPI-INF-3DHP. The location of the predicted pelvis and hip joints are moved towards the neck in a fixed ratio (0.2) before evaluation \cite{zhou2017towards}.

\textbf{Human3.6m}: Table~\ref{table1} shows results on Human3.6m under defined protocol in  \cite{ionescu2014human3} using Model I, which is trained on Human3.6m dataset under
full supervision. As shown in the table, we achieve greater accuracy than the state-of-the-art methods on most of the actions including difficult actions such as Sitting, Greeting, etc in terms of MPJE (Mean Per Joint Error, mm). On average we have an overall improvement of 6\% over our baseline method\cite{martinez2017simple}, which is also trained on 2d pose ground-truth. This improvement in accuracy can be attributed to the 3d-to-2d re-projection loss minimization and geometric constraints. We also outperform state-of-the-art method \cite{sun2017compositional} which was trained on the input images from both Human3.6m and MPII, using our Model I trained on Human3.6m alone.

\vspace{1em}
\textbf{MPI-INF-3DHP}: 
For MPI-INF-3DHP dataset, quantitative evaluation has been done using standard metrics PCK, AUC and MPJE as used in state-of-the-art methods \cite{mono-3dhp2017,mehta2017vnect,luo2018orinet,dabral2018learning,zhou2017towards}. 

\textbf{(a) Cross Dataset Evaluation:}
Table~\ref{table2} shows evaluation results on MPI-INF-3DHP with our Model I (trained on Human 3.6M) and Model II (trained on Human 3.6M + MPII) in terms of PCK and AUC for all three different settings (GS, NoGS and Outdoor) for the 2935 testing images. On an average we see an improvement of 2.3\% on PCK (with a threshold of 150 mm) and 6.2\% on AUC over the best performing state-of-the-art method. This establishes the improved cross-dataset generalization of our method when compared to the state-of-the-art methods.

\textbf{(b) Results after Fine-tuning:} We also present a performance analysis of our Model III (Model I fine-tuned on MPI-INF-3DHP dataset) in Table~\ref{table3}. It shows a comparative analysis of the activity-wise performance of Model III with all recent state-of-the-art methods. We have achieved significant improvement over the state-of-the-art on all the actions in terms of all the metrics. On an average we exceed the best accuracy achieved by methods fully supervised on MPI-INF-3DHP by 2\% on PCK, 17\% on AUC and 16\% on MPJE.


\subsection{Qualitative Results}

\begin{figure*}[t!]
        \centering
        \setlength{\belowcaptionskip}{-7pt}
        \begin{tabular}{cccc}
            \includegraphics[keepaspectratio=true, scale = 0.16]{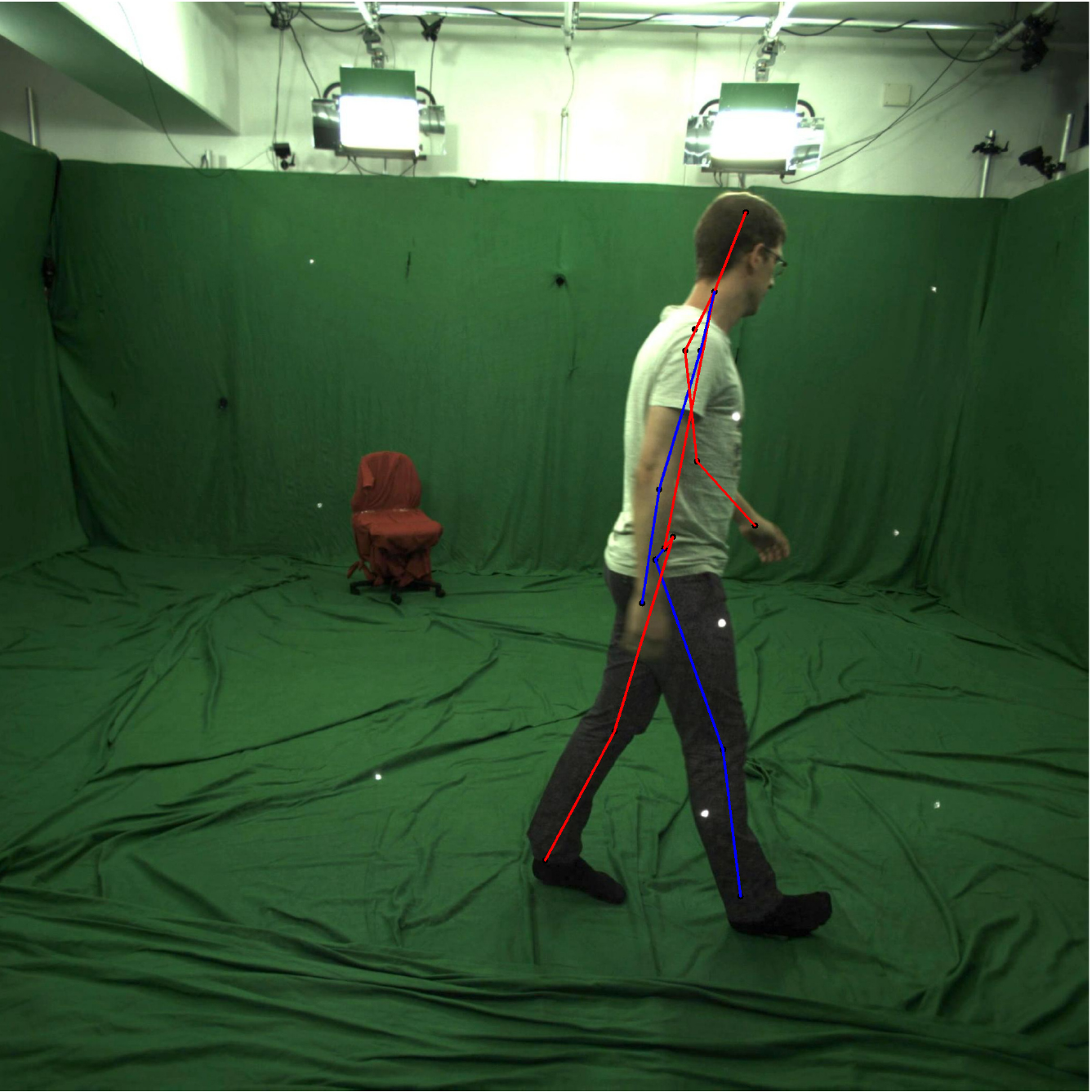} & \includegraphics[keepaspectratio=true, scale = 0.169]{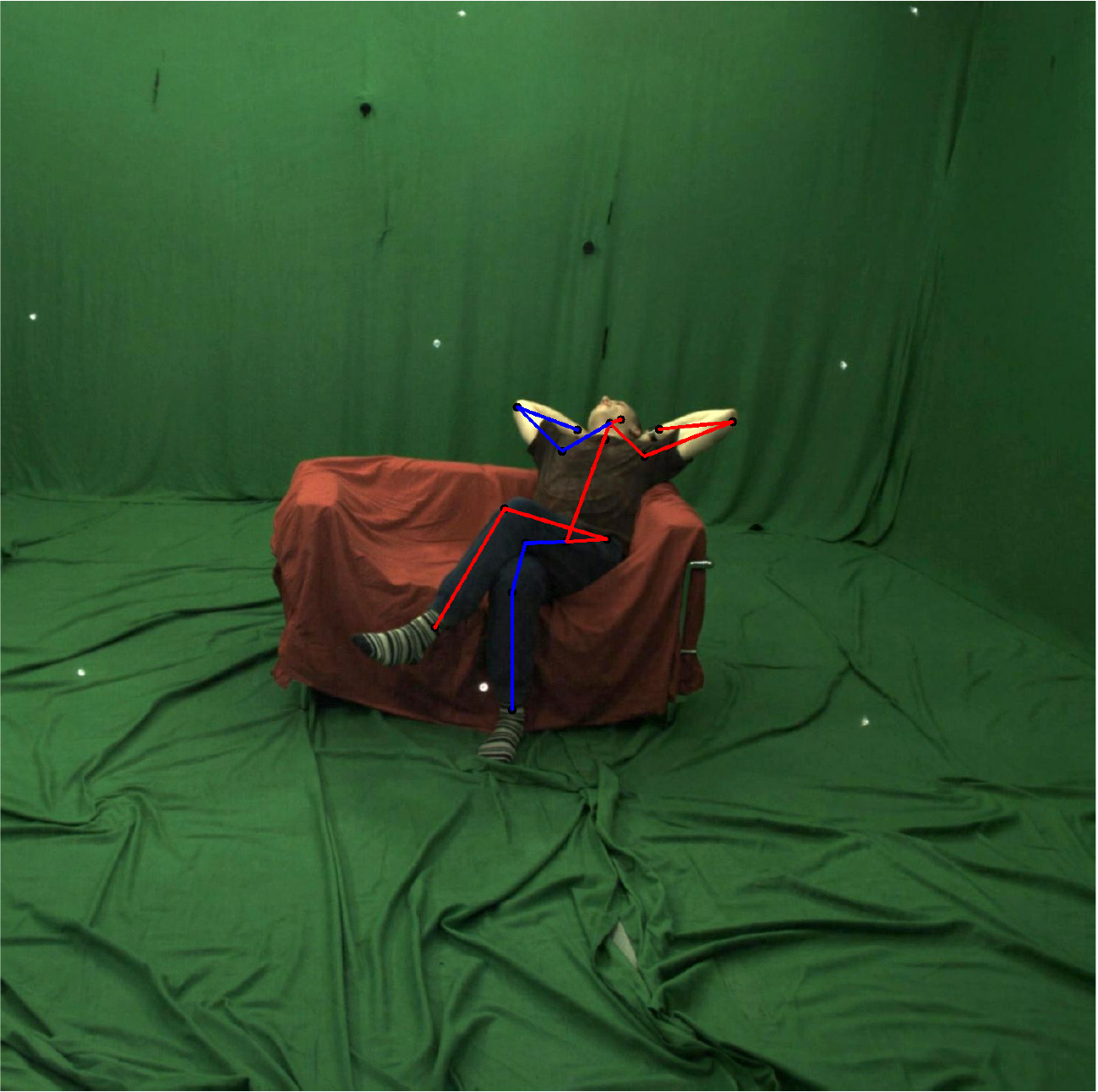} & \includegraphics[keepaspectratio=true, scale = 0.16]{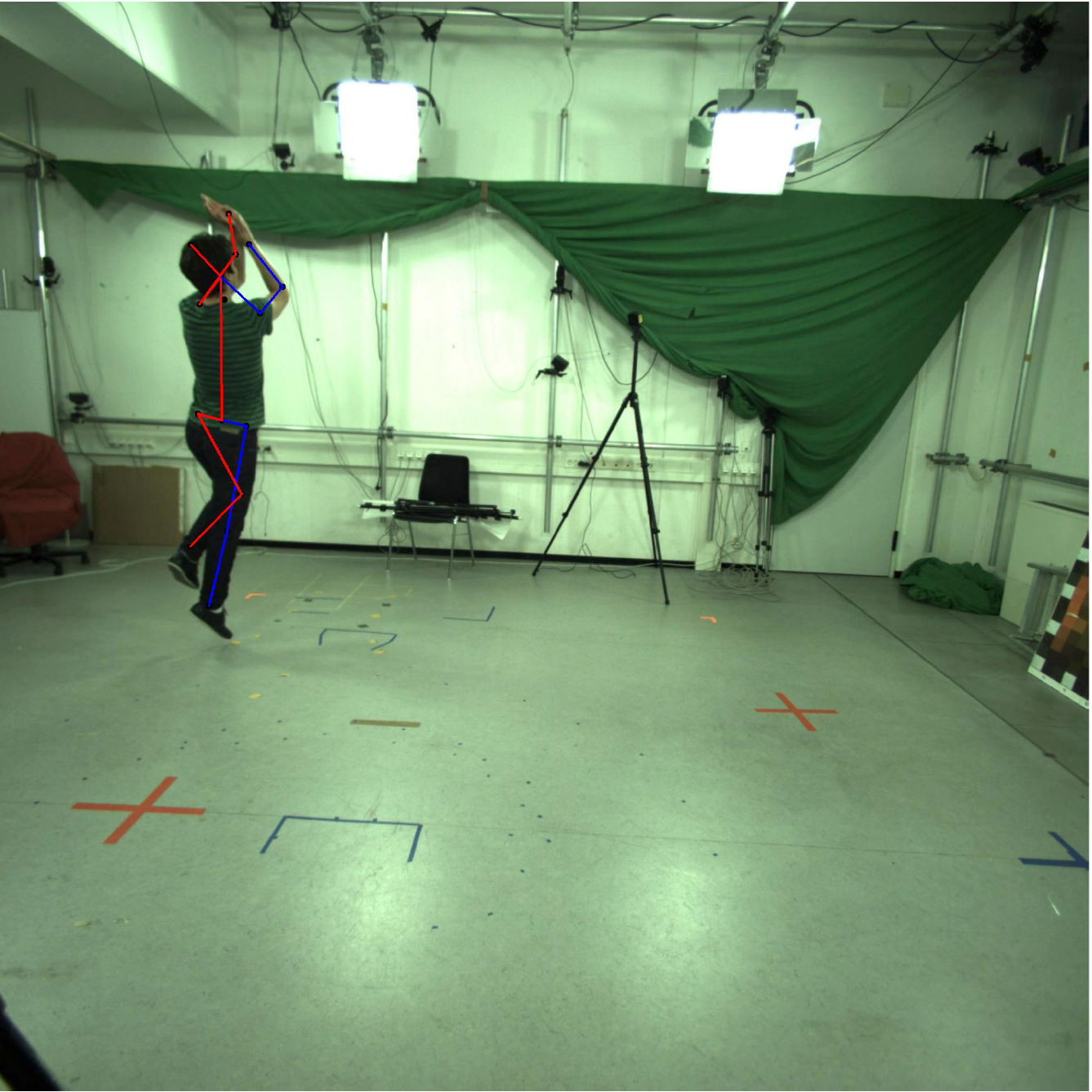} & \includegraphics[keepaspectratio=true, scale = 0.15]{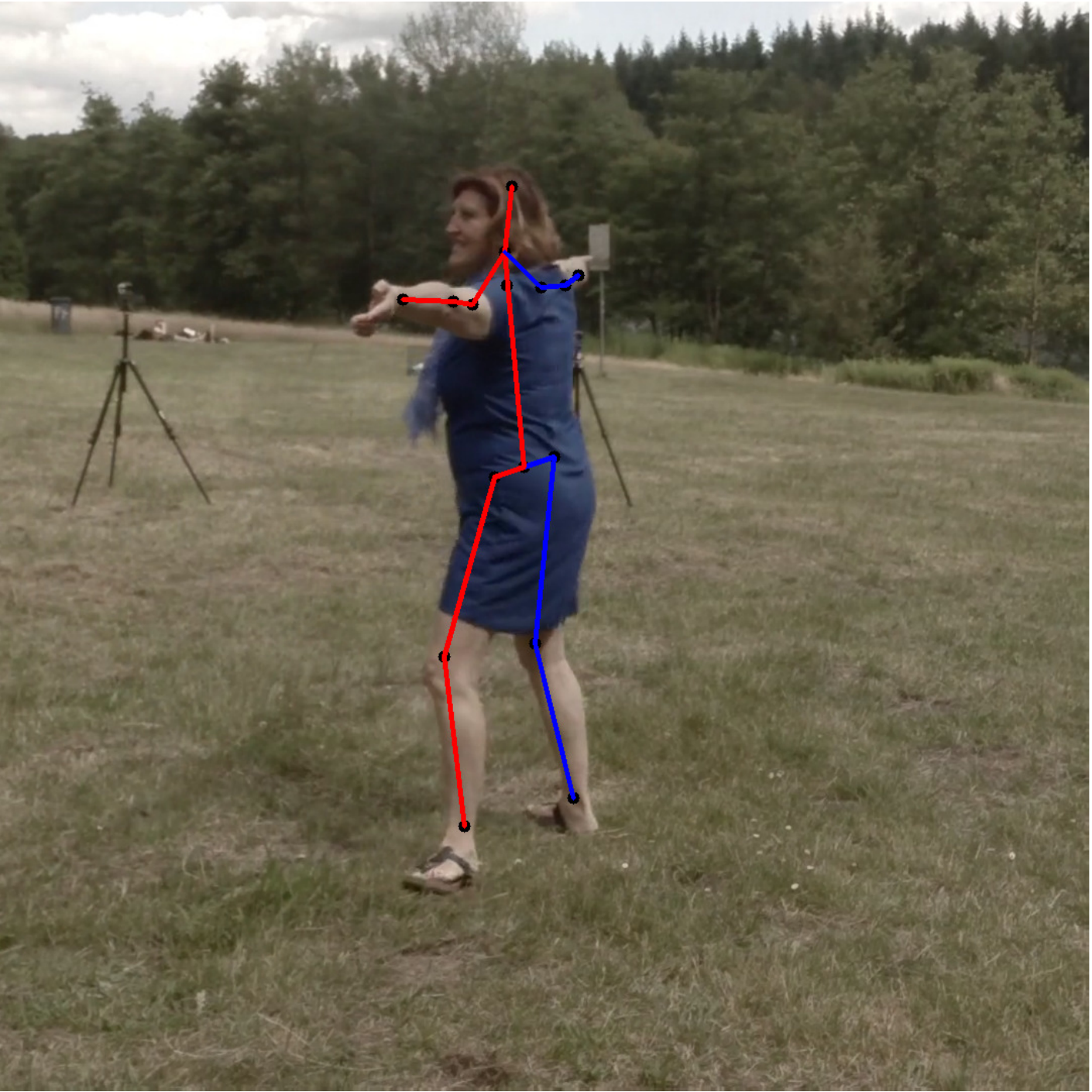} \\   
            \includegraphics[keepaspectratio=true, scale = 0.19]{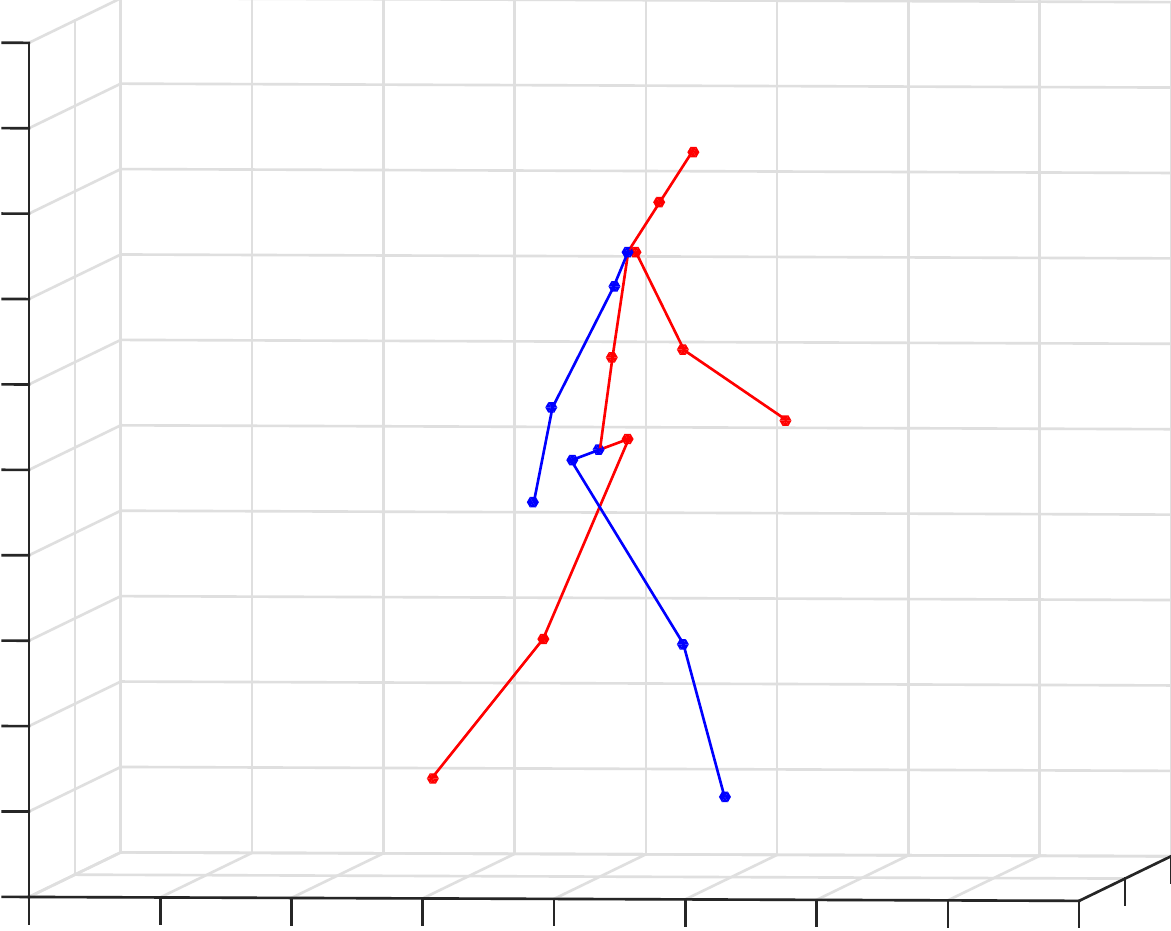} & \includegraphics[keepaspectratio=true, scale = 0.19]{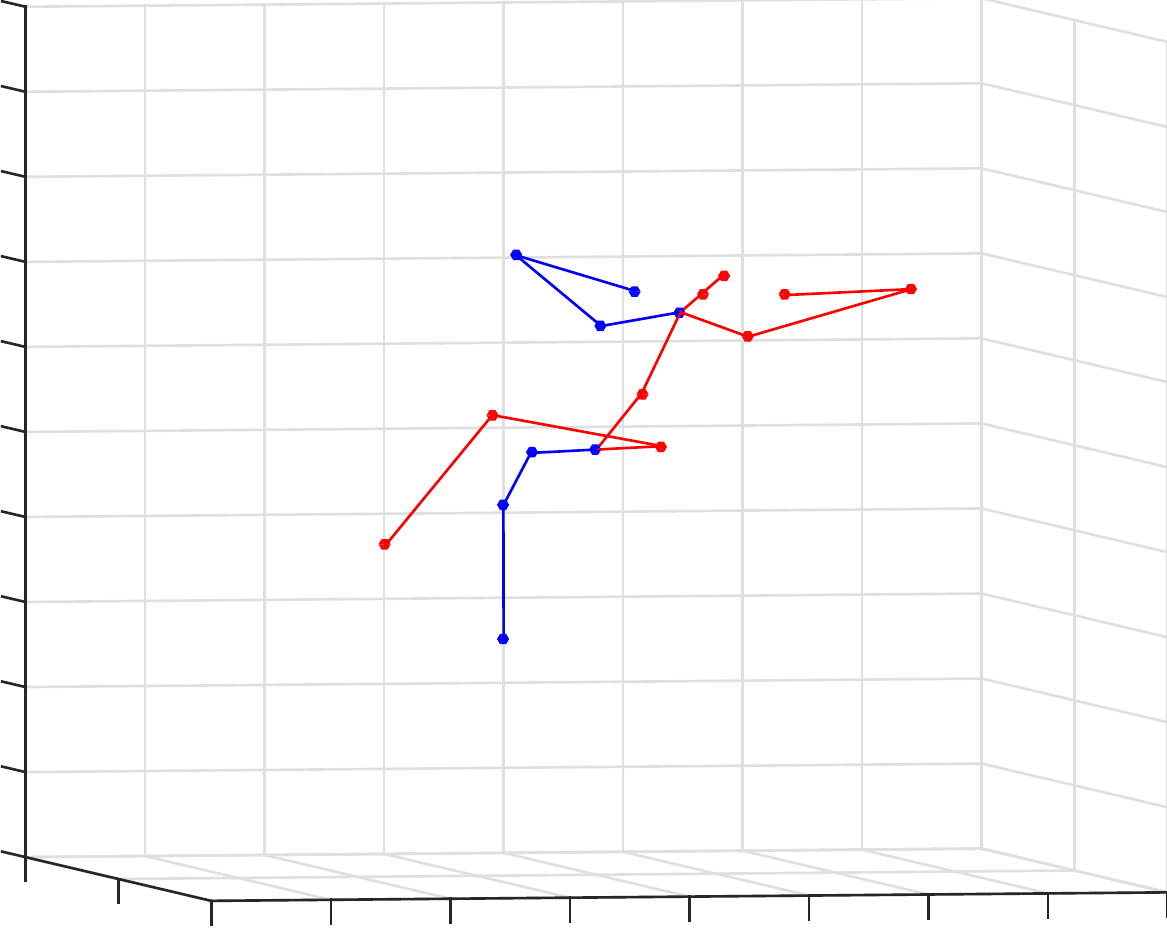} & \includegraphics[keepaspectratio=true, scale = 0.2]{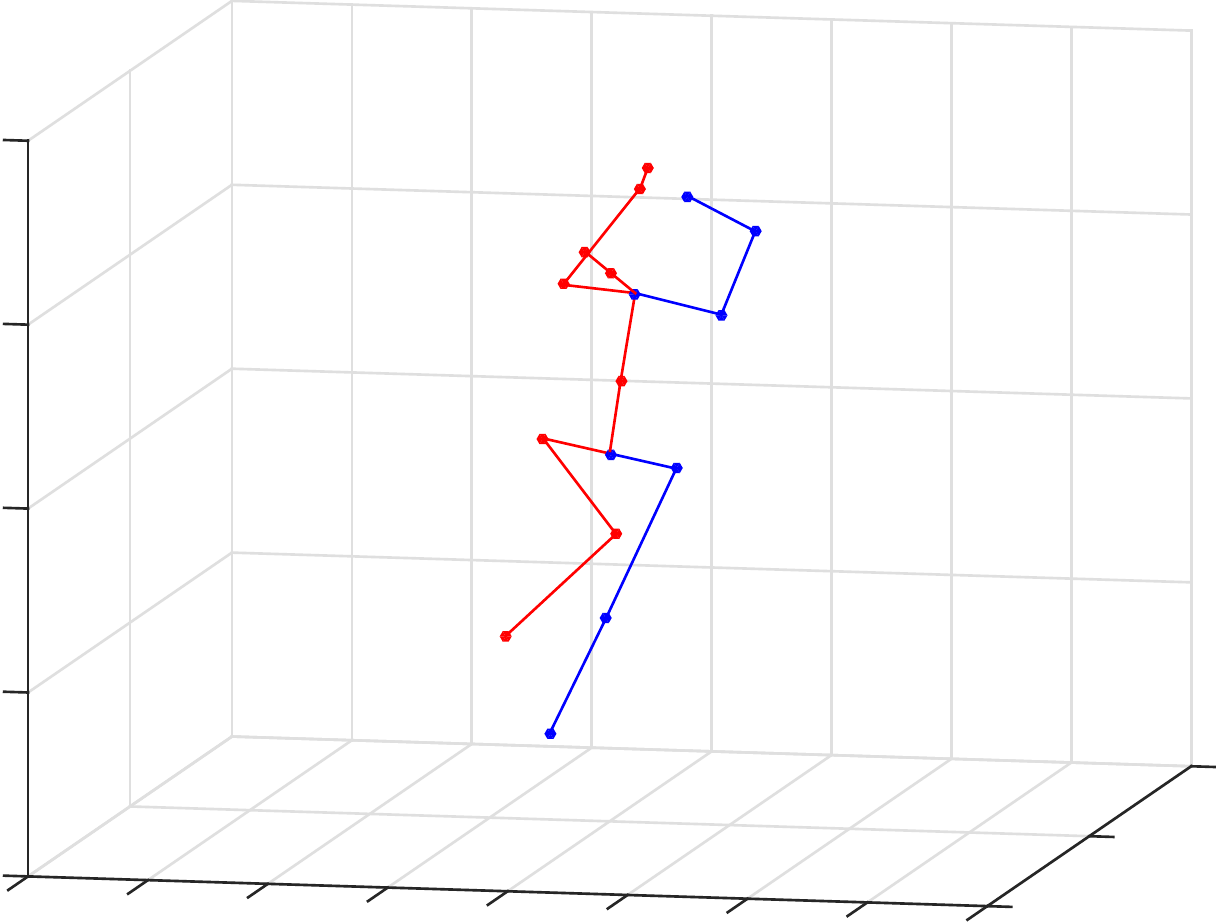} & \includegraphics[keepaspectratio=true, scale = 0.19]{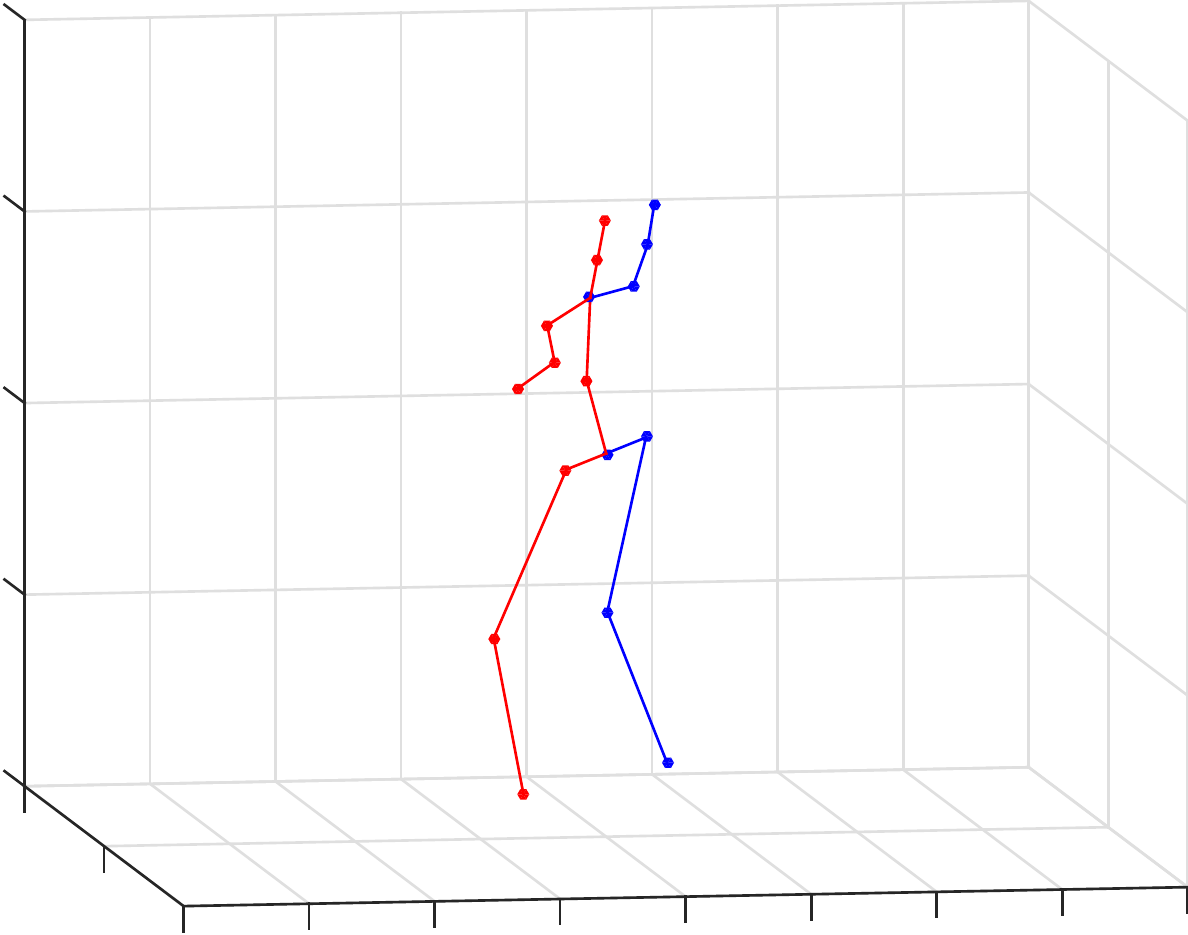} \\
            \includegraphics[keepaspectratio=true, scale = 0.14]{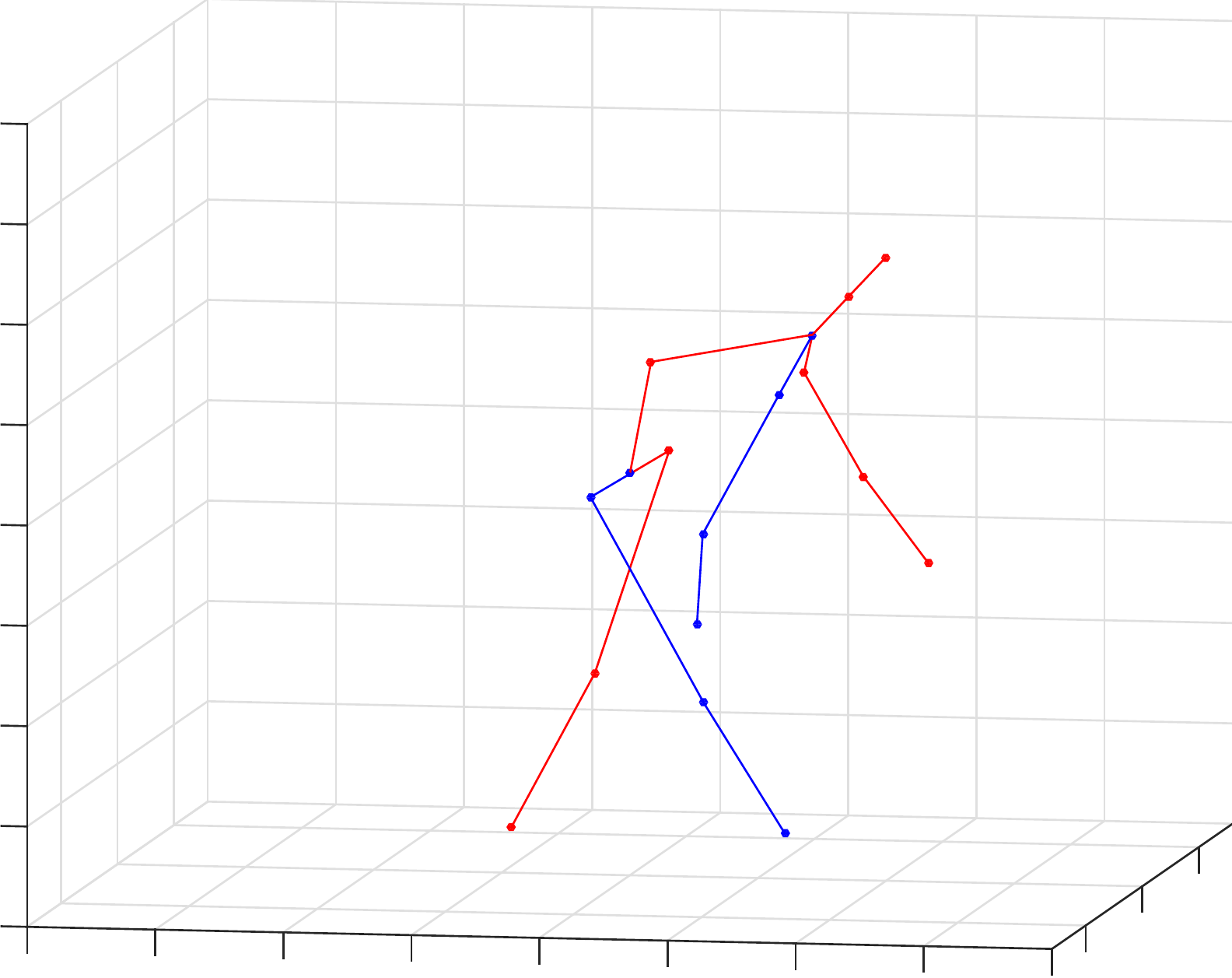} & \includegraphics[keepaspectratio=true, scale = 0.14]{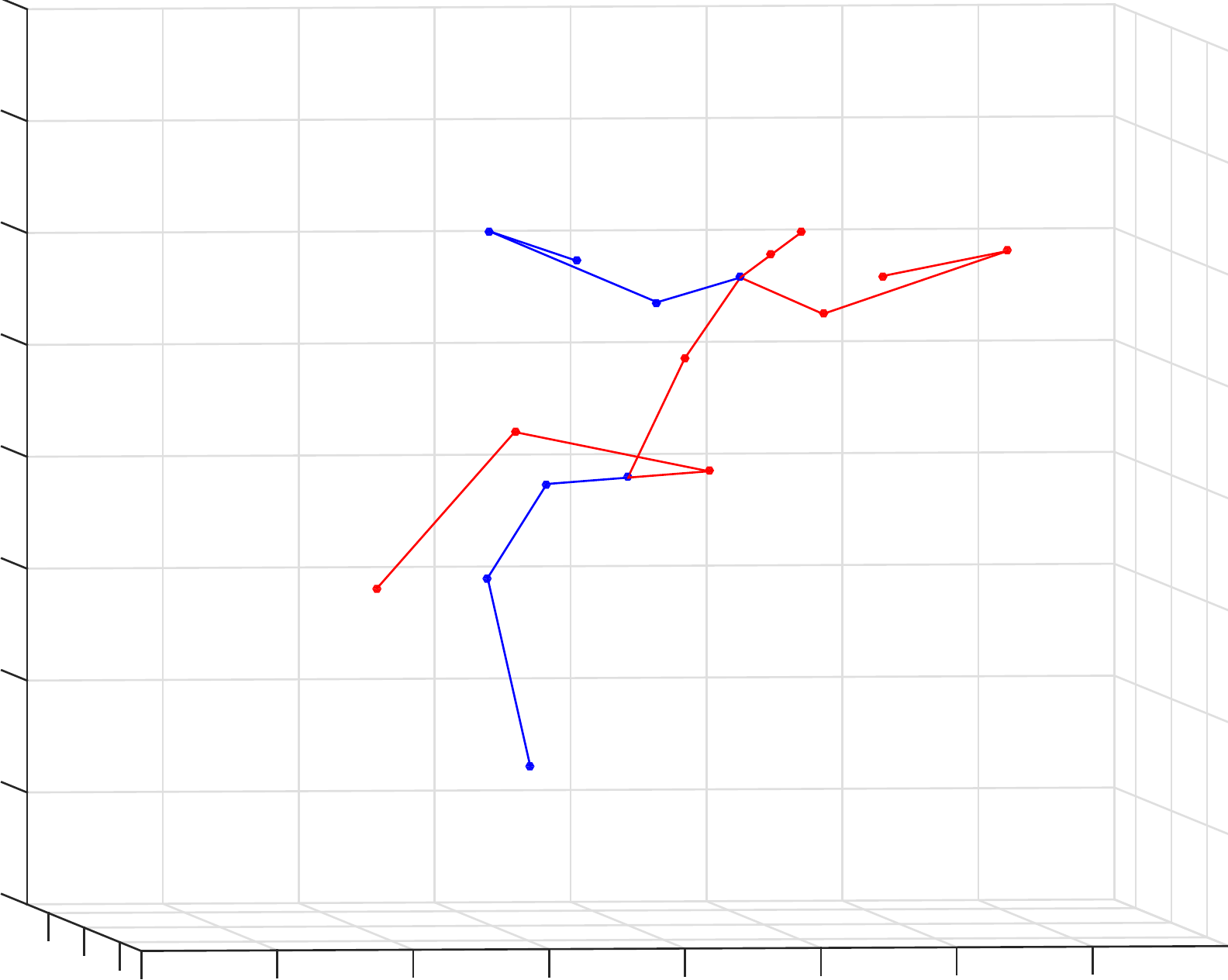} & \includegraphics[keepaspectratio=true, scale = 0.14]{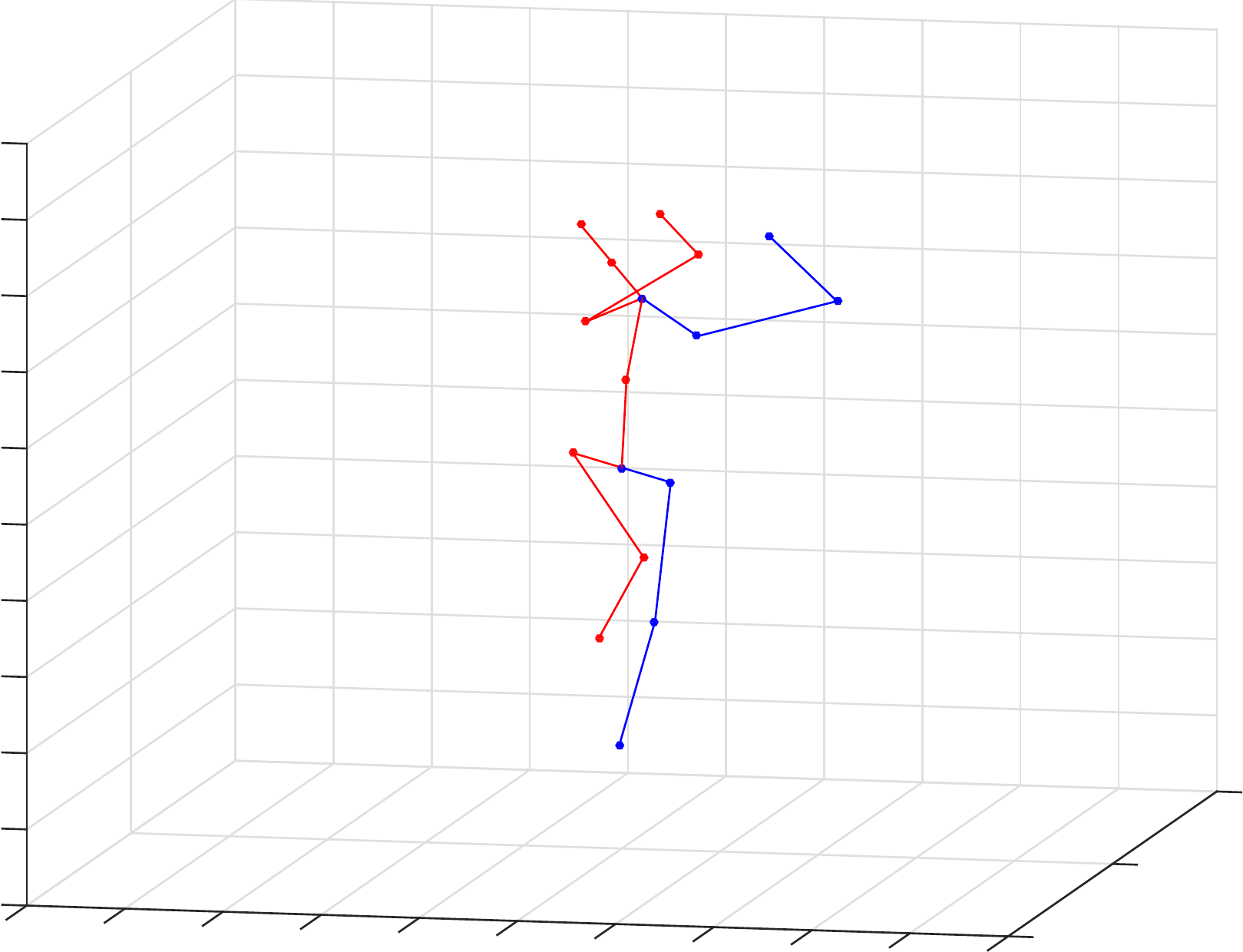} & \includegraphics[keepaspectratio=true, scale = 0.19]{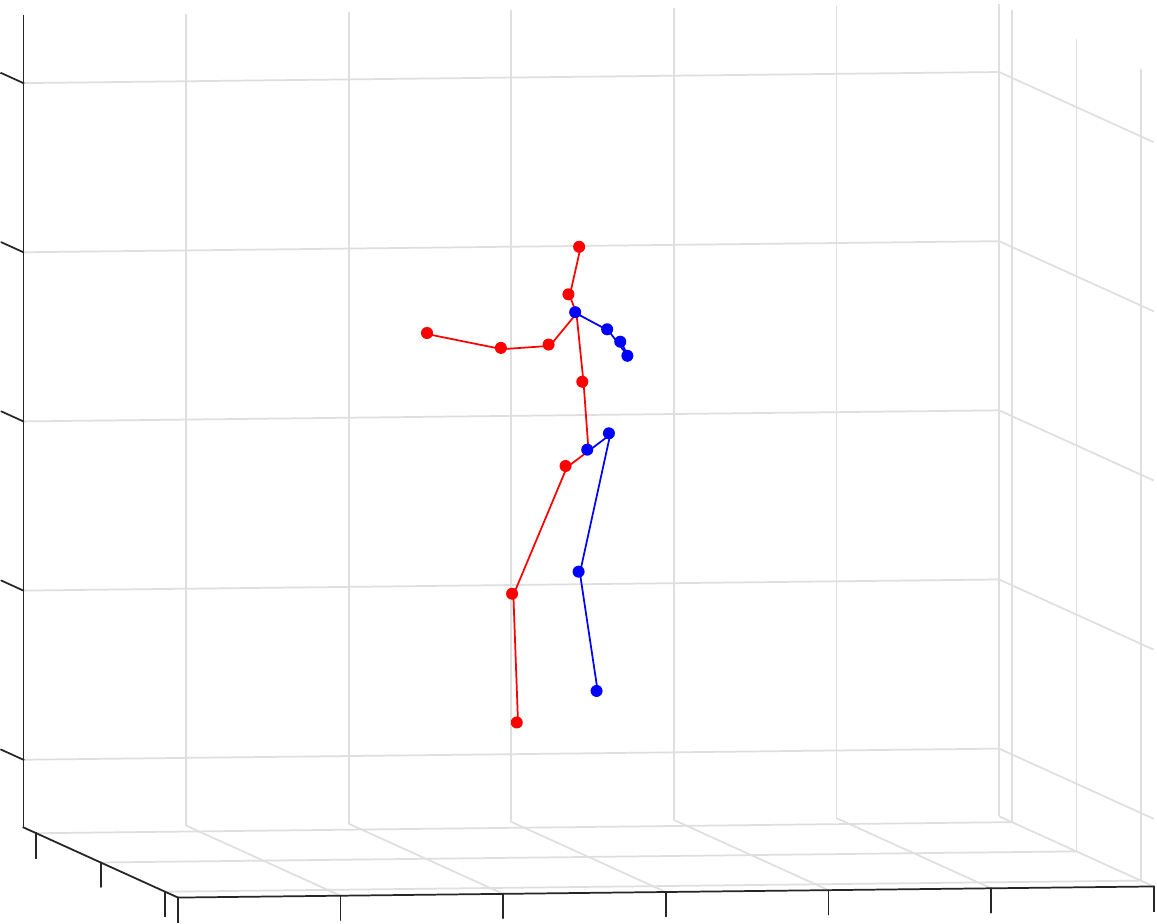} \\
            \includegraphics[keepaspectratio=true, scale = 0.14]{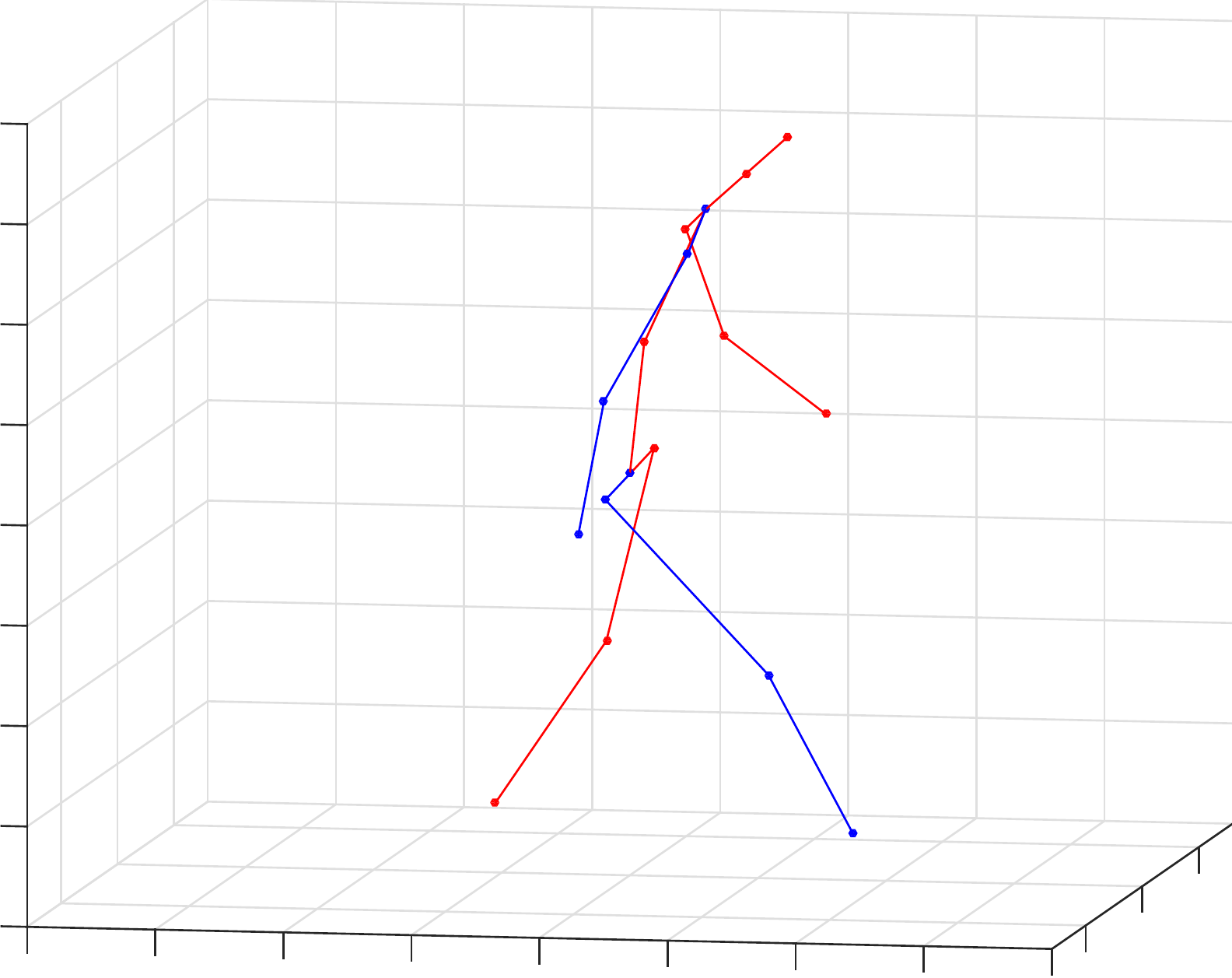} & \includegraphics[keepaspectratio=true, scale = 0.14]{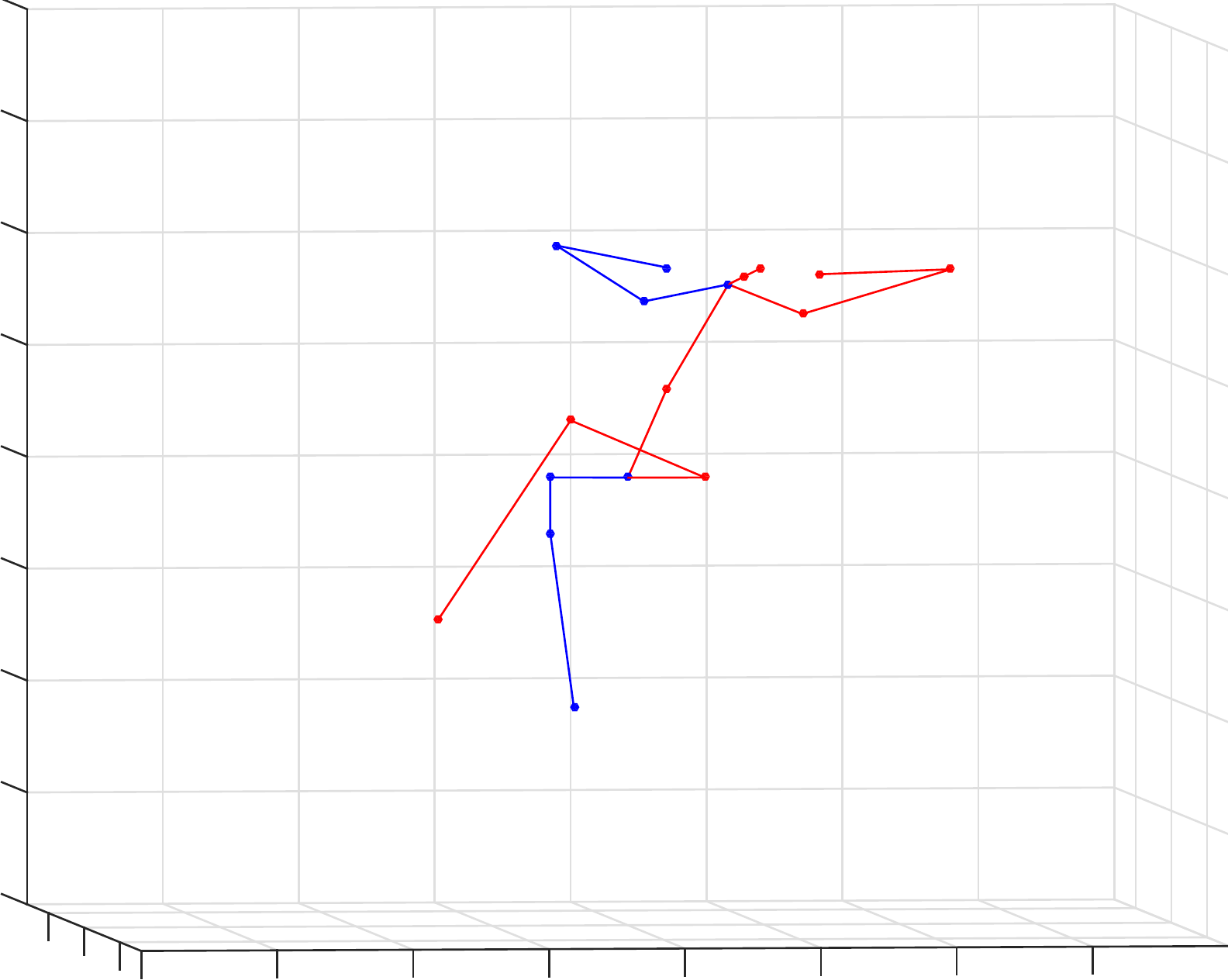} & \includegraphics[keepaspectratio=true, scale = 0.14]{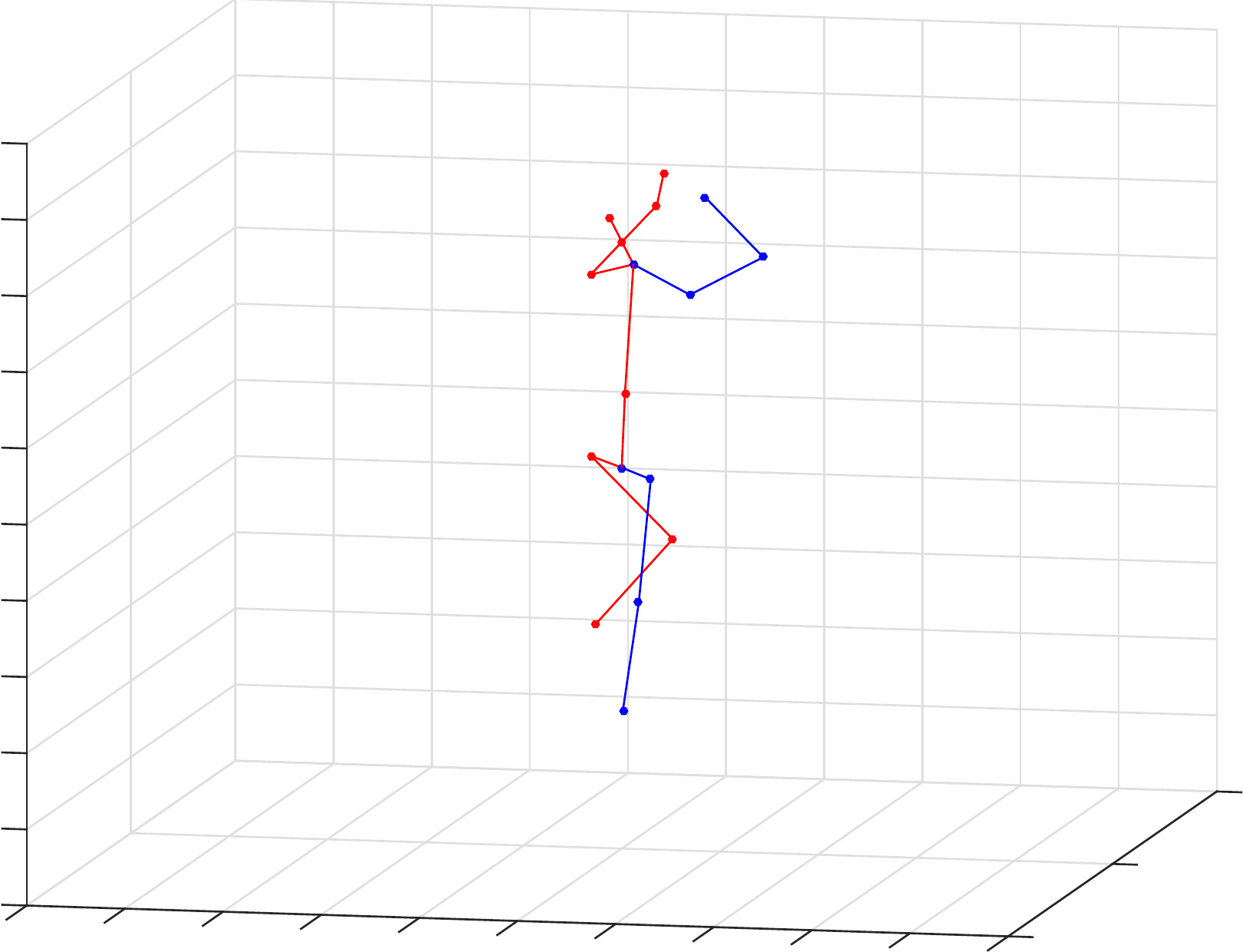} & \includegraphics[keepaspectratio=true, scale = 0.14]{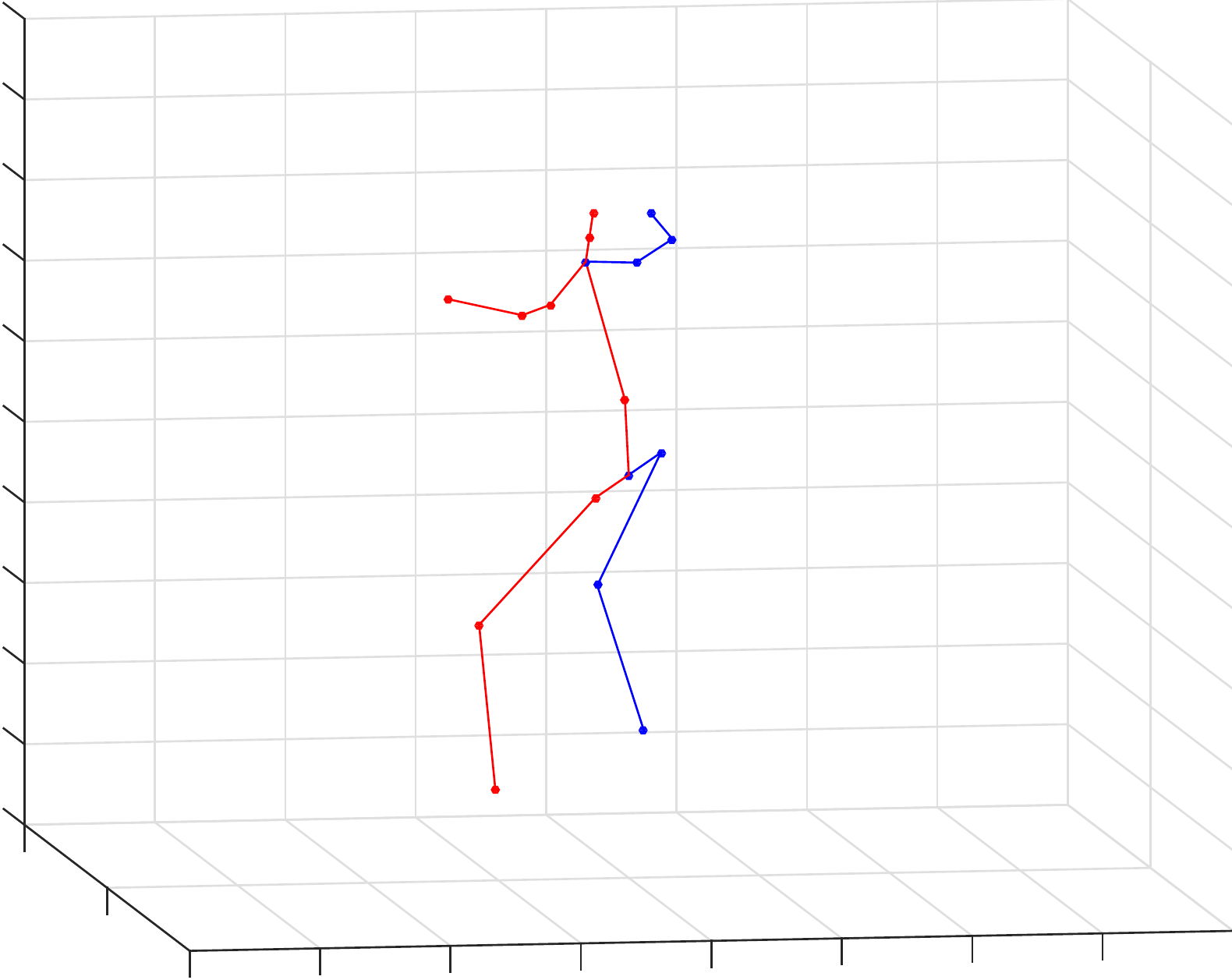} \\
            \includegraphics[keepaspectratio=true, scale = 0.185]{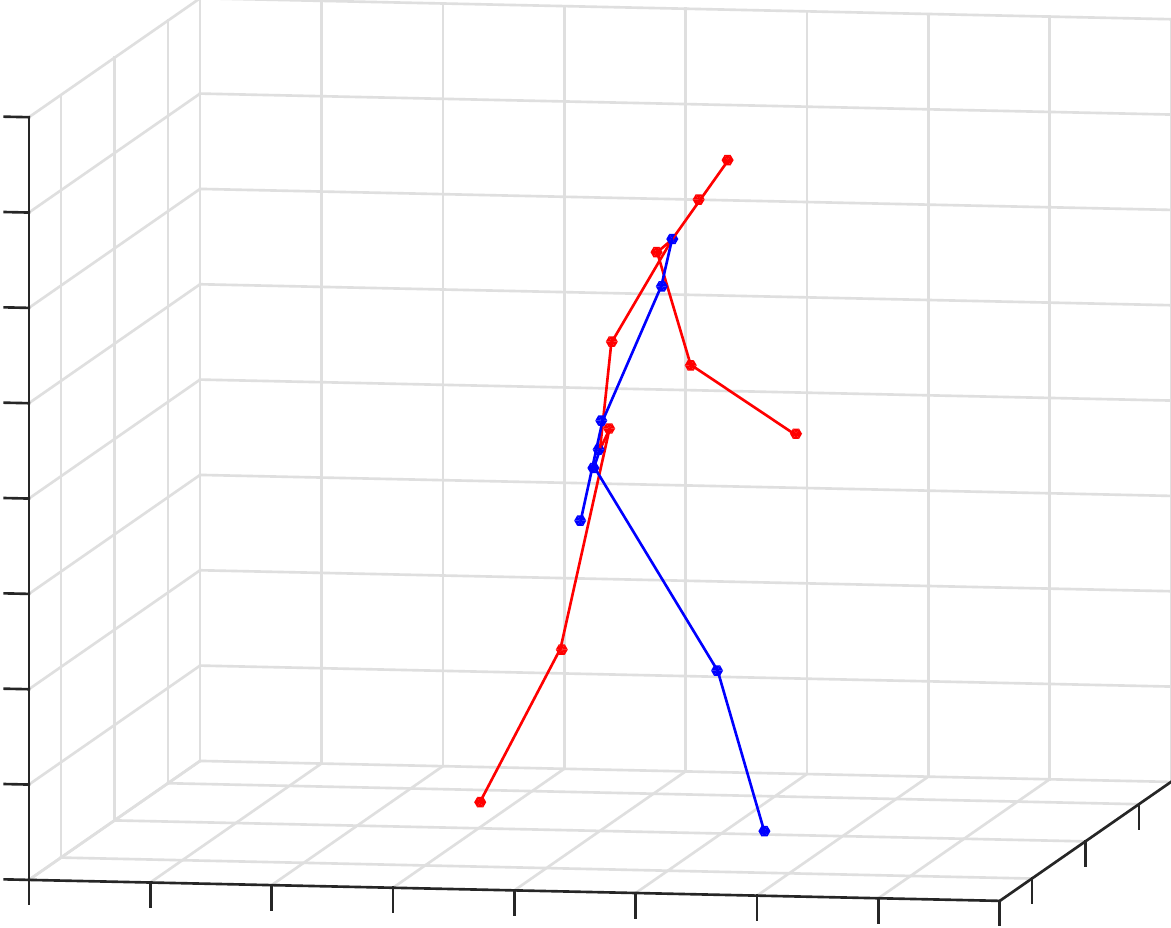} & \includegraphics[keepaspectratio=true, scale = 0.185]{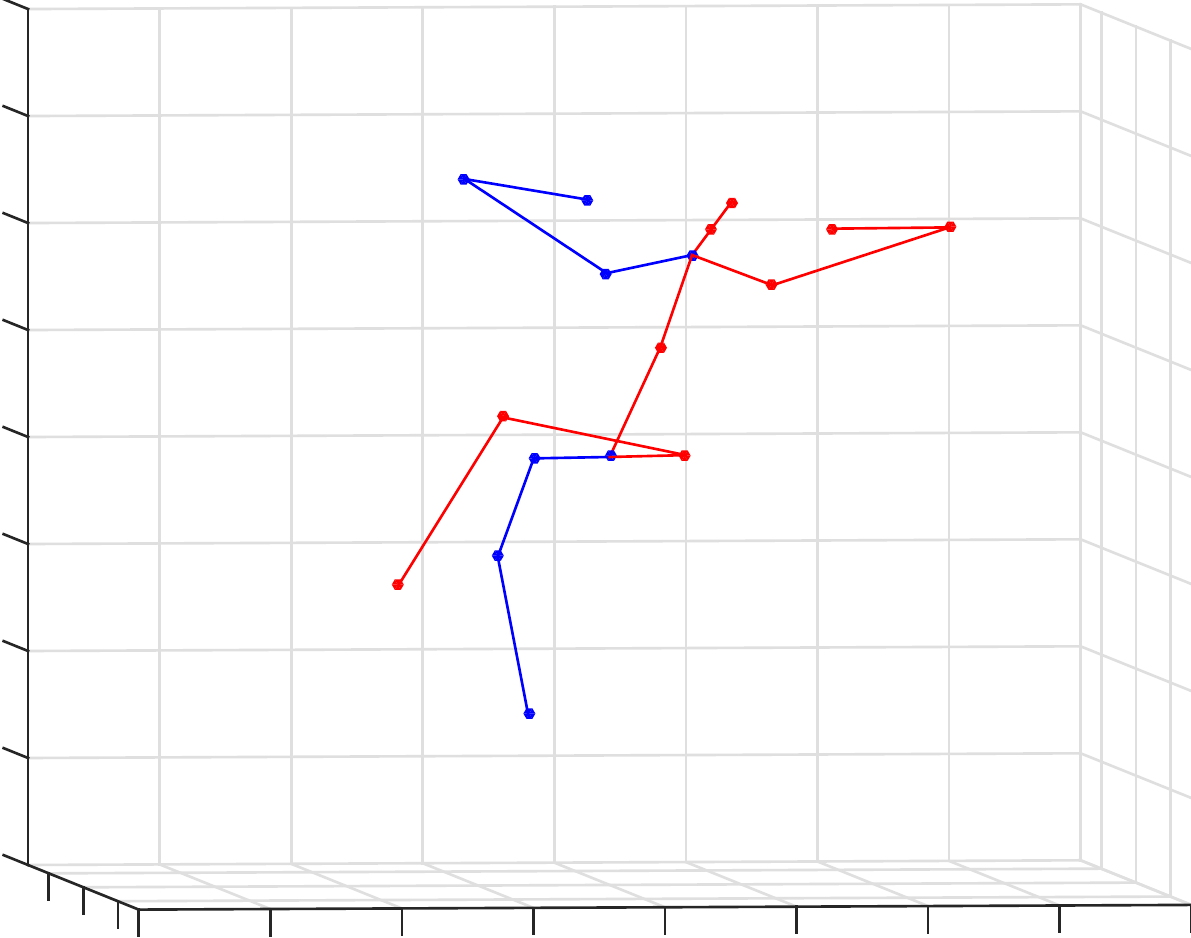} & \includegraphics[keepaspectratio=true, scale = 0.195]{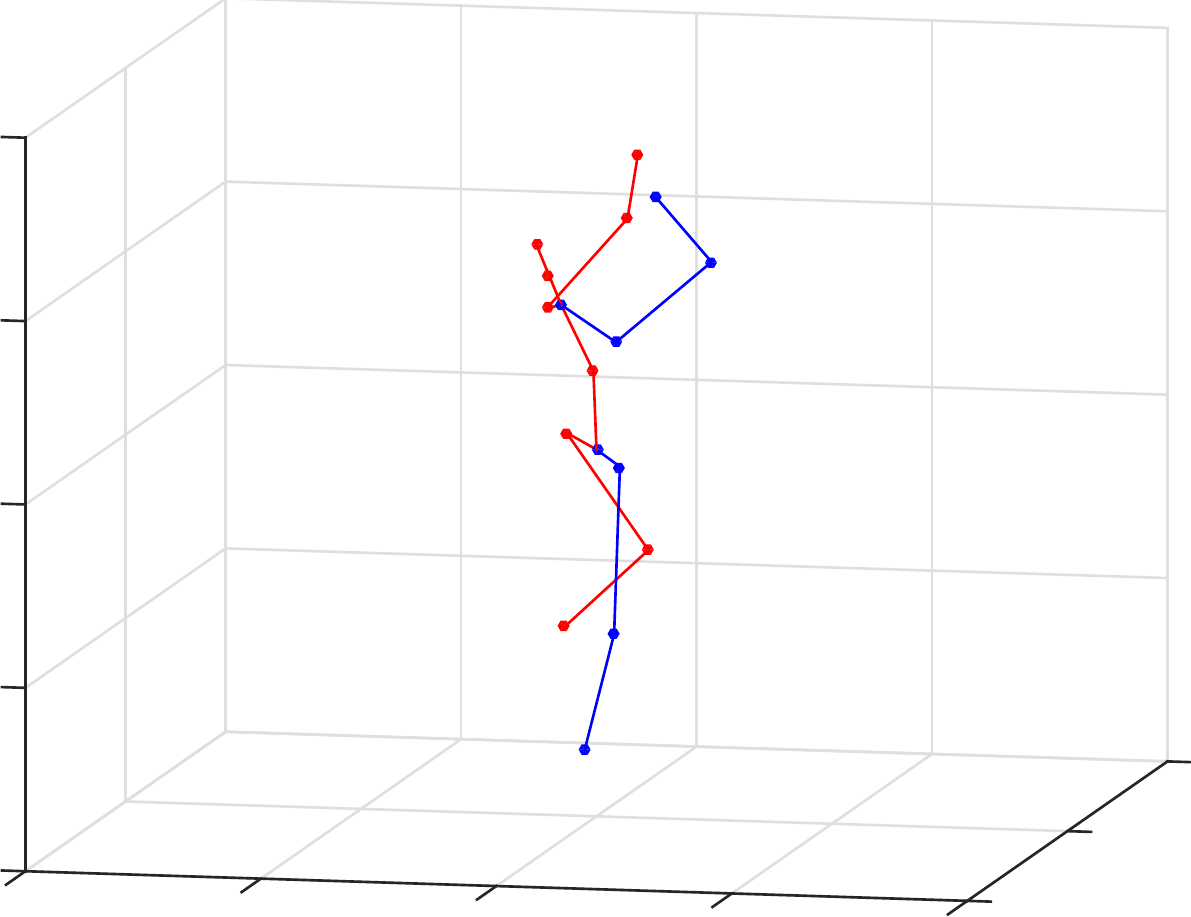} & \includegraphics[keepaspectratio=true, scale = 0.19]{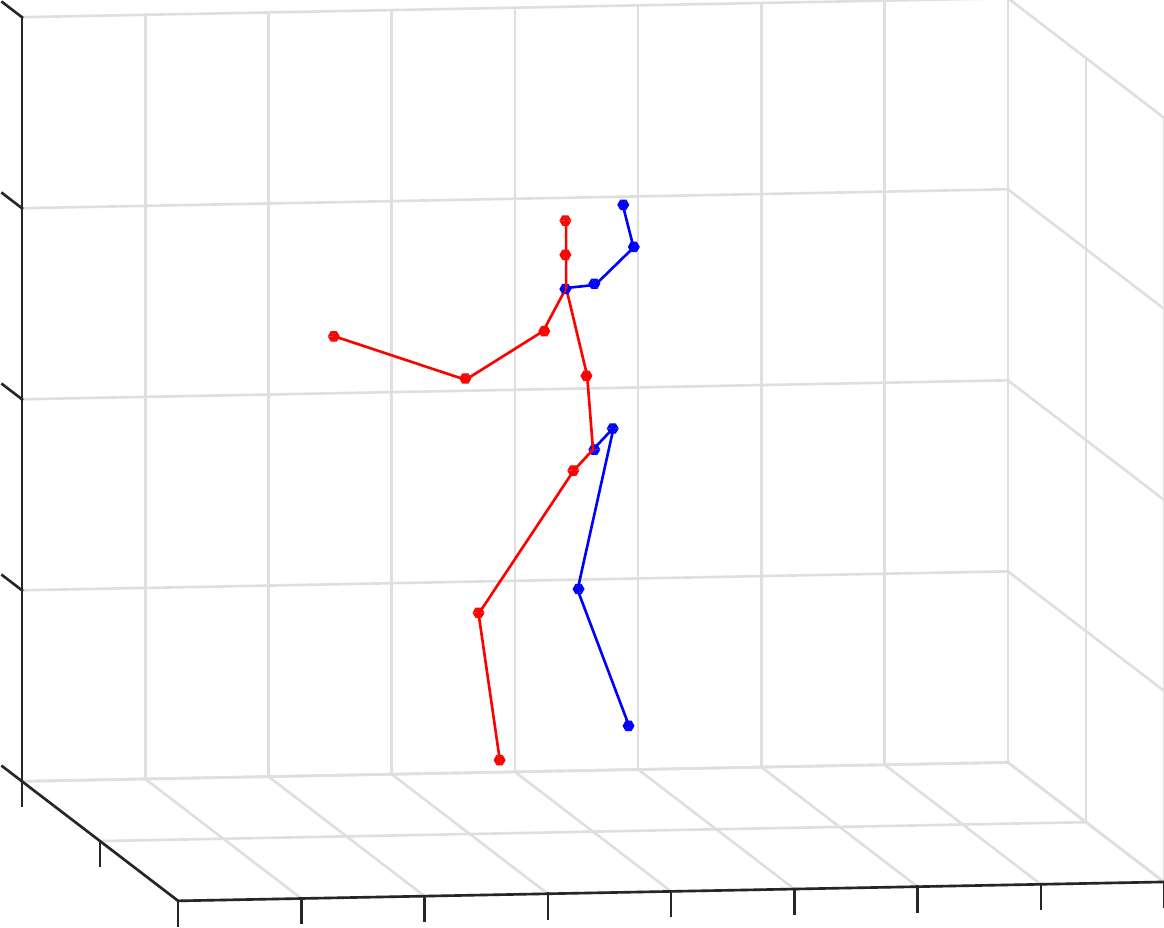} \\
            \includegraphics[keepaspectratio=true, scale = 0.14]{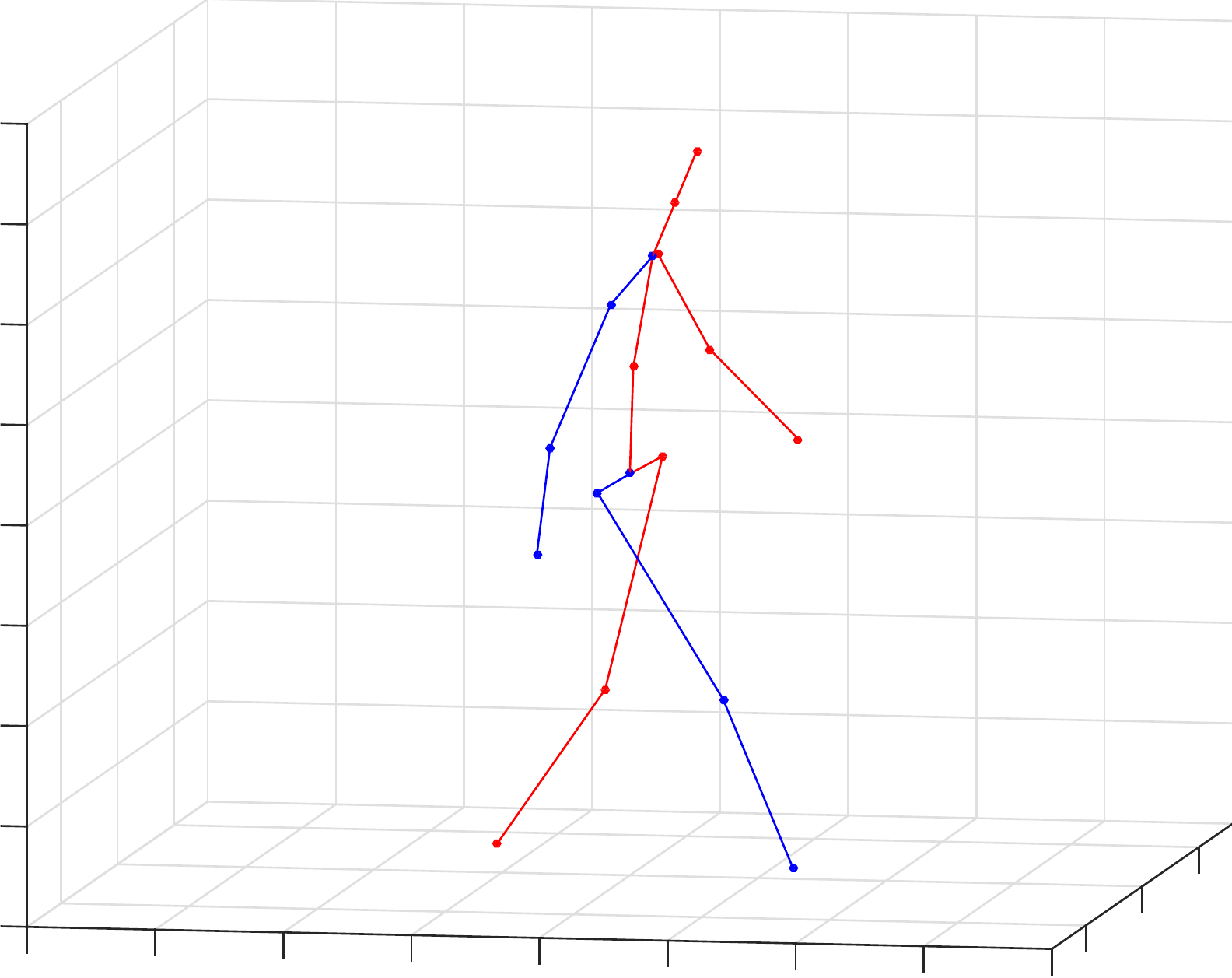} & \includegraphics[keepaspectratio=true, scale = 0.14]{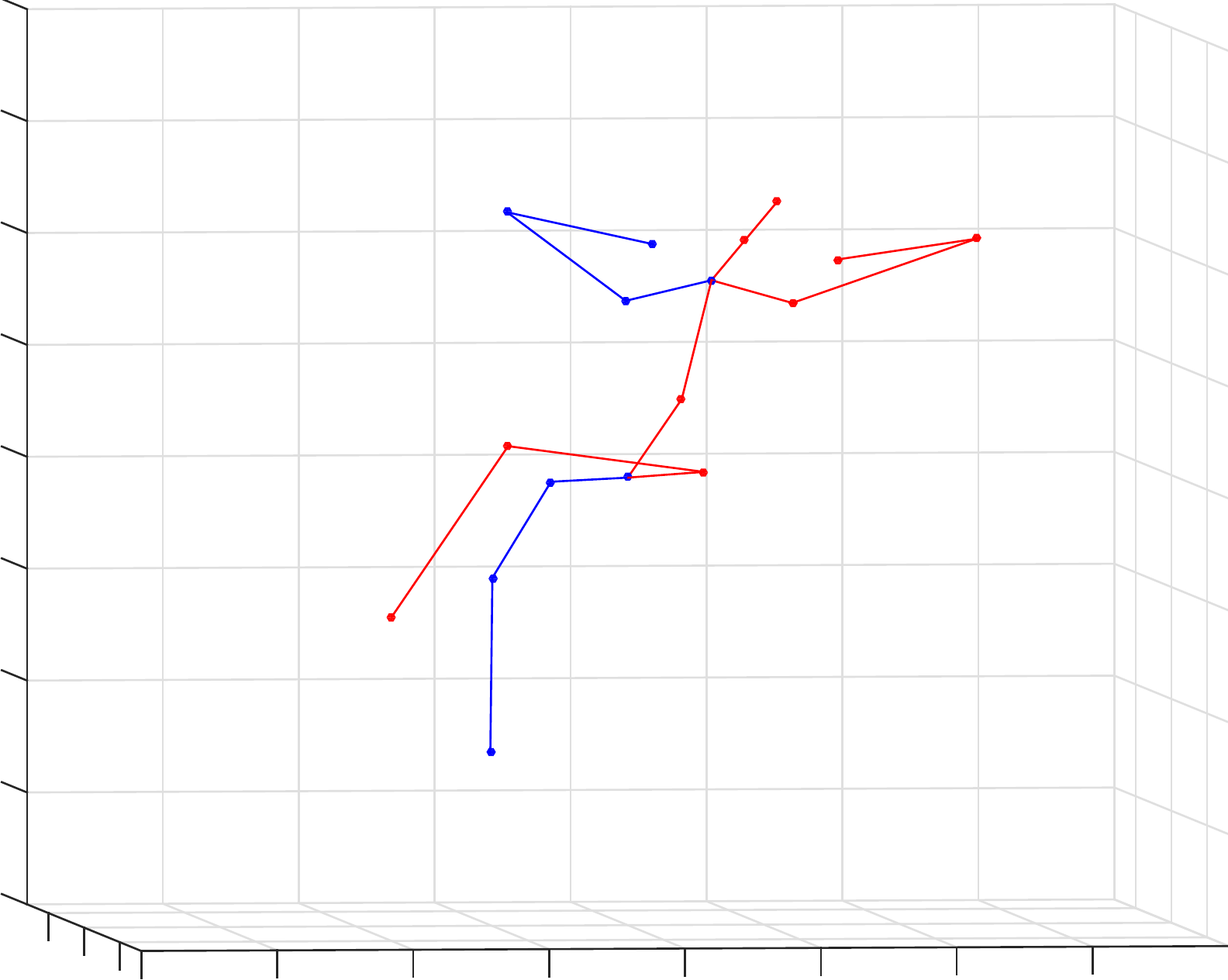} & \includegraphics[keepaspectratio=true, scale = 0.14]{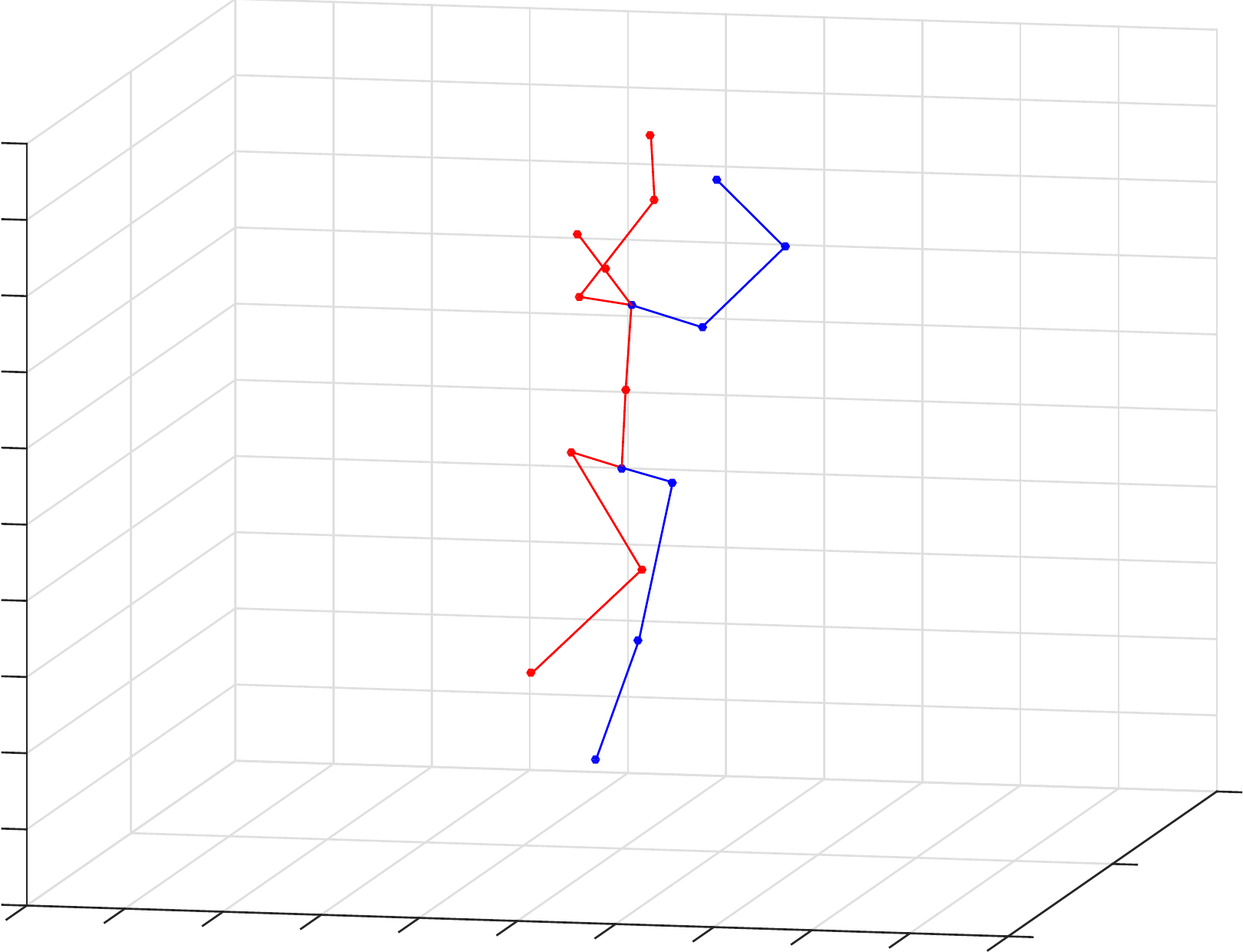} & \includegraphics[keepaspectratio=true, scale = 0.14]{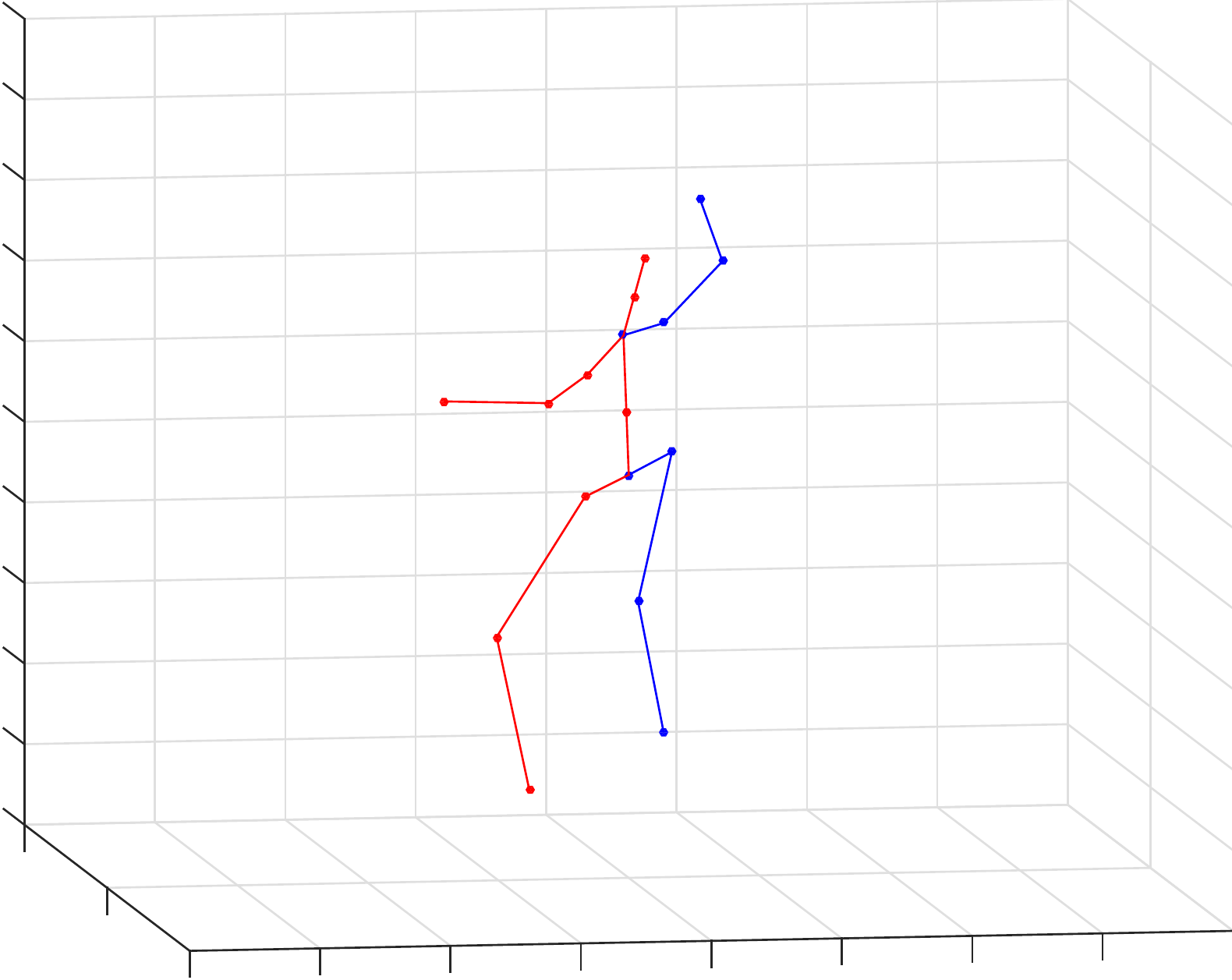} \\
            (a) & (b) & (c) & (d)
        \end{tabular}
        \caption{Qualitative evaluation on MPI-INF-3DHP dataset. \textbf{First} row: Input images with ground truth 2d pose. \textbf{Second} row: 3d ground truth poses. \textbf{Third} row: Prediction of the baseline network.
\textbf{Fourth} row: Prediction of proposed Model I (Model trained on Human3.6m dataset.). \textbf{Fifth} row: Prediction of proposed Model II (Model I fine tuned on MPII dataset.). \textbf{Sixth} row: Prediction of Model III (Model I fine tuned on MPI-INF-3DHP dataset). The baseline model fails to capture proper 3d pose in many cases, e.g in Figure (d), hands of the lady are predicted in a more downward position than that of the ground truth 3d pose. All variations of our proposed model can recover this pose using re-projection loss along with baseline supervised loss. Quantitative results are given in Table~\ref{table2} and Table~\ref{table3}.}
        \label{fig:MPII-INF images}
    \end{figure*}

\begin{figure*}[t!]
        \centering
        \setlength{\belowcaptionskip}{-9pt}
        \begin{tabular}{ccccc}
            \includegraphics[keepaspectratio=true, scale = 0.11]{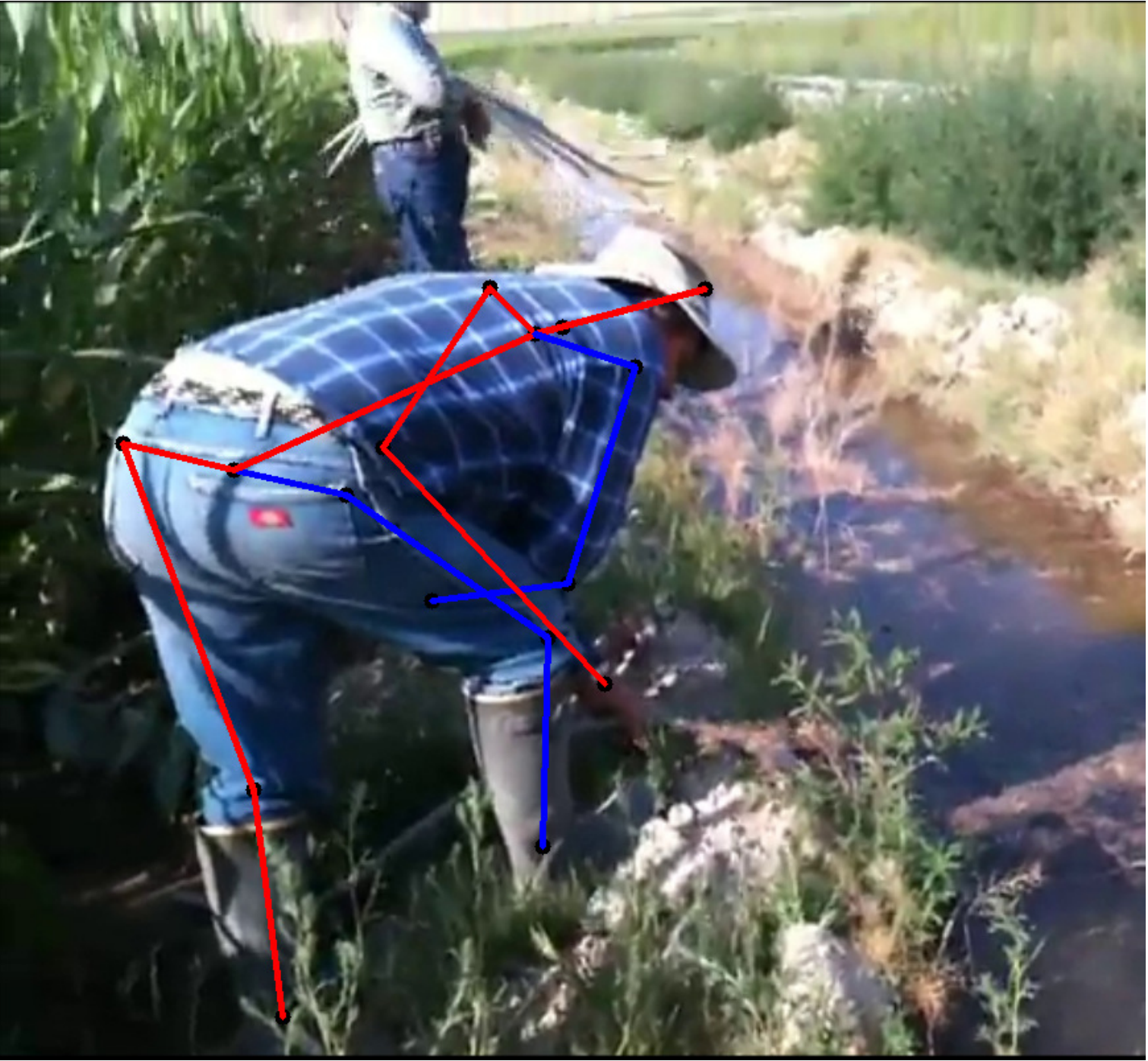} & \includegraphics[keepaspectratio=true, scale = 0.11]{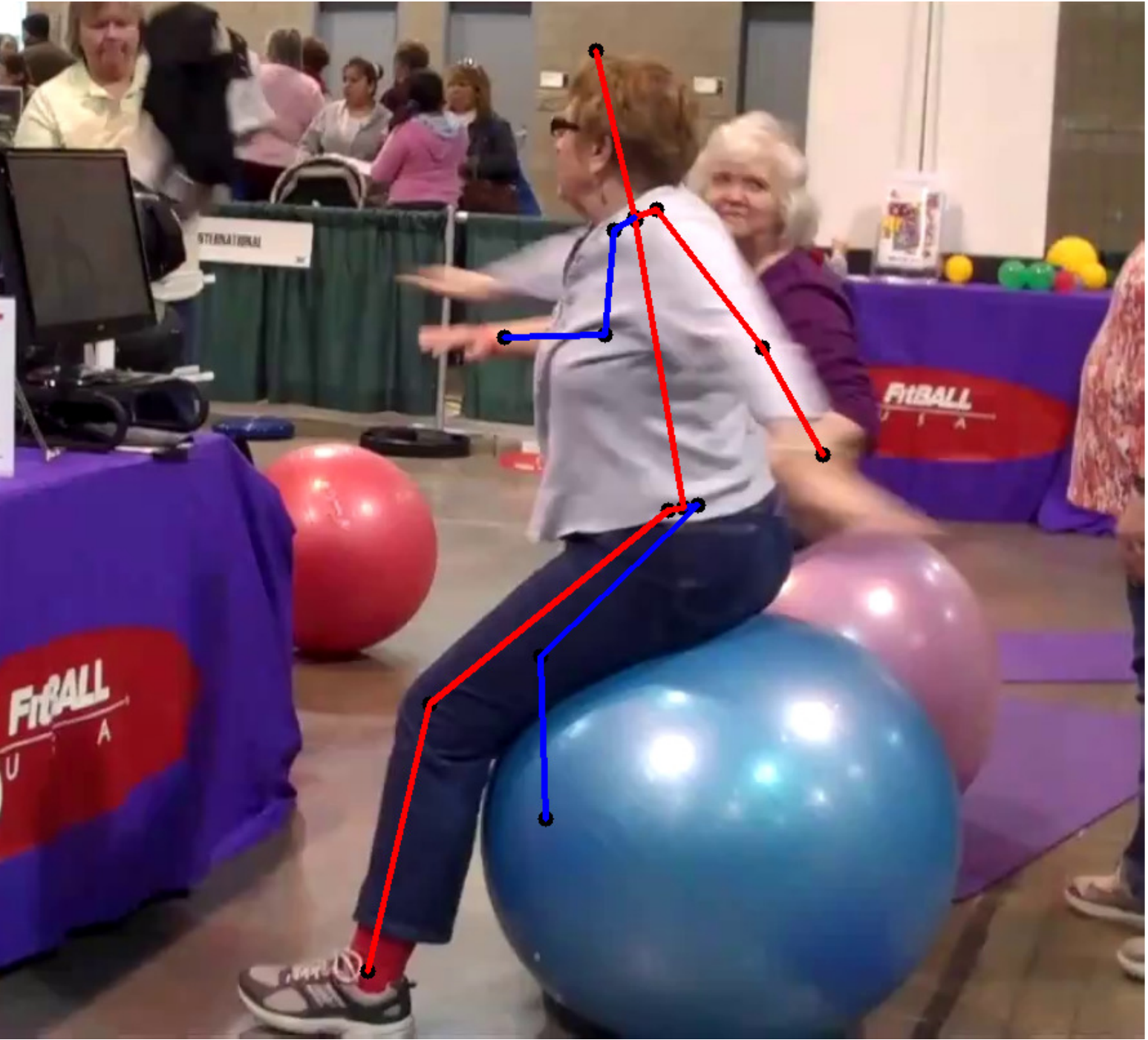} & \includegraphics[keepaspectratio=true, scale = 0.11]{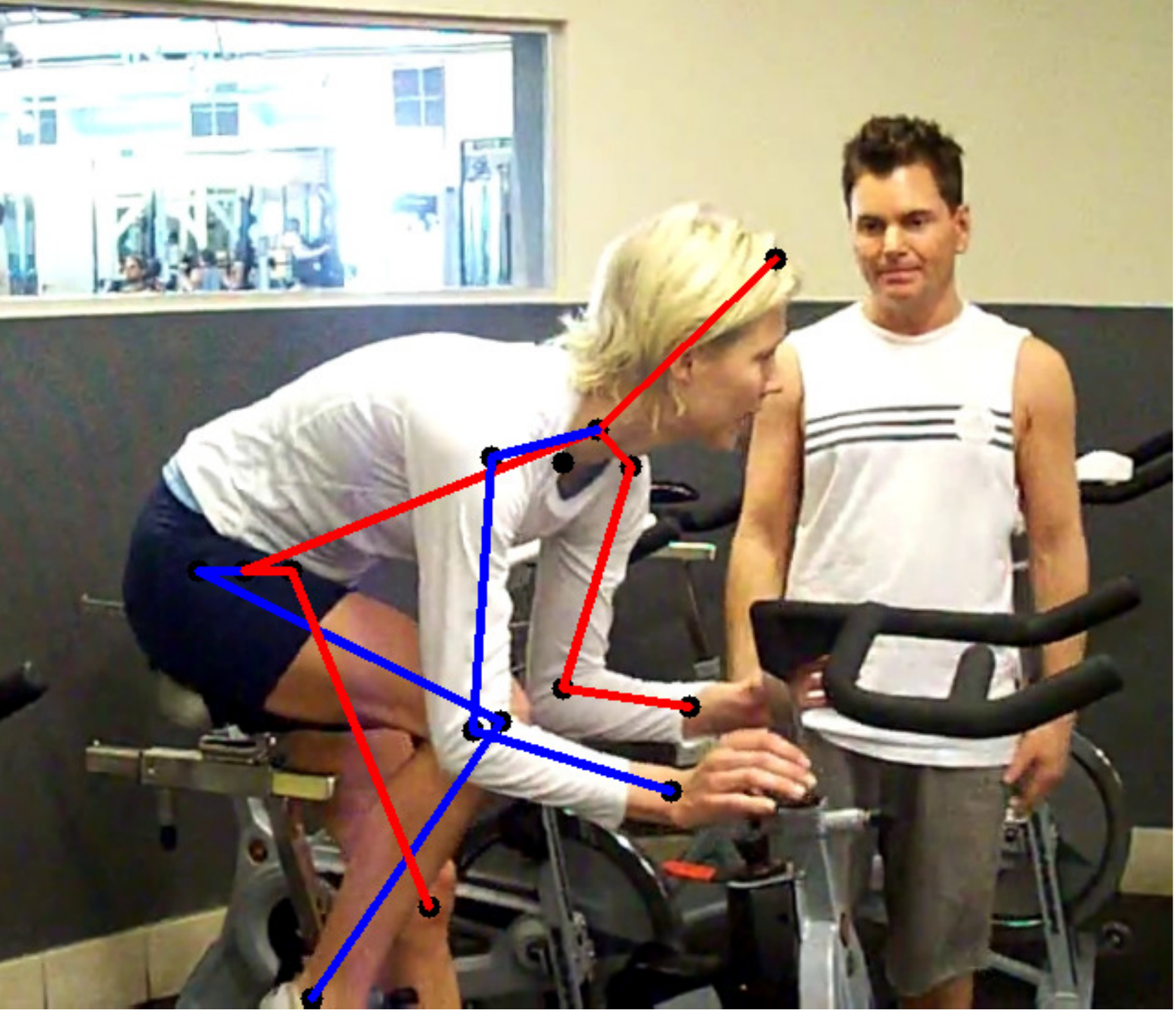} & \includegraphics[keepaspectratio=true, scale = 0.13]{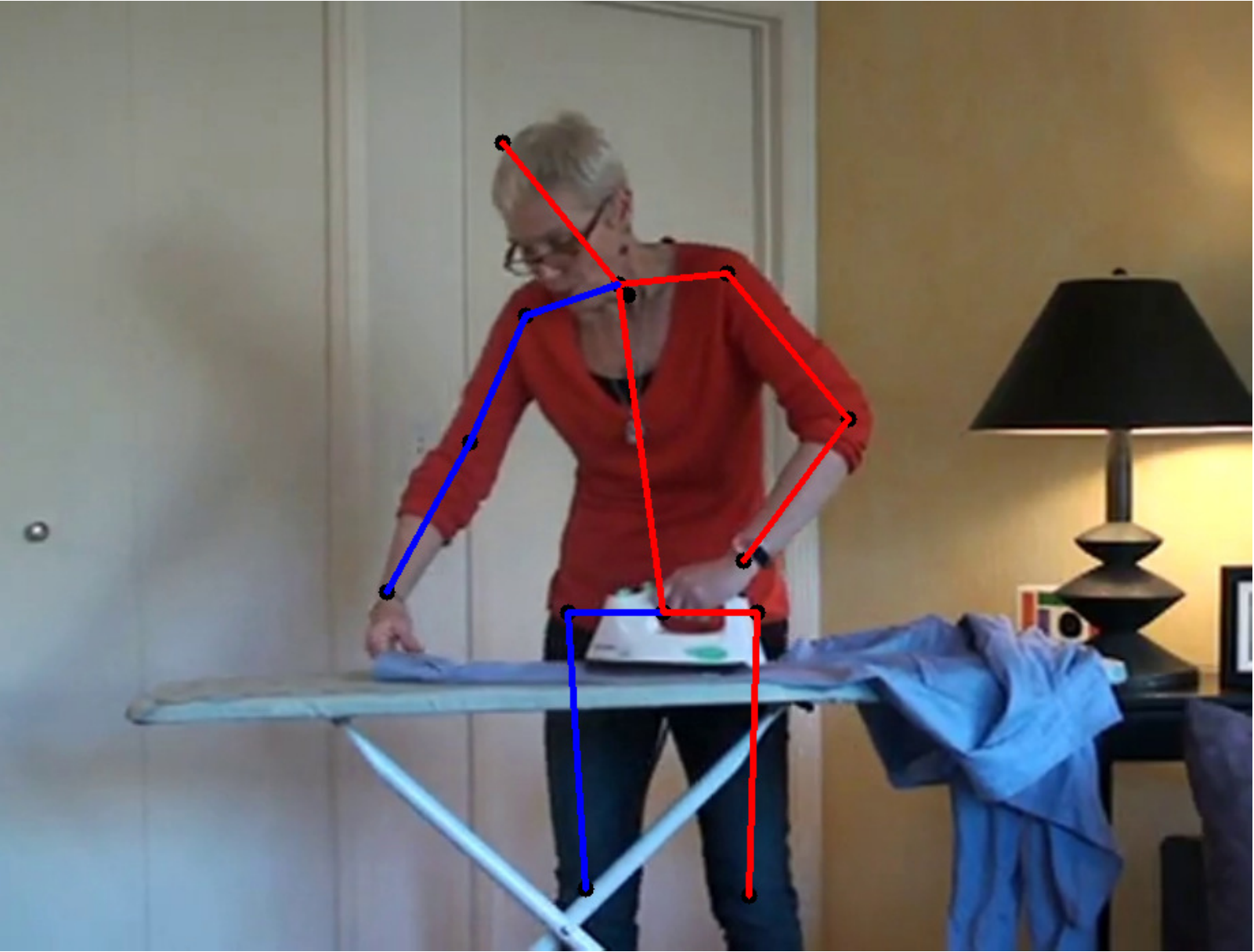} & \includegraphics[keepaspectratio=true, scale = 0.13]{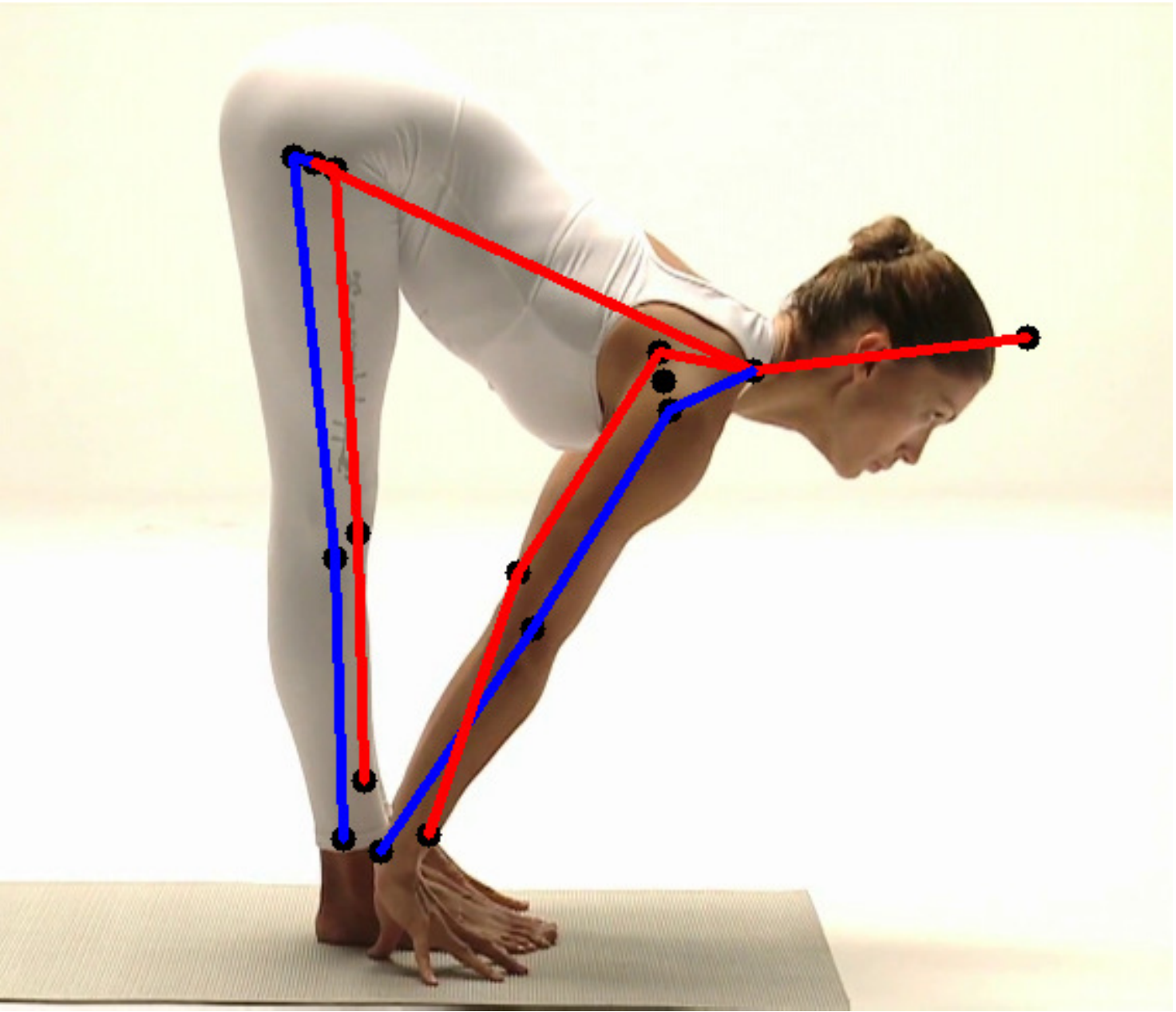}\\
            \includegraphics[keepaspectratio=true, scale = 0.145]{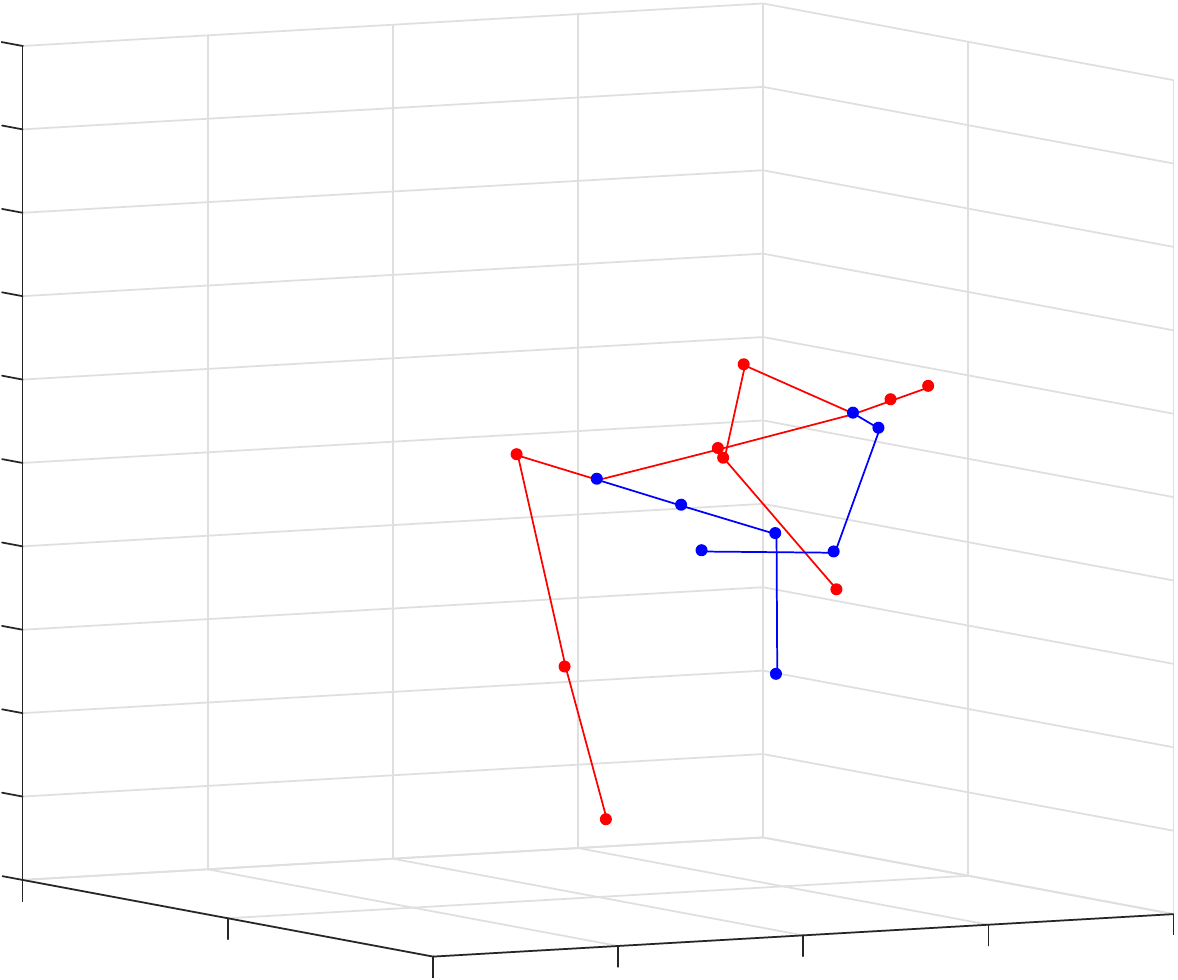} & \includegraphics[keepaspectratio=true, scale = 0.15]{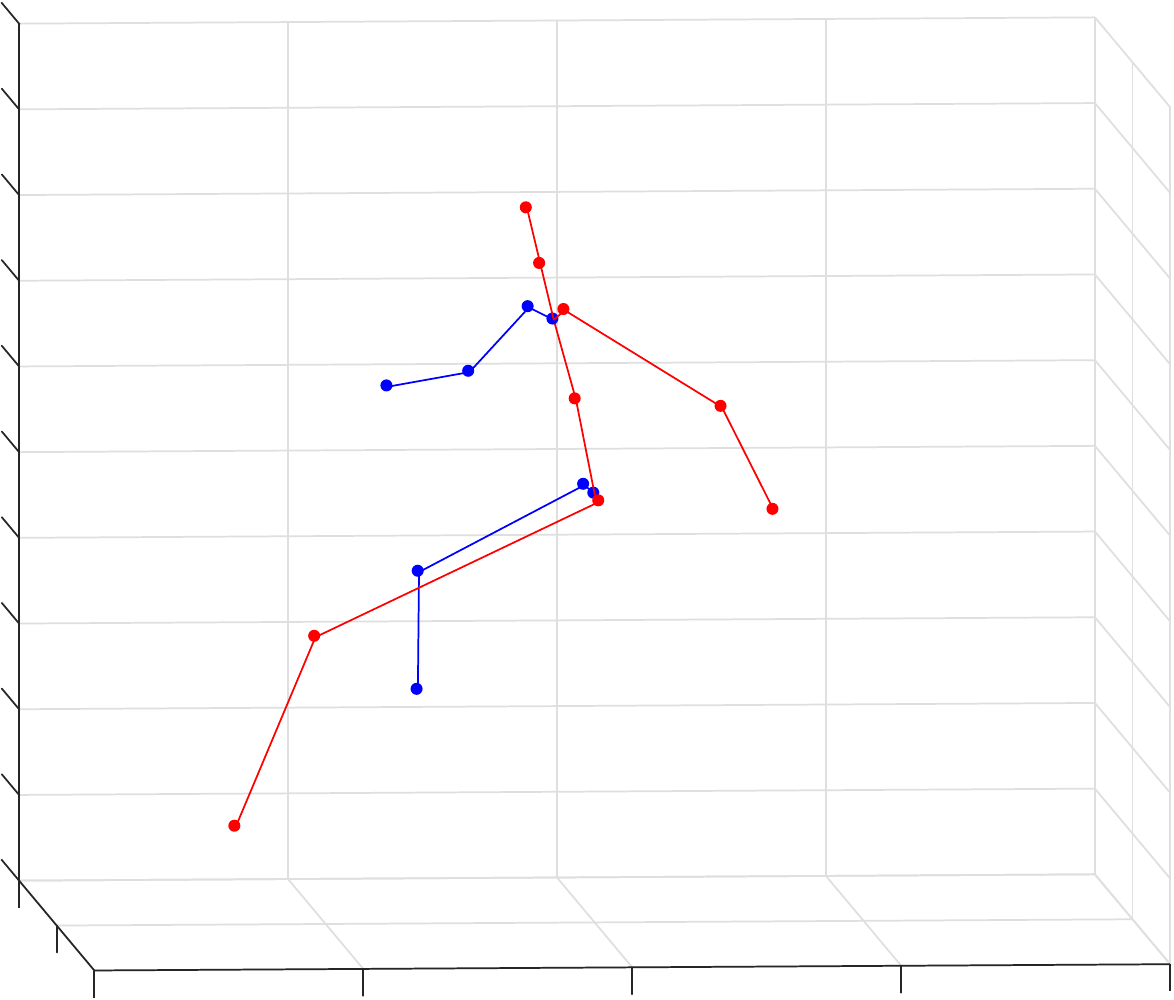} & \includegraphics[keepaspectratio=true, scale = 0.15]{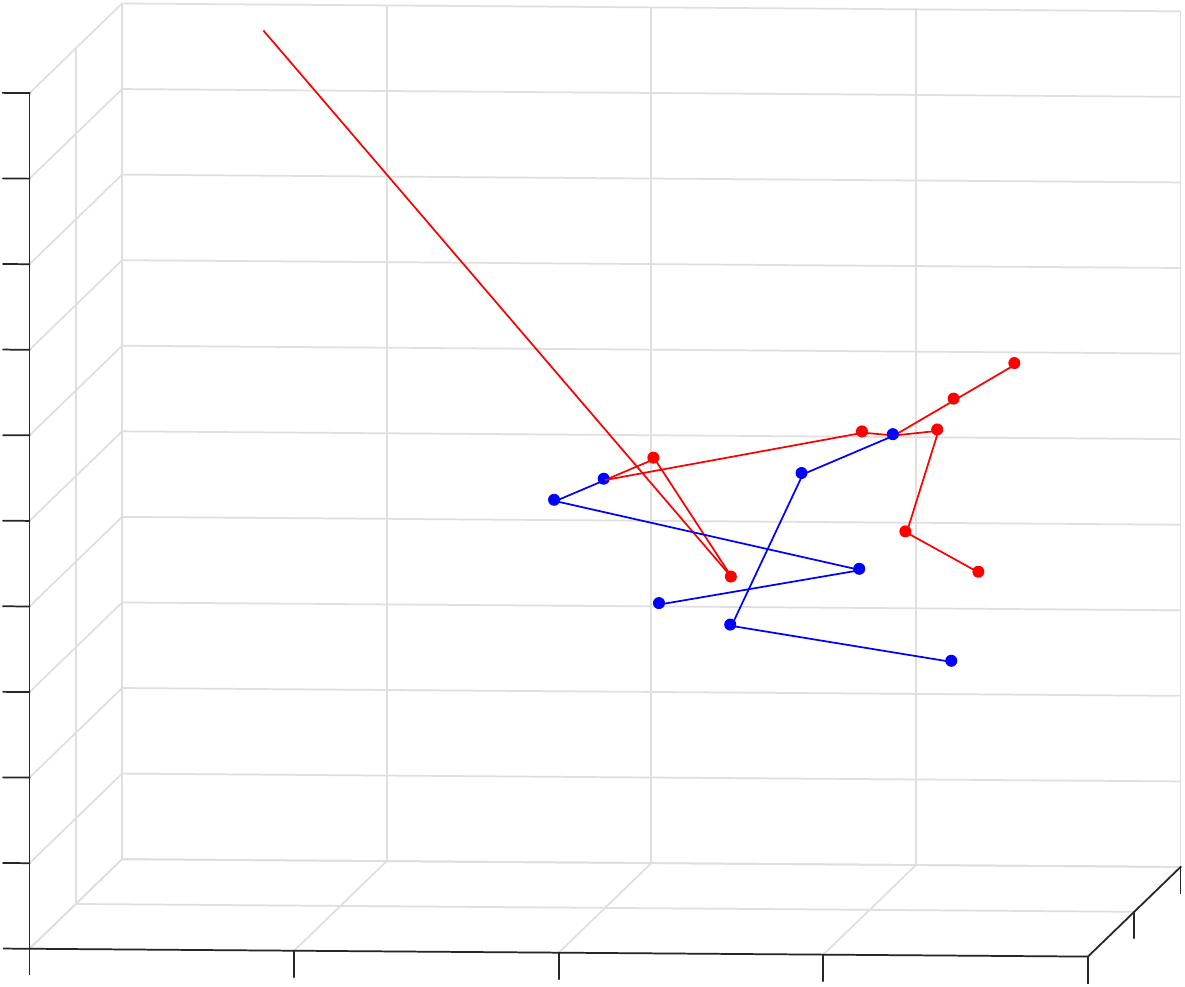} & \includegraphics[keepaspectratio=true, scale = 0.16]{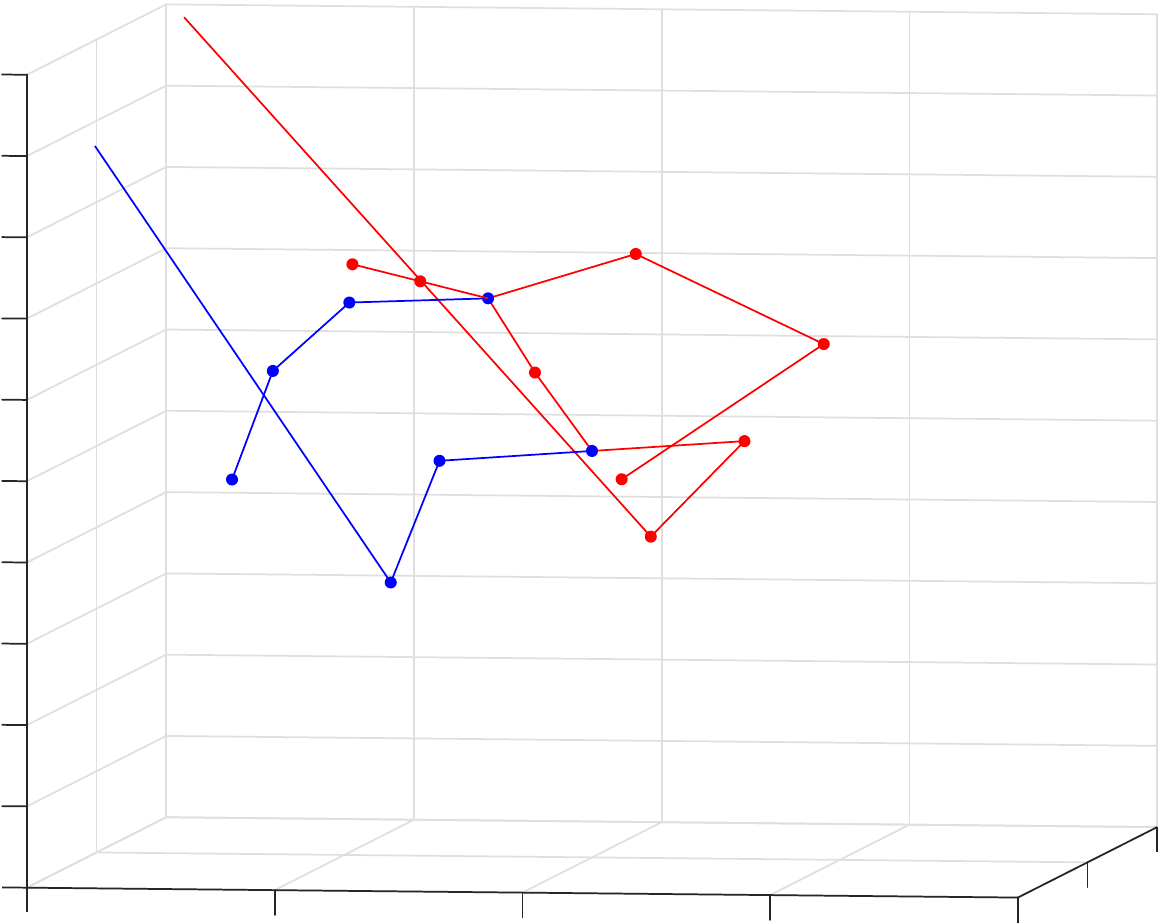}& \includegraphics[keepaspectratio=true, scale = 0.16]{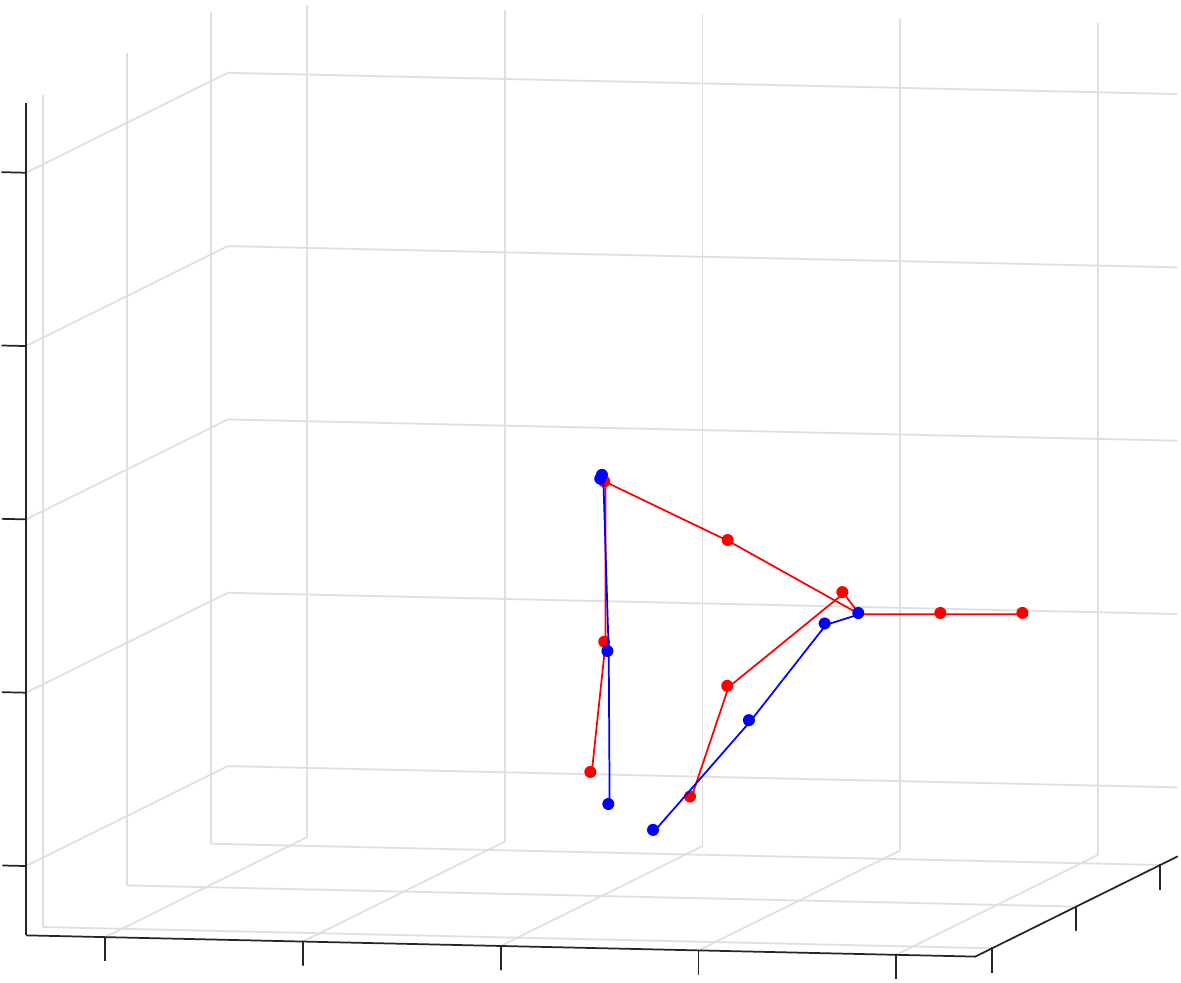} \\
            \includegraphics[keepaspectratio=true, scale = 0.145]{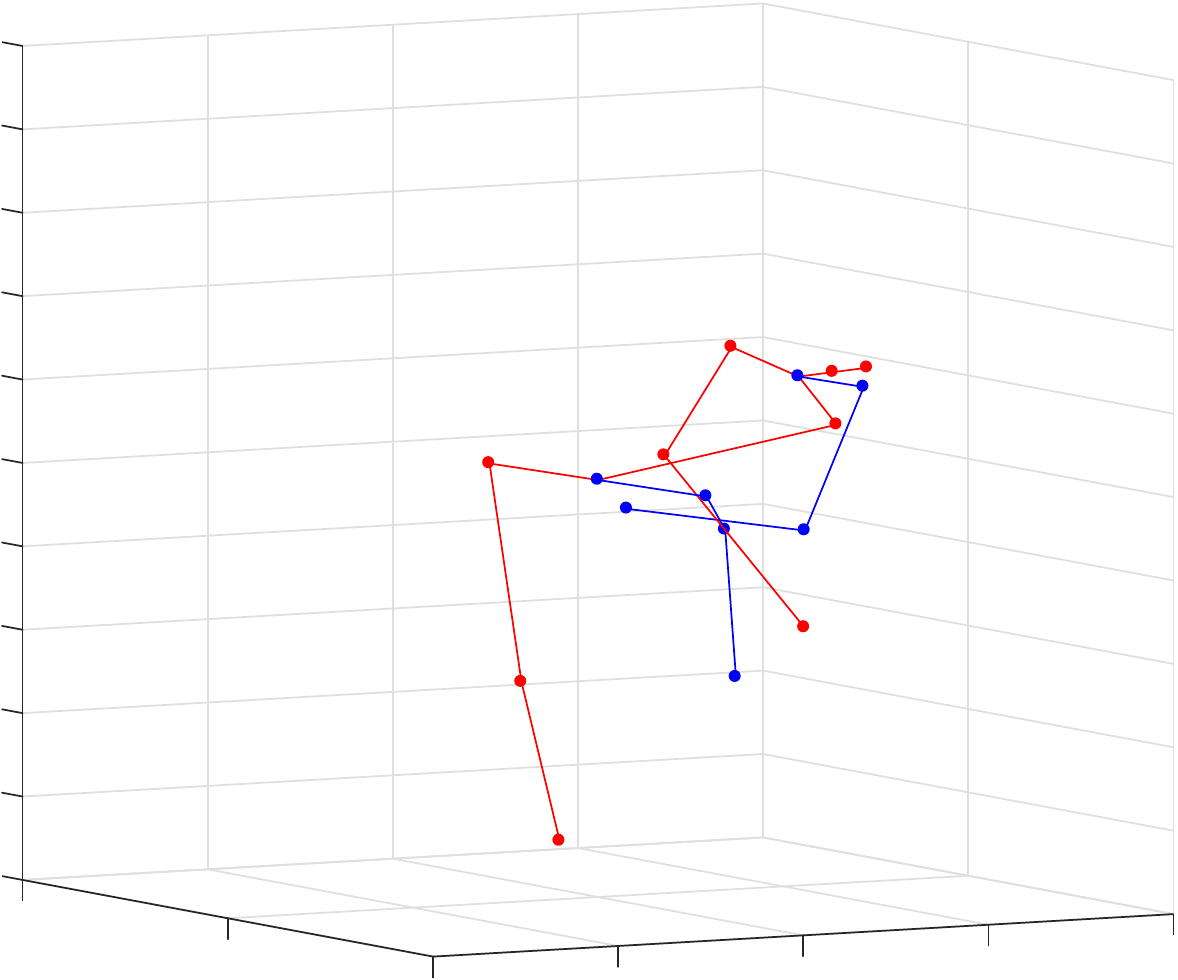} & \includegraphics[keepaspectratio=true, scale = 0.15]{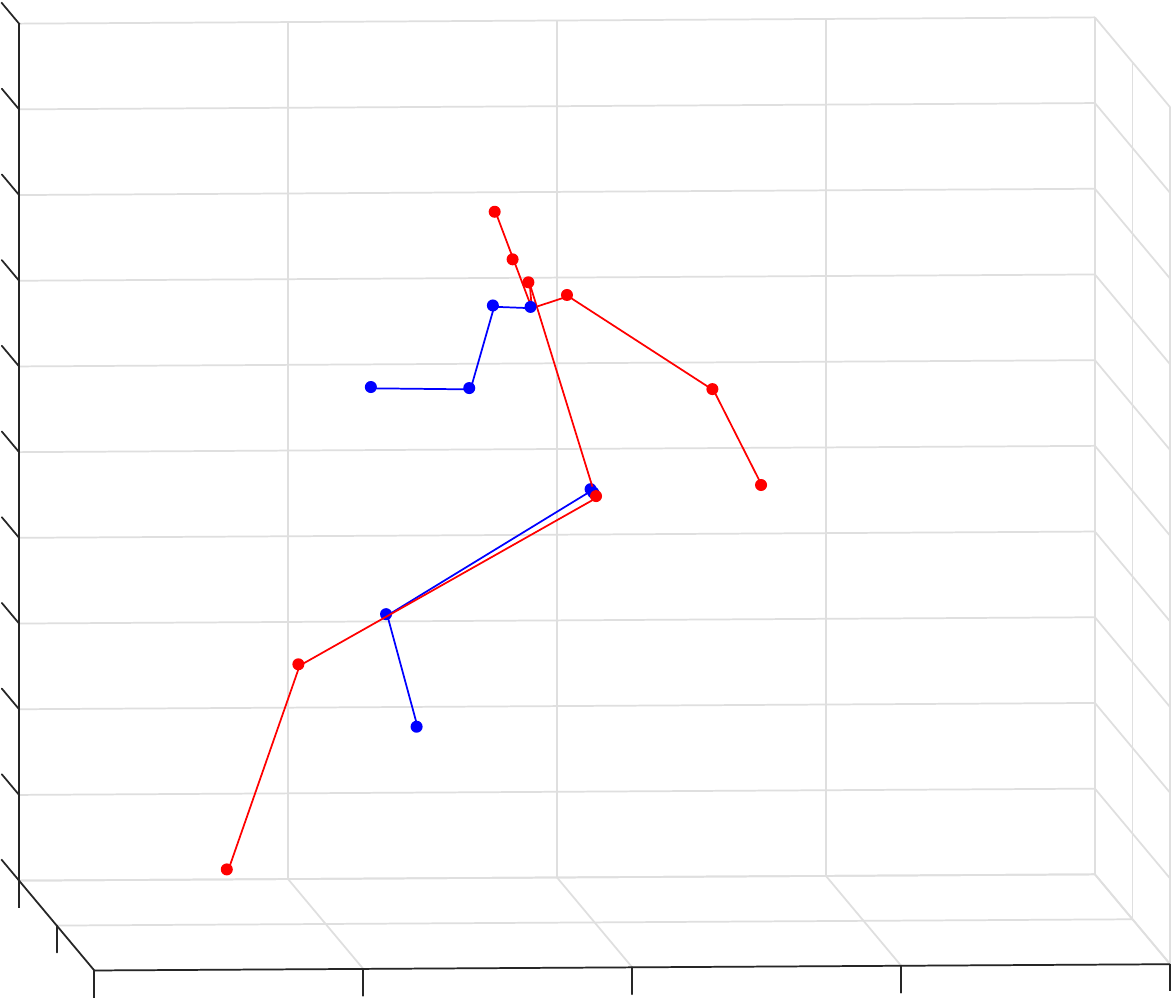} & \includegraphics[keepaspectratio=true, scale = 0.15]{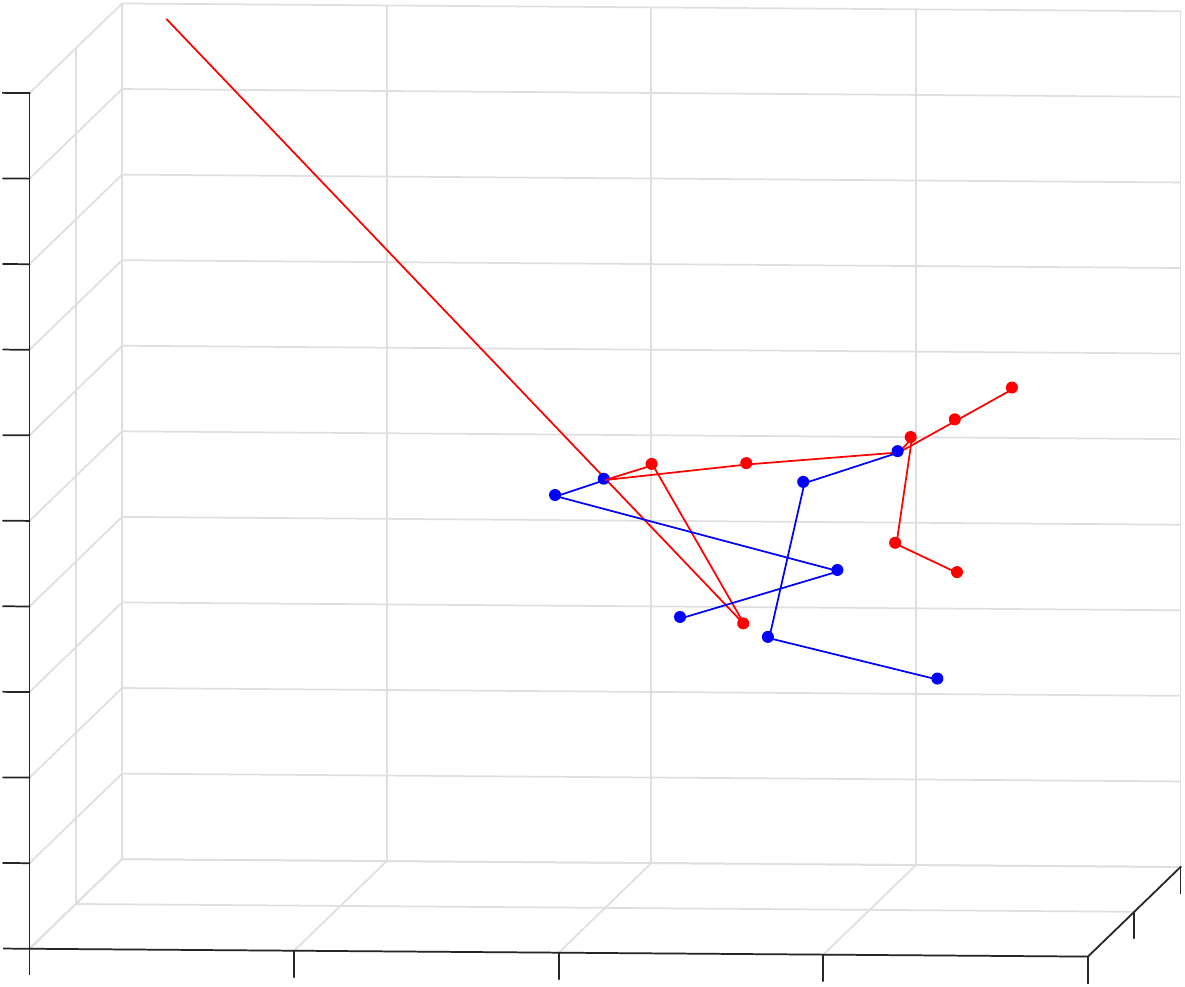} & \includegraphics[keepaspectratio=true, scale = 0.16]{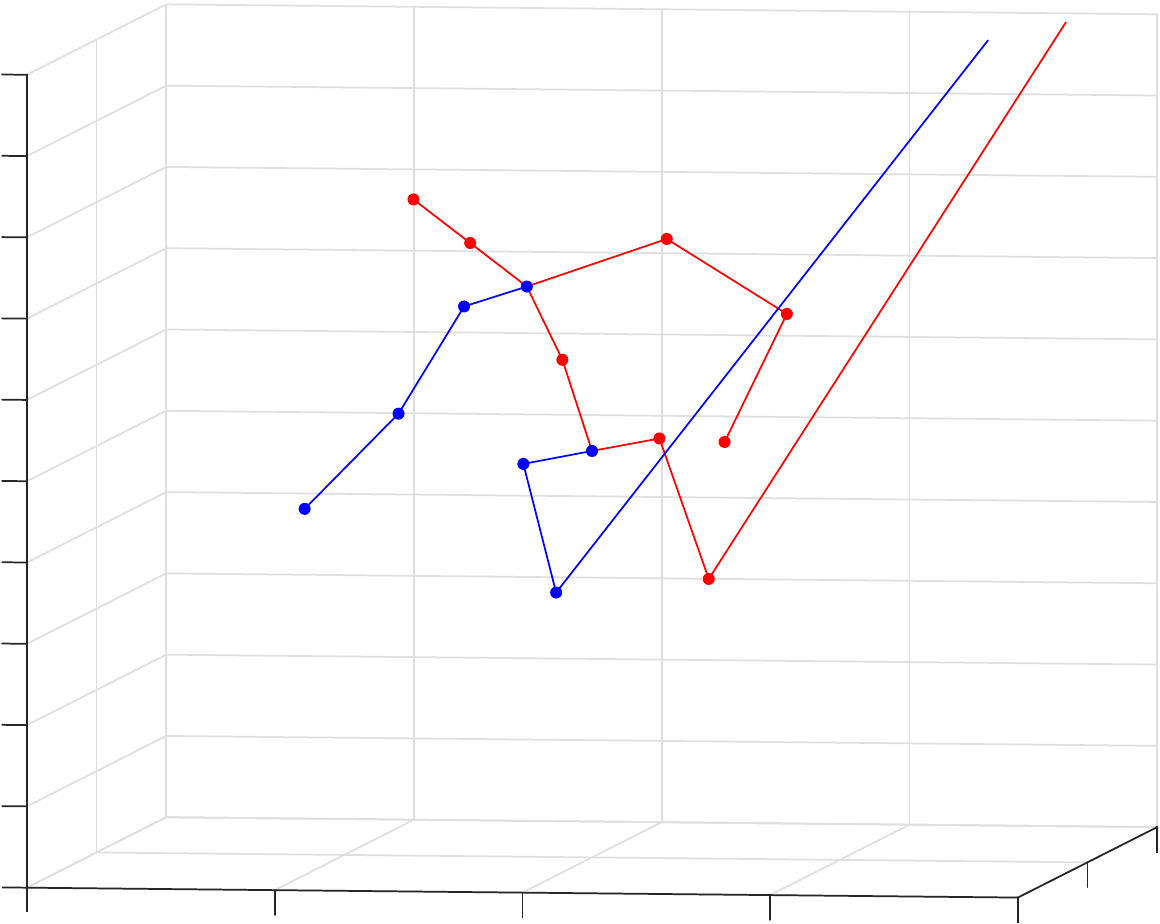}& \includegraphics[keepaspectratio=true, scale = 0.16]{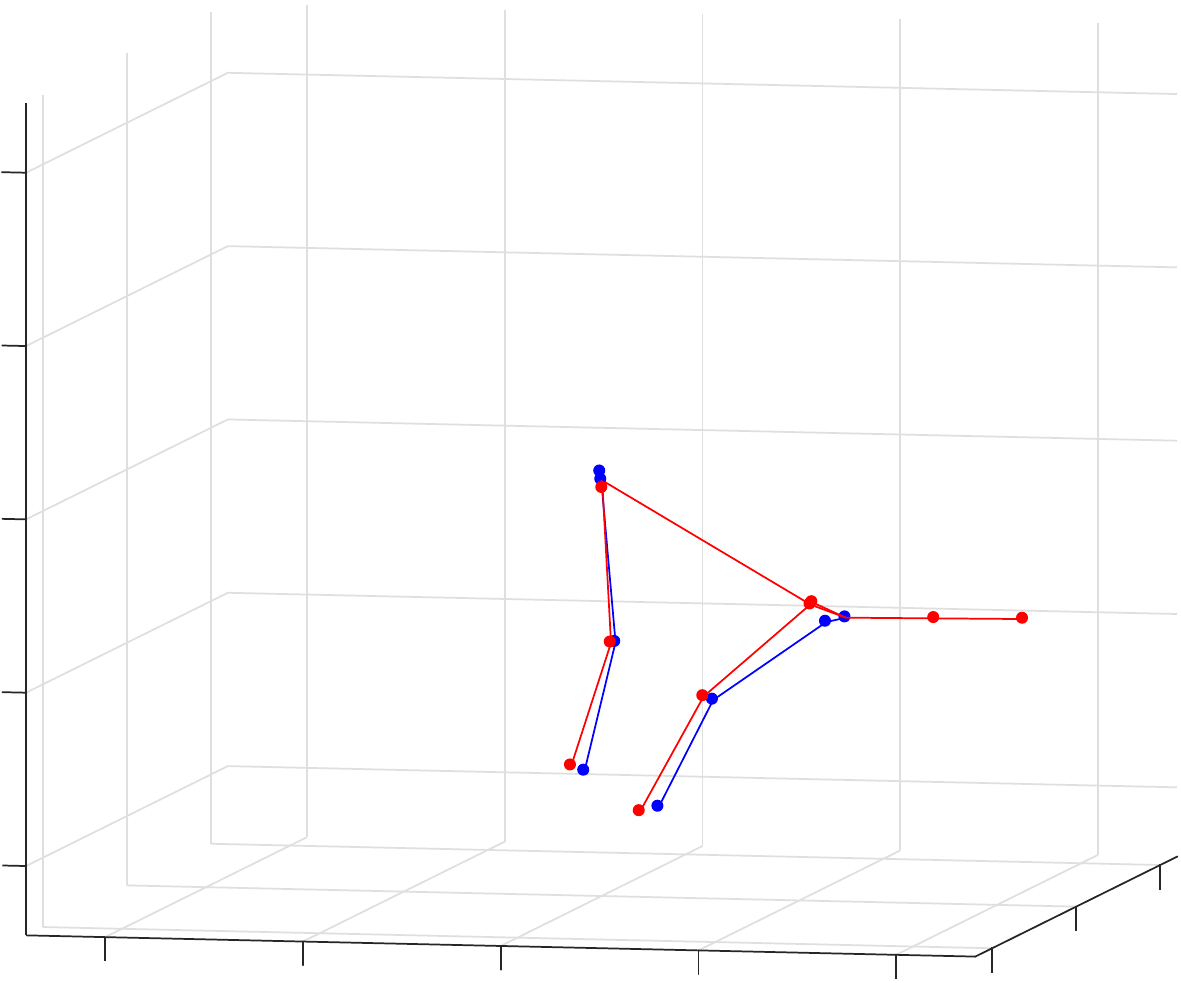} \\
	    \includegraphics[keepaspectratio=true, scale = 0.145]{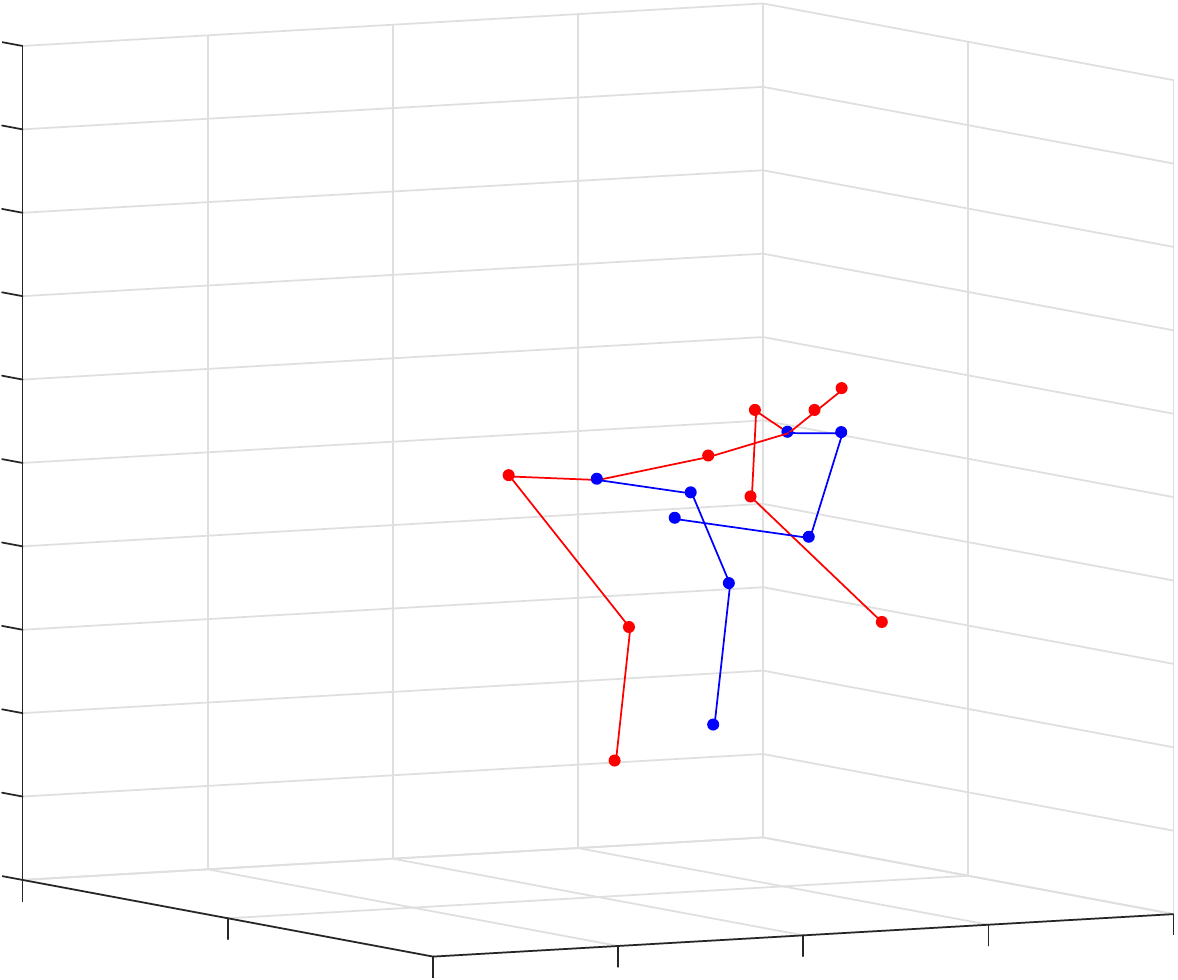} & \includegraphics[keepaspectratio=true, scale = 0.15]{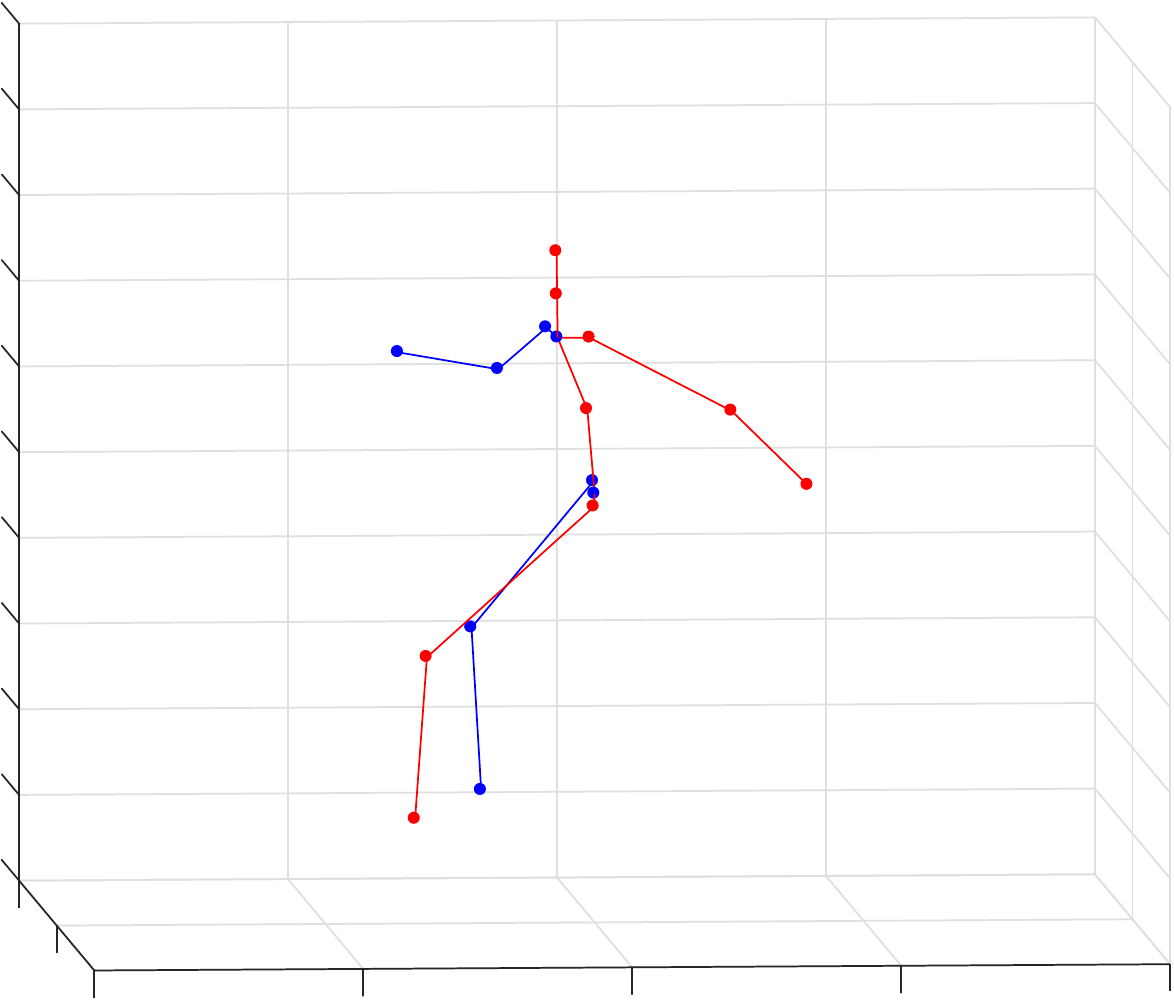} & \includegraphics[keepaspectratio=true, scale = 0.15]{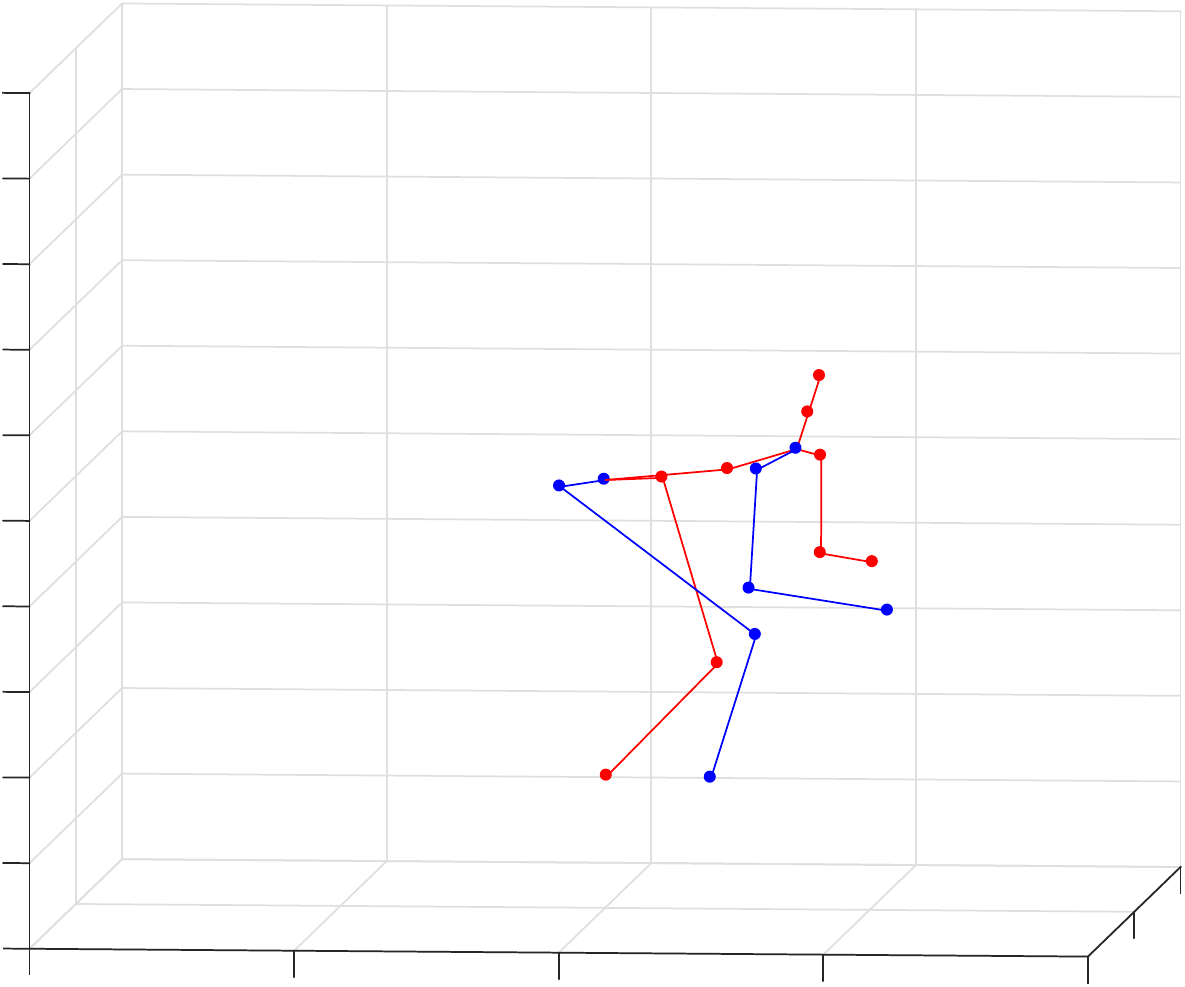} & \includegraphics[keepaspectratio=true, scale = 0.16]{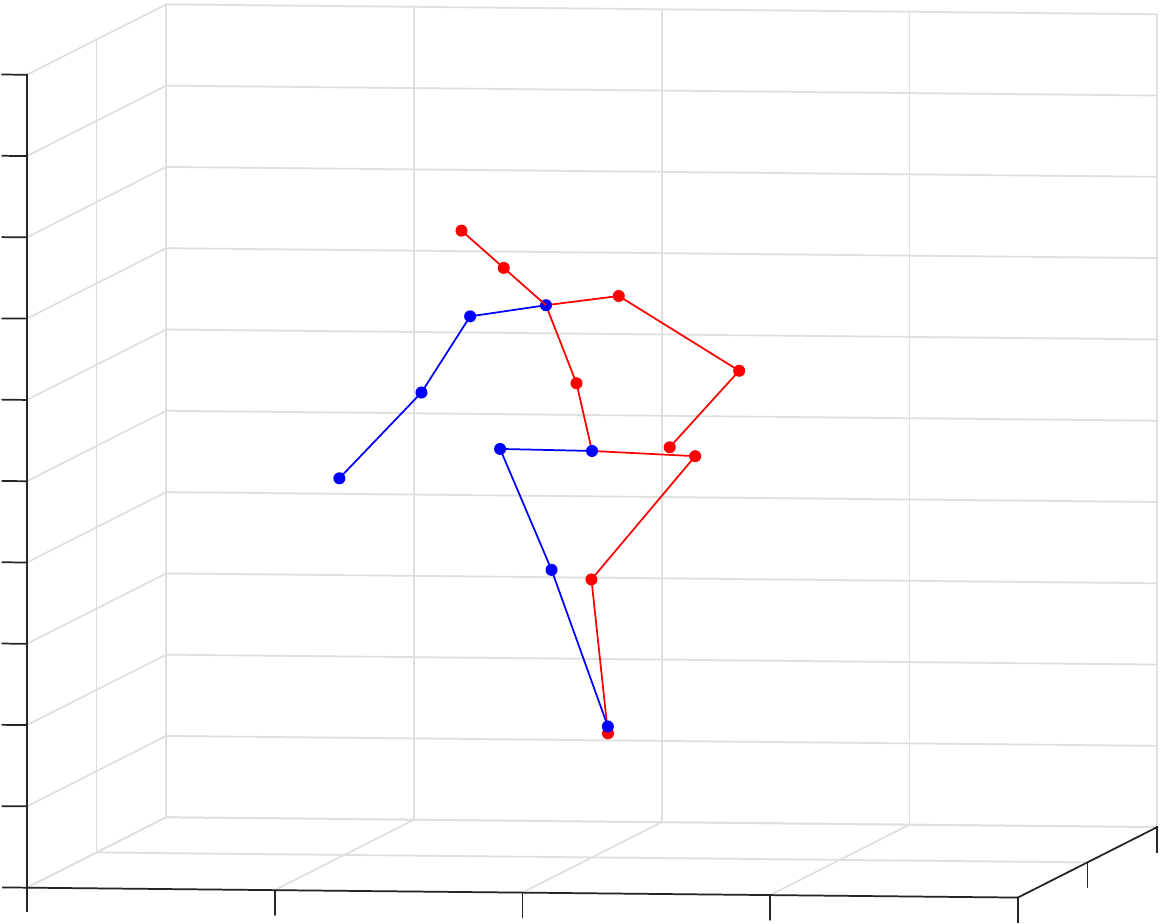}&\includegraphics[keepaspectratio=true, scale = 0.16]{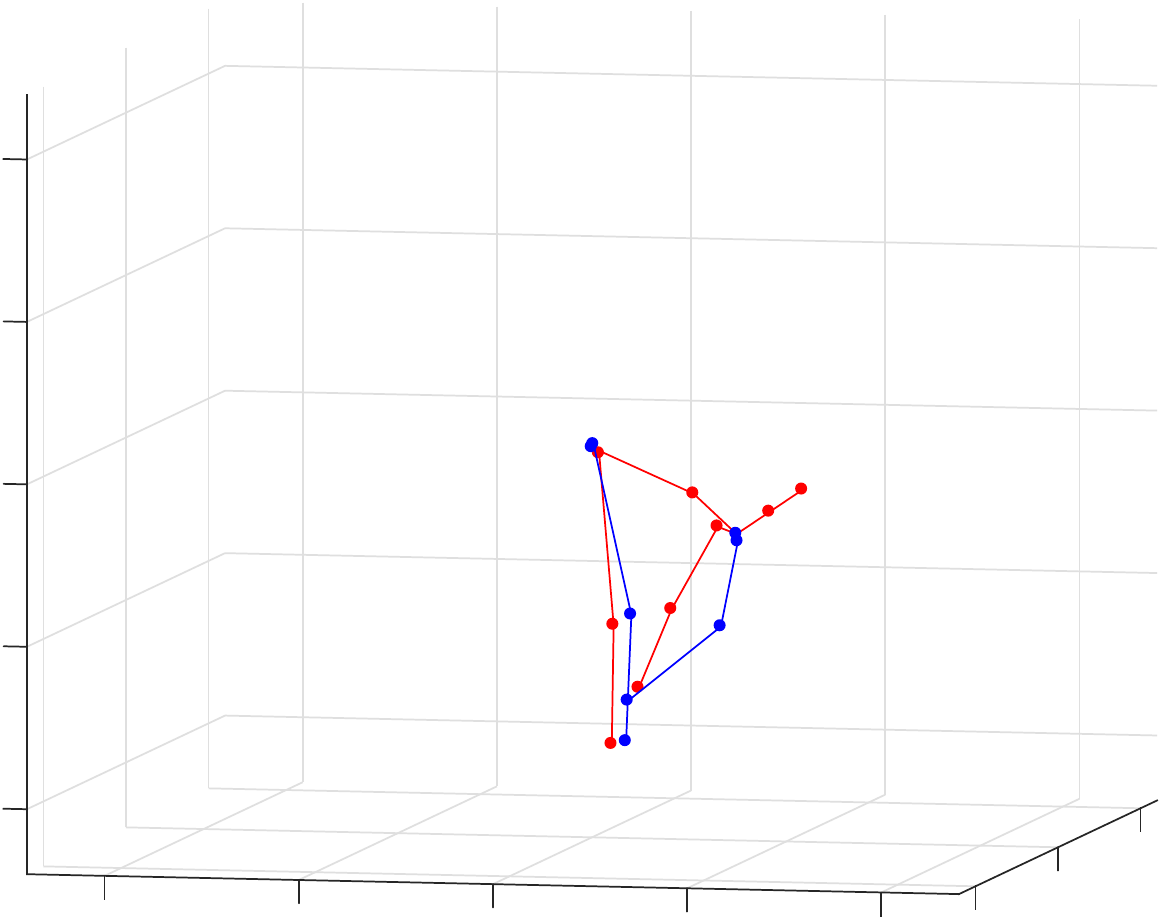} \\
	    (a)&(b)&(c)&(d)&(e)\\
            \end{tabular}
        \caption{Qualitative evaluation of models on MPII dataset. \textbf{First} row: Input images with ground 2d pose. \textbf{Second} row: 3d pose prediction of baseline architecture.
\textbf{Third} row: 3d pose prediction of proposed Model I (trained on Human3.6m dataset). \textbf{Fourth} row: 3d pose prediction of proposed Model II (Model I fine tuned on MPII dataset). A major drawback in the baseline network is, it can not capture poses with occluded or invisible joints. In Figure (c), left foot joint of the person is not visible, hence 2d annotation for this joint is absent. The baseline model can not predict 3d position for this joint, while our proposed model fine-tuned on MPII dataset can predict position of this joint properly even from absent annotation. Similarly, in Figure (d), our model can perfectly predict un-annotated joints.}
        \label{fig:MPII images}
    \end{figure*}

The Qualitative results on Human3.6m, MPII and MPI-INF-3DHP are shown in Figure~\ref{fig:Human36m images}, \ref{fig:MPII images} and \ref{fig:MPII-INF images} respectively. Our fine-tuned model on MPII dataset (Model II) shows significant improvement over the baseline model, on poses where joints are not visible. Hence, joint training of our proposed network on 2d poses with occluded joints (partly annotated 2d pose) along with 3d ground truth enhances its ability to predict occluded poses correctly.

\begin{figure}[t!]
        \centering
        \setlength{\belowcaptionskip}{-5pt}
        \begin{tabular}{ccc}
            \includegraphics[keepaspectratio=true, scale = 0.15]{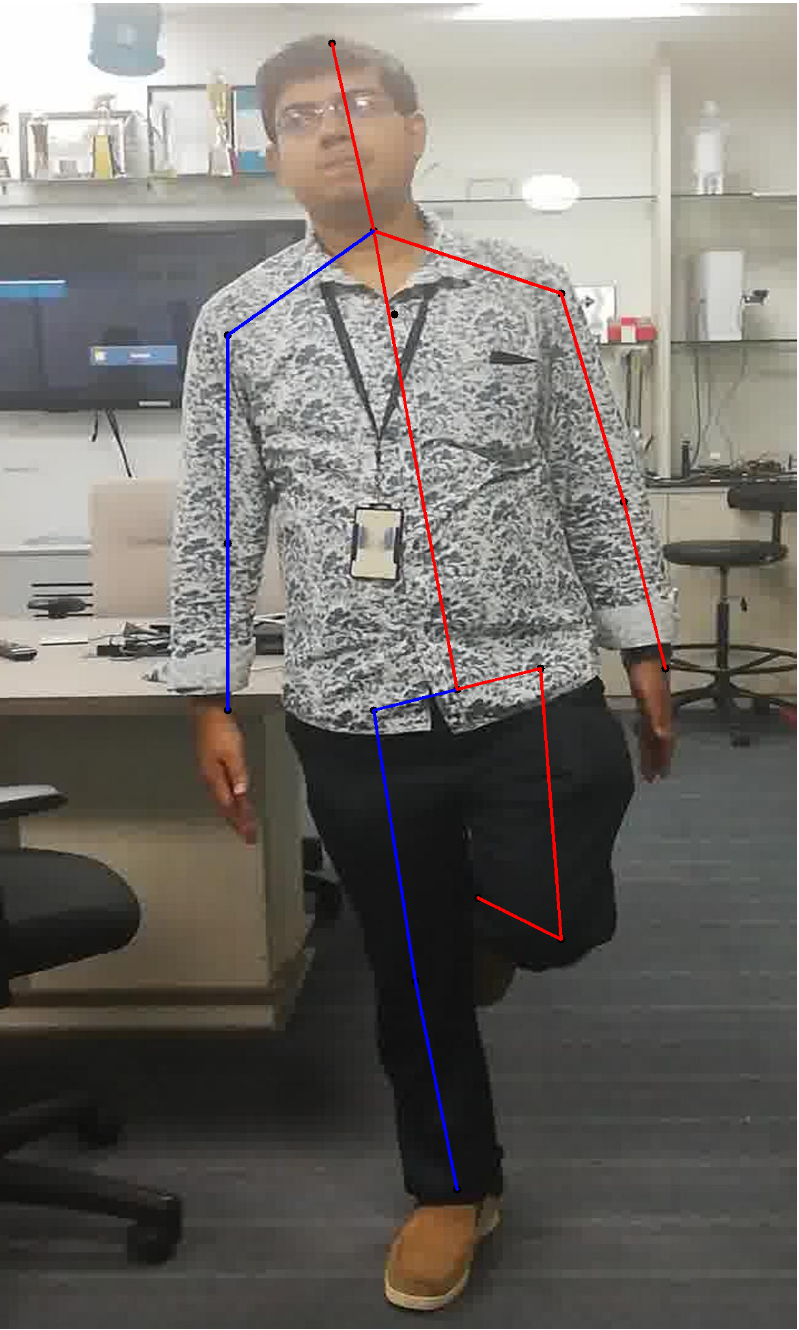} & \includegraphics[keepaspectratio=true, scale = 0.22]{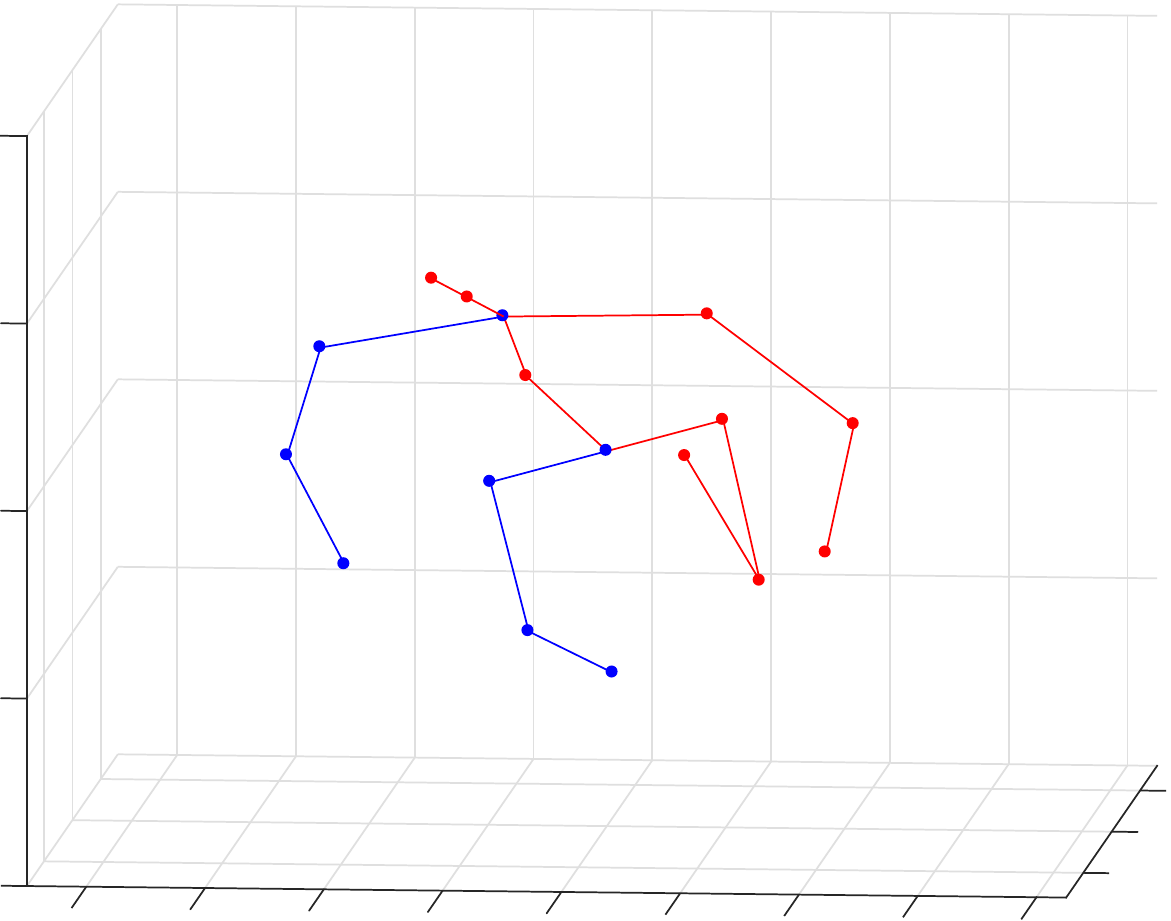} & \includegraphics[keepaspectratio=true, scale = 0.22]{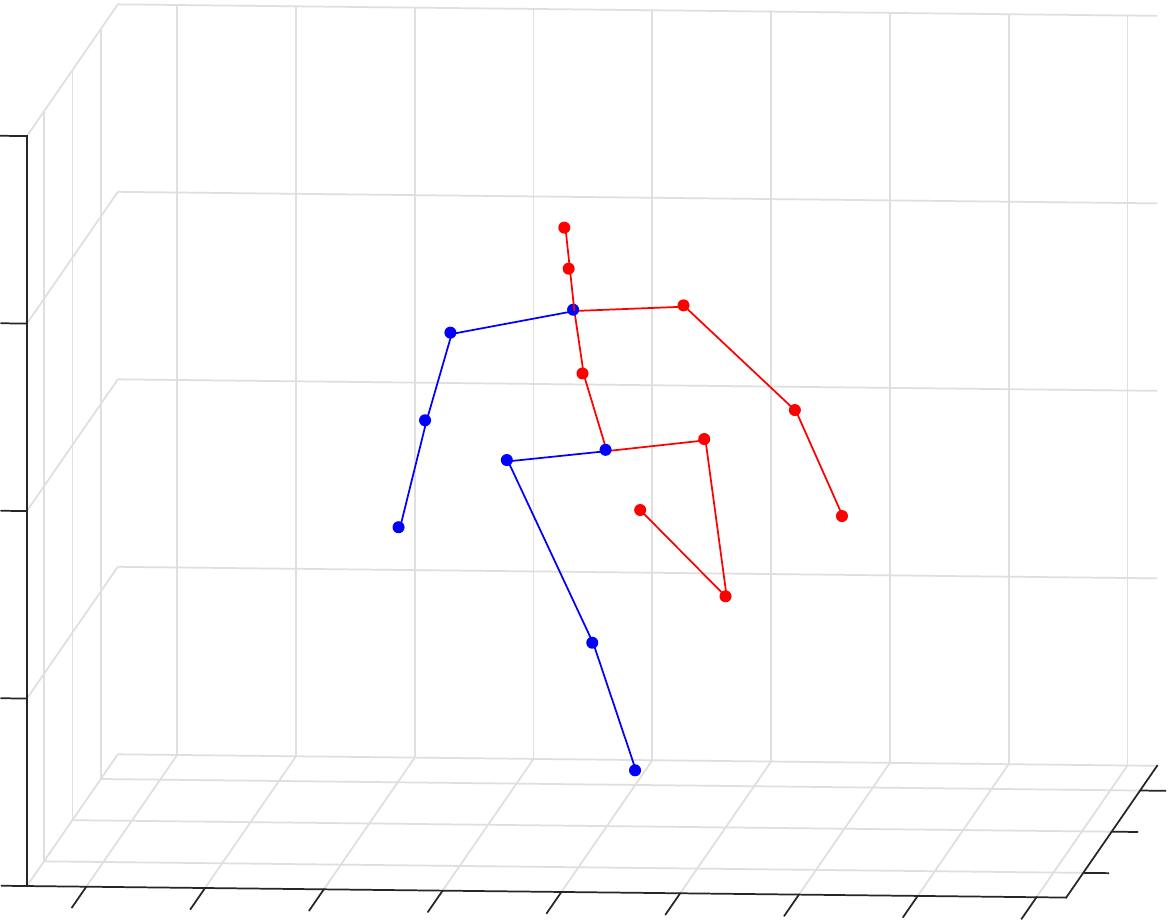} \\
                    \end{tabular}
        \caption{Performance evaluation of proposed model on data captured in our lab condition. \textbf{Left:} Input image with 2d pose (prediction of Stacked Hourglass network \cite{newell2016stacked}). \textbf{Middle:} Predicted 3d pose from baseline network \cite{martinez2017simple}. \textbf{Right:} Predicted 3d pose from our model (Model II). The baseline model fails to capture the proper angular positions of the legs and the overall pose appears to be bent forward.}
        \label{fig: lab data}
    \end{figure}

\textbf{Evaluation on our own dataset}: To further evaluate the generalization capability of our proposed model we have tested the models on our own dataset. A video data has been collected using a mobile camera in our lab environment. Stacked Hourglass network \cite{newell2016stacked} has been used to estimate the 2d poses, which are given as input to our Model. Figure~\ref{fig: lab data} shows a sample image, corresponding predicted 3d pose from the baseline network and predicted 3d pose from our proposed model. Our model gives better prediction (in terms of pose structure, angle between joints) of 3d pose compared to the baseline network.

\begin{table}[htbp]
\centering
	\begin{tabular}{|c|c|c|}
		\hline
	\textbf{Method} &\textbf{PCK} &\textbf{AUC}\\
	\hline
	2d-to-3d (supervised loss) \cite{martinez2017simple}&62.2&27.7\\	
    \hline
	\makecell{Ours 2d-to-3d+3d-to-2d \\ (re-projection loss)} & 64.2 &29.7\\
    \hline
    \makecell{Ours 2d-to-3d+3d-to-2d \\(re-projection loss+ bone symmetry loss)}  
	  &	65.8&31.2\\	
	    \hline
	\end{tabular}	
	\caption{Ablation Study for different losses on MPI-INF-3DHP dataset.}
	\label{table:ablationloss}
\end{table}

\begin{table}[htbp]
\centering
     \begin{tabular}{|c|c|c|}
\hline
\textbf{Method}& \textbf{Re-projection error}& $\Delta$ \\
\hline
w/o batch normalization & 36.2 & 30.7\\
\hline
w/o dropout & 6.49 & 0.99\\
\hline
w/o dropout + w/o batch normalization & 34.79 & 29.29\\
\hline
\end{tabular}
\caption{Ablation study on different network parameters for our (3d-to-2d re-projection module) in terms of re-projection error on Human3.6m. $\Delta$ defines re-projection error differences between current training setup (as mentioned in section \ref{Implementation details}) and the above setups.}
	\label{table:ablationnetwork}
\end{table}

\subsection{Ablation Study:}
Table~\ref{table:ablationloss} and Table~\ref{table:ablationnetwork} shows ablative analysis of different network design parameters and losses used during training. 
Table~\ref{table:ablationloss} shows, the addition of 2d re-projection loss with the supervised 3d loss in baseline network, increases PCK by 3.2\% and AUC by 7.2\% on MPI-INF-3DHP dataset, during cross-dataset validation. Using bone length symmetry loss with re-projection and supervised loss advances network performance further with 6\% and 13\% of improvement in PCK and AUC respectively for similar test-setup. 

For 2d-to-3d and 3d-to-2d module we have used similar architecture as the baseline network. To understand the optimality of performance of our 3d-to-2d module we have performed an ablation study on different choice of design parameters. Table~\ref{table:ablationnetwork}, represents error between input ground truth 2d and re-projected 2d from the 3d-to-2d module for various design choices of the network. This error is measured in terms of Euclidean distance between joints in 2d co-ordinate space. The re-projection error is quite high when the network is trained without dropout or batch normalization between intermediate layers. Hence, the 3d-to-2d module is also trained using batch normalization and dropout similar to the 2d-to-3d module. $\Delta$ defines the re-projection error differences between current training setup and different choices of training setups as mentioned in Table~\ref{table:ablationnetwork}.

\section{Conclusion}
In this paper, we propose a deep neural network for estimating 3d human pose from 2d pose that combines data from 2d pose datasets in the wild and 3d pose datasets captured in controlled environments in a weakly-supervised framework. Our 3d-to-2d re-projection network is necessary for generalization of 3d pose estimation as it learns to predict 3d poses from in-the-wild 2d pose annotations that do not contain 3d pose annotations. Our method outperforms current state-of-the-art methods on a benchmark 3d dataset Human3.6m captured in controlled environments, as well as a challenging 3d dataset MPI-INF-3DHP containing in the wild human poses. Along with benchmark datasets, we also demonstrate the generalization ability of our method on our own dataset. As a future direction, we aim to improve our method by introducing more geometric constraints based on human anatomy, to improve the accuracy of 3d pose estimation in the absence of 3d ground-truth.
\bibliographystyle{IEEEbib}
\bibliography{strings,refs}

\begin{thebibliography}{10}

\bibitem{upperbody}
Sanjana Sinha, Brojeshwar Bhowmick, Kingshuk Chakravarty, Aniruddha Sinha, and
  Abhijit Das,
\newblock ``Accurate upper body rehabilitation system using kinect,''
\newblock in {\em Annual International Conference of the IEEE Engineering in
  Medicine and Biology Society}, 2016, pp. 4605--4609.

\bibitem{SLS}
Kingshuk Chakravarty, Suraj Suman, Brojeshwar Bhowmick, Aniruddha Sinha, and
  Abhijit Das,
\newblock ``Quantification of balance in single limb stance using kinect,''
\newblock in {\em International Conference on Acoustics, Speech and Signal
  Processing}, 2016, pp. 854--858.

\bibitem{mehta2017monocular}
Dushyant Mehta, Helge Rhodin, Dan Casas, Pascal Fua, Oleksandr Sotnychenko,
  Weipeng Xu, and Christian Theobalt,
\newblock ``Monocular 3d human pose estimation in the wild using improved cnn
  supervision,''
\newblock in {\em 3D Vision (3DV), 2017 International Conference on}. IEEE,
  2017, pp. 506--516.

\bibitem{pavlakos2017coarse}
Georgios Pavlakos, Xiaowei Zhou, Konstantinos~G Derpanis, and Kostas
  Daniilidis,
\newblock ``Coarse-to-fine volumetric prediction for single-image 3d human
  pose,''
\newblock in {\em Computer Vision and Pattern Recognition (CVPR), 2017 IEEE
  Conference on}. IEEE, 2017, pp. 1263--1272.

\bibitem{zhou2017towards}
Xingyi Zhou, Qixing Huang, Xiao Sun, Xiangyang Xue, and Yichen Wei,
\newblock ``Towards 3d human pose estimation in the wild: a weakly-supervised
  approach,''
\newblock in {\em IEEE International Conference on Computer Vision}, 2017.

\bibitem{luo2018orinet}
Chenxu Luo, Xiao Chu, and Alan Yuille,
\newblock ``Orinet: A fully convolutional network for 3d human pose
  estimation,''
\newblock {\em arXiv preprint arXiv:1811.04989}, 2018.

\bibitem{tome2017lifting}
Denis Tome, Chris Russell, and Lourdes Agapito,
\newblock ``Lifting from the deep: Convolutional 3d pose estimation from a
  single image,''
\newblock in {\em Proceedings of the IEEE Conference on Computer Vision and
  Pattern Recognition}, 2017, pp. 2500--2509.

\bibitem{yang20183d}
Wei Yang, Wanli Ouyang, Xiaolong Wang, Jimmy Ren, Hongsheng Li, and Xiaogang
  Wang,
\newblock ``3d human pose estimation in the wild by adversarial learning,''
\newblock in {\em Proceedings of the IEEE Conference on Computer Vision and
  Pattern Recognition}, 2018, vol.~1.

\bibitem{lee2018propagating}
Kyoungoh Lee, Inwoong Lee, and Sanghoon Lee,
\newblock ``Propagating lstm: 3d pose estimation based on joint
  interdependency,''
\newblock in {\em Proceedings of the European Conference on Computer Vision
  (ECCV)}, 2018, pp. 119--135.

\bibitem{lin2017recurrent}
Mude Lin, Liang Lin, Xiaodan Liang, Keze Wang, and Hui Cheng,
\newblock ``Recurrent 3d pose sequence machines,''
\newblock in {\em Computer Vision and Pattern Recognition (CVPR), 2017 IEEE
  Conference on}. IEEE, 2017, pp. 5543--5552.

\bibitem{hossain2018exploiting}
Mir Rayat~Imtiaz Hossain and James~J Little,
\newblock ``Exploiting temporal information for 3d human pose estimation,''
\newblock in {\em European Conference on Computer Vision}. Springer, Cham,
  2018, pp. 69--86.

\bibitem{zhou2016sparseness}
Xiaowei Zhou, Menglong Zhu, Spyridon Leonardos, Konstantinos~G Derpanis, and
  Kostas Daniilidis,
\newblock ``Sparseness meets deepness: 3d human pose estimation from monocular
  video,''
\newblock in {\em Proceedings of the IEEE conference on computer vision and
  pattern recognition}, 2016, pp. 4966--4975.

\bibitem{dabral2018learning}
Rishabh Dabral, Anurag Mundhada, Uday Kusupati, Safeer Afaque, Abhishek Sharma,
  and Arjun Jain,
\newblock ``Learning 3d human pose from structure and motion,''
\newblock in {\em Proceedings of the European Conference on Computer Vision
  (ECCV)}, 2018, pp. 668--683.

\bibitem{newell2016stacked}
Alejandro Newell, Kaiyu Yang, and Jia Deng,
\newblock ``Stacked hourglass networks for human pose estimation,''
\newblock in {\em European Conference on Computer Vision}. Springer, 2016, pp.
  483--499.

\bibitem{cao2017realtime}
Zhe Cao, Tomas Simon, Shih-En Wei, and Yaser Sheikh,
\newblock ``Realtime multi-person 2d pose estimation using part affinity
  fields,''
\newblock in {\em Proceedings of the IEEE Conference on Computer Vision and
  Pattern Recognition}, 2017, pp. 7291--7299.

\bibitem{andriluka14cvpr}
Mykhaylo Andriluka, Leonid Pishchulin, Peter Gehler, and Bernt Schiele,
\newblock ``2d human pose estimation: New benchmark and state of the art
  analysis,''
\newblock in {\em IEEE Conference on Computer Vision and Pattern Recognition
  (CVPR)}, June 2014.

\bibitem{sigal2010humaneva}
Leonid Sigal, Alexandru~O Balan, and Michael~J Black,
\newblock ``Humaneva: Synchronized video and motion capture dataset and
  baseline algorithm for evaluation of articulated human motion,''
\newblock {\em International journal of computer vision}, vol. 87, no. 1-2, pp.
  4, 2010.

\bibitem{h36m_pami}
Catalin Ionescu, Dragos Papava, Vlad Olaru, and Cristian Sminchisescu,
\newblock ``Human3.6m: Large scale datasets and predictive methods for 3d human
  sensing in natural environments,''
\newblock {\em IEEE Transactions on Pattern Analysis and Machine Intelligence},
  vol. 36, no. 7, pp. 1325--1339, jul 2014.

\bibitem{martinez2017simple}
Julieta Martinez, Rayat Hossain, Javier Romero, and James~J Little,
\newblock ``A simple yet effective baseline for 3d human pose estimation,''
\newblock in {\em International Conference on Computer Vision}, 2017, vol.~1,
  p.~5.

\bibitem{mehta2017vnect}
Dushyant Mehta, Srinath Sridhar, Oleksandr Sotnychenko, Helge Rhodin, Mohammad
  Shafiei, Hans-Peter Seidel, Weipeng Xu, Dan Casas, and Christian Theobalt,
\newblock ``Vnect: Real-time 3d human pose estimation with a single rgb
  camera,''
\newblock {\em ACM Transactions on Graphics (TOG)}, vol. 36, no. 4, pp. 44,
  2017.

\bibitem{fang2018learning}
Hao-Shu Fang, Yuanlu Xu, Wenguan Wang, Xiaobai Liu, and Song-Chun Zhu,
\newblock ``Learning pose grammar to encode human body configuration for 3d
  pose estimation,''
\newblock in {\em Proc. of the AAAI Conference on Artificial Intelligence},
  2018.

\bibitem{sun2017compositional}
Xiao Sun, Jiaxiang Shang, Shuang Liang, and Yichen Wei,
\newblock ``Compositional human pose regression,''
\newblock in {\em The IEEE International Conference on Computer Vision (ICCV)},
  2017, vol.~2, p.~7.

\bibitem{mono-3dhp2017}
Dushyant Mehta, Helge Rhodin, Dan Casas, Pascal Fua, Oleksandr Sotnychenko,
  Weipeng Xu, and Christian Theobalt,
\newblock ``Monocular 3d human pose estimation in the wild using improved cnn
  supervision,''
\newblock in {\em 3D Vision (3DV), 2017 Fifth International Conference on}.
  IEEE, 2017.

\bibitem{ionescu2014human3}
Catalin Ionescu, Dragos Papava, Vlad Olaru, and Cristian Sminchisescu,
\newblock ``Human3. 6m: Large scale datasets and predictive methods for 3d
  human sensing in natural environments,''
\newblock {\em IEEE transactions on pattern analysis and machine intelligence},
  vol. 36, no. 7, pp. 1325--1339, 2014.

\bibitem{li20143d}
Sijin Li and Antoni~B Chan,
\newblock ``3d human pose estimation from monocular images with deep
  convolutional neural network,''
\newblock in {\em Asian Conference on Computer Vision}. Springer, 2014, pp.
  332--347.

\end{thebibliography}

\end{document}